\documentclass[10pt,journal,compsoc]{IEEEtran}
\ifCLASSOPTIONcompsoc
  \usepackage[nocompress]{cite}
\else
  \usepackage{cite}
\fi

\usepackage[pdftex]{graphicx}

\usepackage{amsmath}
\usepackage{amsfonts}
\usepackage{booktabs}
\usepackage{multirow}
\usepackage{color}
\usepackage{balance}
\usepackage{tabularx}
\usepackage{etoolbox}

\makeatletter
\patchcmd{\@makecaption}
  {\\}
  {\,: }
  {}
  {}
\makeatother
\usepackage{enumitem}
\usepackage{makecell}
\usepackage[backref,breaklinks,colorlinks]{hyperref}  
\usepackage{url}
\usepackage{algorithm}
\usepackage{algpseudocode}
\usepackage{xspace}
\usepackage[edges]{forest}
\definecolor{hidden-draw}{RGB}{20,68,106}
\definecolor{hidden-pink}{RGB}{255,245,247}

\newtheorem{definition}{Definition}

\makeatletter
\def\UrlAlphabet{%
      \do\a\do\b\do\c\do\d\do\e\do\f\do\g\do\h\do\i\do\j%
      \do\k\do\l\do\m\do\n\do\o\do\p\do\q\do\r\do\s\do\t%
      \do\u\do\v\do\w\do\x\do\y\do\z\do\A\do\B\do\C\do\D%
      \do\E\do\F\do\G\do\H\do\I\do\J\do\K\do\L\do\M\do\N%
      \do\O\do\P\do\Q\do\R\do\S\do\T\do\U\do\V\do\W\do\X%
      \do\Y\do\Z}
\def\UrlDigits{\do\1\do\2\do\3\do\4\do\5\do\6\do\7\do\8\do\9\do\0}
\g@addto@macro{\UrlBreaks}{\UrlOrds}
\g@addto@macro{\UrlBreaks}{\UrlAlphabet}
\g@addto@macro{\UrlBreaks}{\UrlDigits}
\makeatother

\newcommand{\para}[1]{\textbf{#1}.\xspace}

\newcommand{\tabref}[1]{Table \ref{#1}}

\newcommand{\LLM}{LLM\xspace}
\newcommand{\LLMs}{LLMs\xspace}

\newcommand{\MLLMs}{MLLMs\xspace}
\newcommand{\MLLM}{MLLM\xspace}
\newcommand{\MModal}{Multi-Modal\xspace}
\newcommand{\Mmodal}{Multi-modal\xspace}
\newcommand{\mmodal}{multi-modal\xspace}
\newcommand{\CODEV}{Data-Model Co-Development\xspace}
\newcommand{\Codev}{Data-model co-development\xspace}
\newcommand{\codev}{data-model co-development\xspace}

\begin{document}
\title{The Synergy between Data and Multi-Modal Large Language Models: A Survey from Co-Development Perspective}

\author{Zhen~Qin*,
Daoyuan~Chen*,~
Wenhao~Zhang,~
Liuyi~Yao,~
Yilun~Huang,~ \\
Bolin~Ding,
Yaliang~Li$\dag$,
Shuiguang~Deng$\dag$,~\IEEEmembership{Senior Member,~IEEE}
\IEEEcompsocitemizethanks{
\IEEEcompsocthanksitem * Equal Contributions. Work done during Z. Qin's internship at Alibaba Group. ~~~~~~ $\dag$ Corresponding Authors 
\IEEEcompsocthanksitem Z. Qin and S. Deng are with the College of Computer Science and Technology, Zhejiang University.
E-mail: \{zhenqin, dengsg\}@zju.edu.cn 
\IEEEcompsocthanksitem D. Chen, W. Zhang, L. Yao, Y. Huang, B. Ding and Y. Li are with Alibaba Group. 
E-mail: \{daoyuanchen.cdy, zwh434786, yly287738, lielin.hyl, bolin.ding, yaliang.li\}@alibaba-inc.com
}
}

\IEEEtitleabstractindextext{%
\begin{abstract}
The rapid development of large language models (\LLMs) has been witnessed in recent years. 
Based on the powerful \LLMs, \mmodal \LLMs (\MLLMs) extend the modality from text to a broader spectrum of domains, attracting widespread attention due to the broader range of application scenarios.
As \LLMs and \MLLMs rely on vast amounts of model parameters and data to achieve emergent capabilities, the importance of data is receiving increasingly widespread attention and recognition.
Tracing and analyzing recent data-oriented works for \MLLMs, we find that the development of models and data is not two separate paths but rather interconnected.
On the one hand, vaster and higher-quality data contribute to better performance of \MLLMs; on the other hand, \MLLMs can facilitate the development of data. 
The co-development of \mmodal data and \MLLMs requires a clear view of 1) at which development stages of \MLLMs specific data-centric approaches can be employed to enhance certain \MLLM capabilities, and 2) how \MLLMs, utilizing those capabilities, can contribute to \mmodal data in specific roles. 
To promote the \codev for \MLLM community, we systematically review existing works related to \MLLMs from the \codev perspective. 
A regularly maintained project associated with this survey is accessible at \url{https://github.com/modelscope/data-juicer/blob/main/docs/awesome_llm_data.md}.
\end{abstract}

}
\maketitle

\section{Introduction}
\label{sec-intro}
\begin{figure}
    \centering
    \includegraphics[width=0.66\linewidth]{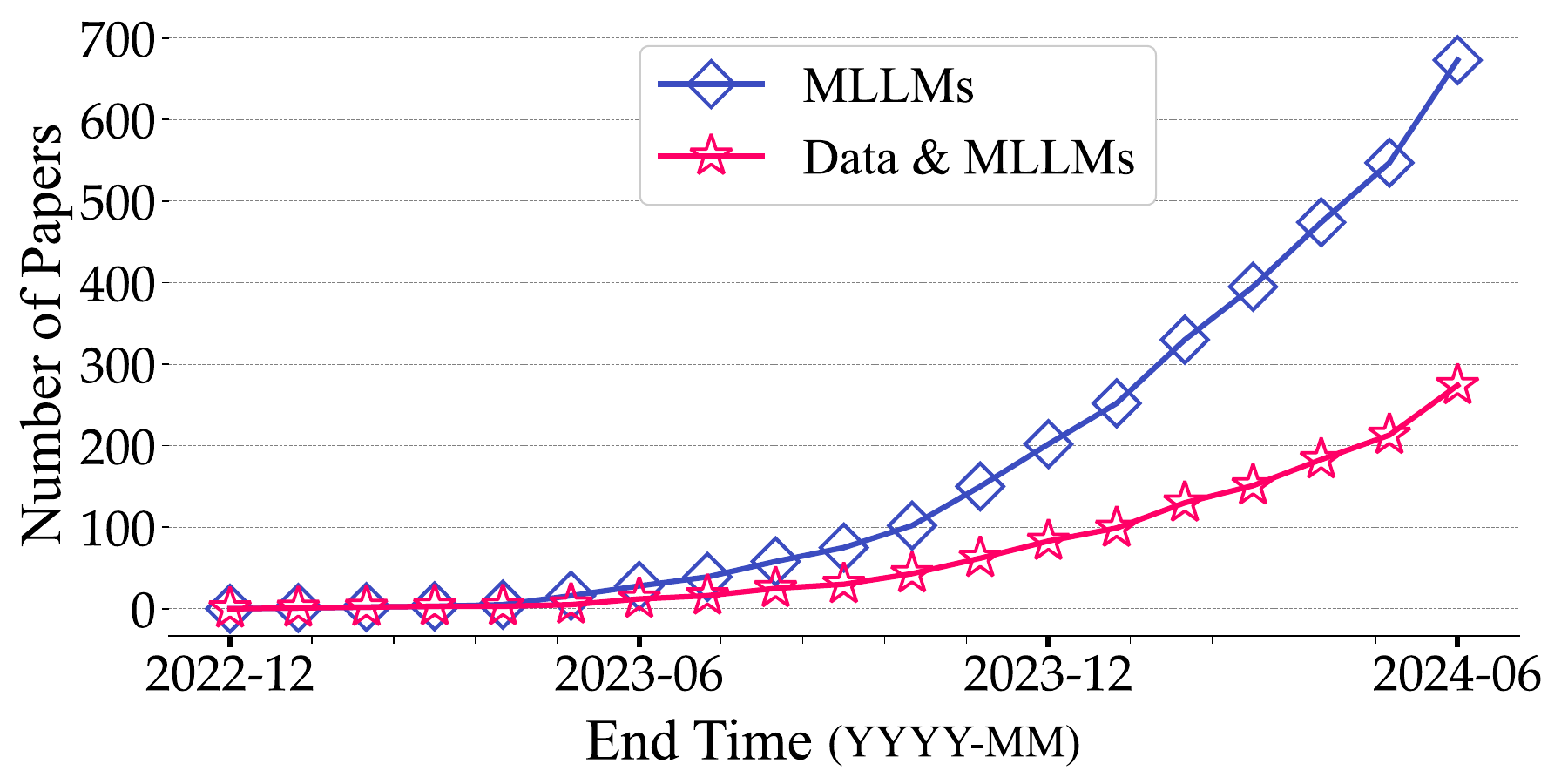}
    \caption{The trends of cumulative numbers of papers on arXiv \protect\footnotemark[1] related to \emph{\MLLMs} and those related to \emph{both \MLLMs and data}, respectively.}
    \label{pic-paper-trend}
\end{figure}

\IEEEPARstart{L}{arge} language models (\LLMs) demonstrate impressive performances across a wide range of tasks in recent years, with their associated technologies making significant advancements. 
Since human senses are not limited to text modality, \mmodal \LLMs (\MLLMs) have come into view, such as \texttt{Gemini-1.5} \cite{reid2024gemini} and \texttt{Sora} \cite{openai2024sora} that are capable of processing inputs or outputs in modalities beyond text, and \texttt{GPT-4o} \cite{openai2024gpt4o} and \texttt{NExT-GPT} \cite{wu2023nextgpt} that can even interact between multiple modalities in both input and output.
\MLLMs have gained widespread attention in the past two years. 
As shown in Fig. \ref{pic-paper-trend}, research related to \MLLMs has been emerging at an increasing speed since 2023.

The outstanding performance of \MLLMs stems from the emergent abilities brought by the scaling up in the number of parameters \cite{survey-zhao2023survey}. 
Many works indicate that scaling up the model size needs to be complemented by even more massive amounts of data \cite{villalobos2022will,goyal2024scaling,gao2024lumina}, such as scaling law \cite{kaplan2020scaling,aghajanyan2023scaling}. 
In light of this, a series of works shift the focus from merely model architectures and training techniques to data-centric approaches that focus on the curation of data \cite{survey-zha2023datacentric,Salehi2023Data-green,survey-jakubik2024datacentric,gadre2023datacomp,fan2023improving,he2024efficient}, which serve as the basis to unlock the potential of large models. 
From Fig. \ref{pic-paper-trend}, among existing papers for \MLLMs, those related to data-centric approaches likewise exhibit a strong growth trend on the count and occupy a significantly important portion. 
\footnotetext[1]{
The statistics are obtained with the advanced search of arXiv, merely for reflecting trends and may not be absolutely precise on numerical results, especially for subtle connections to \mmodal data.
}

As technical works related to \MLLMs continue to emerge, some reviews for \MLLMs have gradually appeared as well \cite{
mmsurvey-carolan2024review,
mmsurvey-caffagni2024revolution,
mmsurvey-xu2024survey,
mmsurvey-li2023multimodal,
mmsurvey-zhang2024mmllms,
mmsurvey-10445007,
liu2024multimodal,
tang2024video,
survey-jin2024efficient,
mmsurvey-huang2023visual,
mmsurvey-yin2024survey,
mmsurvey-wu2023multimodal,
survey-10123038,
zhao2024deep,
mmsurvey-wang2024exploring,
zhou2024survey,
mmsurvey-zhao2024survey
}.
These surveys are mainly conducted from the model-centric perspective, however, the importance of the data needs further emphasis.
One recent survey emphasizes the data-centric perspective for \MLLMs \cite{bai2024survey}, organizing existing data-centric approaches according to the stages in the proposed data pipeline. 

Actually, \emph{the development of data and models is intertwined rather than separate}. 
Large-scale and high-quality data enhance model performance, while well-trained models can help further improve the data.
It reduces labor costs and expands data quantity, and has been successfully demonstrated by the training of Segment Anything model (SAM) \cite{kirillov2023segment} which utilizes segmentation masks that need to be annotated for training. 
As the SAM becomes more proficient with training, it gradually replaces humans in annotation tasks, and thereby a repetitive cycle that improves both the model and the dataset is formed. 
Such a progressive and virtuous cycle advances \MLLM development, i.e., \MLLMs benefited from high-quality datasets can help improve the training data, which in turn further enhances the \MLLMs. 

\Codev is promising for \MLLMs but has not been fully investigated. 
From our investigation, there currently lacks a review for \MLLMs from the perspective of \codev.
Existing surveys have yet to establish a relationship between data-centric approaches and \MLLM capabilities, and have not clearly articulated how the capabilities of \MLLMs can assist in datasets.
The key to enabling the \codev for \MLLMs lies in elucidating which data approaches can enhance each specific capability of \MLLMs, as well as understanding the roles that models can assume to improve \mmodal data. 
Thus, this survey aims to advance the \codev for \MLLMs by answering the following research questions with a comprehensive review: 
\begin{itemize}[leftmargin=1em]
    \item \textbf{RQ1}: At which stage throughout the lifecycle of an \MLLM can specific data-centric approaches be employed to enhance specific \MLLM capabilities?
    \item \textbf{RQ2}: What roles can models play in facilitating the curation of \mmodal data, and what specific capabilities of models are leveraged in each case?
\end{itemize}

To answer these two key research questions, we first propose a novel taxonomy grounded in the \codev paradigm for \MLLMs. 
We categorize previous efforts into two principal dual types: 
the data contributions to models and the reciprocal model contributions to data, establishing their in-depth connections anchored in the capabilities of \MLLMs. 
Subsequently, we provide a comprehensive examination of existing works for \MLLMs from the \codev perspective, which uncovers considerable untapped potential for advancing the \codev paradigm, primarily due to the lack of dedicated focus on the synergistic interplay between data and models.
Built upon the insights garnered, we delineate several progressive future directions in the \codev of \MLLMs, forming a roadmap to better leverage the complementarity between data and models, spanning from infrastructures to various self-boosting degrees of \codev.
The main contributions of this survey are three-fold: 
\begin{itemize}[leftmargin=1em]
    \item \textbf{A New Perspective for \MLLM Development}: we propose a new taxonomy that emphasizes the synergy between \mmodal data and \MLLMs, aiming to mine the mutual benefits for both data and model development. This taxonomy is systematically organized based on the hierarchy of data-related techniques essential for developing \MLLMs, offering a clear view of the whole life cycle to advance \MLLMs for researchers and developers.
    \item \textbf{An Up-to-date Review for \MLLMs from \CODEV Perspective}: we systematically review the fast-growing works on \MLLMs and elucidate 1) which \MLLM capabilities can be enhanced by specific data-centric approaches, and 2) how the capabilities of well-trained models in turn nourish \mmodal data. To the best of our knowledge, this is the \emph{first} survey on \MLLMs from \codev perspective. 
    \item \textbf{A Roadmap for Future \MLLMs}: focusing on the internal interplay between data and models for \MLLMs, we provide a roadmap progressively organized by several advanced and promising directions.
    With this work, we hope to offer a source of inspiration and guidance to both academic researchers and industry practitioners navigating the evolving landscape of \MLLMs.
\end{itemize}

\para{Organizations} 
The rest of this survey is organized as follows. 
Sec. \ref{sec-preliminary} provides preliminaries, including the background, taxonomy, comparisons to related surveys, and \MLLM architectures.
Sec. \ref{sec-data-contrib-scaling} discusses data contributions for scaling up \MLLMs.
Sec. \ref{sec-data-contrib-usability} reviews data contributions for improving the usability of \MLLMs.
Sec. \ref{sec-model-contrib-synthesis} describes the capabilities of models that directly help dataset curation for \MLLMs.
Sec. \ref{sec-model-contrib-application} explores the applications of models acting as data scientists to assist in dataset curation for \MLLMs.
Sec. \ref{sec-resources} lists some public datasets for \MLLMs, with the participation of models for curation indicated.
Sec. \ref{sec-future} discusses a roadmap for future development of \MLLMs.
\section{Preliminary}
\label{sec-preliminary}

\subsection{Background}
\label{subsec-preliminary-background}
\begin{figure*}
  \centering
  \includegraphics[width=\linewidth]{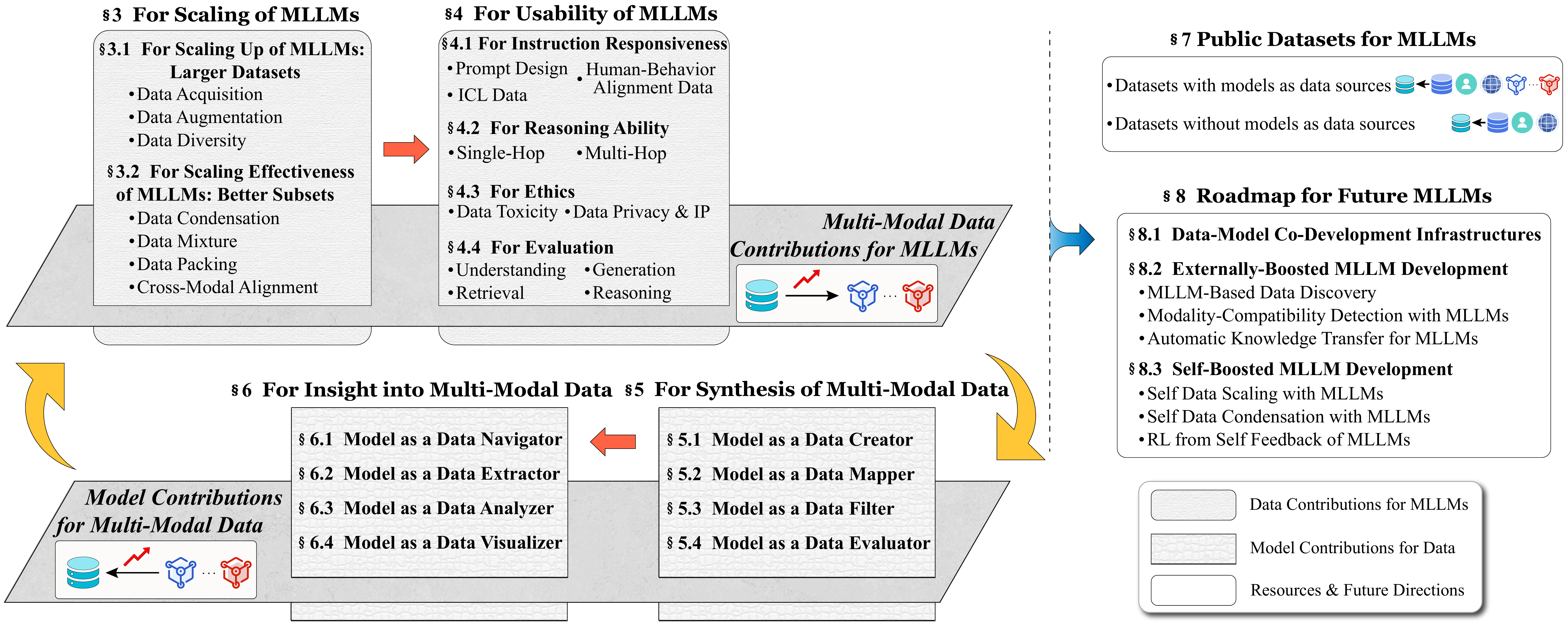}
  \caption{
    Taxonomy for \MLLMs from the \codev perspective and overview of \S\ref{sec-data-contrib-scaling}-\ref{sec-future} with their inter-relationships.
    Data contributions for \MLLMs (\S\ref{sec-data-contrib-scaling} \& \ref{sec-data-contrib-usability}) are organized in an objective-driven manner and ordered according to \MLLM development stages, i.e., first scaling up for better performance then improving the usability. 
    Model contributions for data (\S\ref{sec-model-contrib-synthesis} \& \ref{sec-model-contrib-application}) are organized by the roles played by models. 
  }
  \label{pic-taxonomy}
\end{figure*}
\textbf{\Codev} for \MLLMs attempts to improve model performance while leveraging this model to optimize the data, with the final objective aimed at a well-performing \MLLM. 
It is characterized by \textit{dynamic} training data for a \textit{dynamic} model \cite{xu2023cit}. 
Given the current lack of a formal definition for \codev, we attempt to provide a formal definition with a large \mmodal generation model as an example, which is capable of generating outputs in modalities different from the input or in more modalities than the input, such as image-to-text dialog \cite{chen2024allava} and text-to-image synthesis \cite{rombach2022high}.
\begin{definition}
\label{def-co-development}
(\CODEV for Large Generation Model). 
Let 
1) $p_{data}(\mathbf{u})$ denote an ideal distribution that each real-world data $\mathbf{u}$ follows, where $\mathbf{u}$ is generally composed by a context $\mathbf{c}$ and a response $\mathbf{r}$, denoted by $\mathbf{u} = \left\{\mathbf{c}, \mathbf{r} \right\}$, 
2) $\mathbf{w}$ be the large generation model to be trained, which generates a response $\mathbf{r}'$ given a context $\mathbf{c}'$, as $\mathbf{r}' = \mathbf{w}(\mathbf{c}')$, 
3) $\mathcal{D} = \left\{\mathbf{u}^{\mathcal{D}}_1, \mathbf{u}^{\mathcal{D}}_2, \ldots \right\}$ be the dataset to train $\mathbf{w}$, which is usually characterized by large scale, 
4) $\mathcal{D}^*$ denote the ideally optimal $\mathcal{D}$, 
and 
5) $Q(\cdot,\cdot)$ denote the function that quantifies the similarity between two data samples, with a larger value indicating higher similarity, 
\codev can be formalized as a bi-level optimization problem, as
\begin{align}
\label{eq-obj-model}
\!\!\!\!\!\!&\max_{\mathbf{w}} \ \  \mathbb{E}_{\left\{\mathbf{c}, \mathbf{r}\right\}\in\mathcal{D}^*} Q(\mathbf{r}, \mathbf{w}(\mathbf{c})), \\
\label{eq-obj-data}
\!\!\!\!\!\!\text{s.t.} \ \  \mathcal{D}^* \in \arg & \min_{\mathcal{D}} \ \mathbb{E}_{\mathbf{u} \sim p_{data}(\mathbf{u}), \forall\mathbf{x}=\left\{\mathbf{c}, \mathbf{r}\right\} \in \mathcal{D}} - Q(\mathbf{u}, \mathbf{x}).
\end{align}
\end{definition}
Eq. \eqref{eq-obj-data} curates data for training $\mathbf{w}$. 
Eq. \eqref{eq-obj-model} is the final objective that teaches $\mathbf{w}$ to produce data resembling real-world data. 
\Codev can be classified into two paradigms based on the optimization tools for $\mathcal{D}$: 
\begin{enumerate}[leftmargin=1em]
    \item (Self-Boosted Paradigm). The model $\mathbf{w}$ to be trained is also used to improve dataset $\mathcal{D}$, where Eq. \eqref{eq-obj-data} and Eq. \eqref{eq-obj-model} are usually optimized alternatively. In this paradigm, Eq. \eqref{eq-obj-data} can be further reformulated as $\mathcal{D}^* \in \arg \min_{\mathcal{D} \mid \mathbf{w}} \ \mathbb{E}_{\mathbf{u} \sim p_{data}(\mathbf{u}), \forall\mathbf{x}=\left\{\mathbf{c}, \mathbf{r}\right\} \in \mathcal{D}} Q(\mathbf{u}, \mathbf{x})$.
    \item (Externally-Boosted Paradigm). As an alternative, $\mathcal{D}$ can be curated with a well-trained model $\mathbf{w}^*$ such as \texttt{GPT-4V}, or even human efforts. Accordingly, we can rewrite Eq. \eqref{eq-obj-data} as $\mathcal{D}^* \in \arg \min_{\mathcal{D} \mid \mathbf{w}^*} \ \mathbb{E}_{\mathbf{u} \sim p_{data}(\mathbf{u}), \forall\mathbf{x}=\left\{\mathbf{c}, \mathbf{r}\right\} \in \mathcal{D}} Q(\mathbf{u}, \mathbf{x})$
    or 
    $\mathcal{D}^* \in \arg \min_{\mathcal{D} \mid \text{human}} \ \mathbb{E}_{\mathbf{u} \sim p_{data}(\mathbf{u}), \forall\mathbf{x}=\left\{\mathbf{c}, \mathbf{r}\right\} \in \mathcal{D}} Q(\mathbf{u}, \mathbf{x})$. 
\end{enumerate}
From our survey, the self-boosted paradigm is proven effective for uni-modal \LLMs \cite{li2023self} and vision foundation models \cite{xu2023cit,kirillov2023segment}, but still lacks investigation for \MLLMs. 

\Codev is promising as increasing attention shifts towards data-centric approaches, where data often serves as the primary variable, rather than merely focusing on model architectures \cite{survey-jakubik2024datacentric,Salehi2023Data-green,survey-zha2023data,survey-zha2023datacentric,he2024efficient,bai2024survey}. 
As \MLLMs require increasingly large volumes of data, models are gradually used to assist or directly build the data samples. 
Thus, the development of data and models have become interdependent and inseparable: massive and high-quality data can lead to well-performing \MLLMs, and in turn, well-performing \MLLMs can help construct more high-quality data. 
Therefore, it is necessary to understand how data approaches enhance specific capabilities of \MLLMs, and how \MLLMs assist in data approaches, thereby advancing the \codev for \MLLMs.

\subsection{Taxonomy}
\label{subsec-preliminary-taxonomy}
\begin{table*}[t]
  \centering
  \renewcommand\arraystretch{0.8}
  \caption{Qualitative comparisons between our survey and closely related surveys. 
  A survey may appear in more than one row due to the diversity of emphases. 
  Model-centric surveys for uni-modal \LLMs are omitted due to their secondary relevance to this survey.
  }
  \label{tab-related}
  \setlength\tabcolsep{3pt}
  \begin{tabularx}{\linewidth}{ccllX}
    \toprule[1.0pt]
    \multicolumn{1}{c}{\multirow{1}{*}{Modality}} & 
    \multicolumn{1}{c}{\multirow{1}{*}{Perspective}} &
    \multicolumn{1}{c}{\multirow{1}{*}{Taxonomy}} & 
    \multicolumn{1}{c}{\multirow{1}{*}{Highlights}} & 
    \multicolumn{1}{c}{\multirow{1}{*}{References}} \\
    \midrule[1.0pt]
    \multirow{6}{*}{\MModal\vspace{-1.9cm}} & 
    \multirow{5}{*}{Model-Centric\vspace{-1.6cm}}  & 
    \multirow{1}{*}{Based on Training Techniques\vspace{-0.3cm}} &
    \multirow{1}{*}{Training stages \& algorithms for \MLLMs\vspace{-0.3cm}} &
    \cite{mmsurvey-carolan2024review,mmsurvey-caffagni2024revolution,mmsurvey-xu2024survey,mmsurvey-li2023multimodal,mmsurvey-zhang2024mmllms,mmsurvey-10445007,liu2024multimodal,tang2024video,survey-jin2024efficient} \\
    \cmidrule{3-5}
     &
     &
    \multirow{1}{*}{Based on \MLLM Architectures\vspace{-0.3cm}} &
    \multirow{1}{*}{The architectural components of \MLLMs\vspace{-0.3cm}} &
    \cite{mmsurvey-huang2023visual,mmsurvey-yin2024survey,mmsurvey-wu2023multimodal,mmsurvey-caffagni2024revolution,mmsurvey-xu2024survey,mmsurvey-zhang2024mmllms,survey-10123038,mmsurvey-10445007,zhao2024deep,survey-jin2024efficient} \\
    \cmidrule{3-5}
     & 
     &
    \multirow{2}{*}{Based on \MLLM Capabilities\vspace{-0.15cm}} & 
    Reasoning abilities of \MLLMs &
    \cite{mmsurvey-yin2024survey,mmsurvey-li2023multimodal,mmsurvey-wang2024exploring} \\
    \cmidrule{4-5}
     & 
     &
     &
    Applications of \MLLMs &
    \cite{zhou2024survey,mmsurvey-caffagni2024revolution} \\
    \cmidrule{3-5}
     &
     &
    Based on \MLLM Systems &
    \makecell[l]{Key considerations \& applications \\ in \MLLM System Design} &
    \cite{mmsurvey-zhao2024survey,mmsurvey-xu2024survey} \\
    \cmidrule{2-5}
     &
    Data-Centric & 
    Based on Data Pipeline & 
    \makecell[l]{Data approaches for \MLLMs across stages \\ in the data pipeline} &
    \cite{bai2024survey} \\
    \cmidrule{1-5}
    \multirow{2}{*}{Uni-Modal\vspace{-0.9cm}} & 
    \multirow{2}{*}{Data-Centric\vspace{-0.9cm}} &
     Based on Data Pipeline & 
    \makecell[l]{Data approaches for \LLMs across stages \\ in the data pipeline} &
    \cite{survey-zha2023datacentric,survey-wang2023data,long2024llms} \\
    \cmidrule{3-5}
     &
     &
    Based on Training Stage &
    Data-centric approaches for each training stage &
    \cite{survey-albalak2024survey-selection} \\
    \cmidrule{3-5}
     &
     &
    Based on Adopted Techniques & 
    Techniques for specific data approaches&
    \cite{survey-wang2024survey-selection,survey-ding2024data-aug,Salehi2023Data-green} \\
    \midrule[0.9pt]
    \multirow{1}{*}{\MModal} & 
    \vspace{-0.6cm}\makecell[c]{\textbf{Data-Model} \\ \textbf{Co-Development}}\vspace{0.6cm} &
    \makecell[l]{Based on \textbf{Mutual Benefits} \\ ~~\textbf{between Model \& Data}} & 
    \makecell[l]{
      $\bullet$ Matching data-centric approaches to \MLLM \\ properties  (\textbf{data contributions for \MLLMs}) \vspace{0.1cm}\\
      $\bullet$ Roles acted by models to facilitate \mmodal \\ data (\textbf{model contributions for \mmodal data})
    }    &
    \multirow{1}{*}{\textbf{Ours}} \\
    \bottomrule[1.0pt]
  \end{tabularx}
\end{table*}
The taxonomy and the relationships between the items are illustrated in Fig. \ref{pic-taxonomy}. 
According to our investigations, both the contributions of data to models and vice versa can be categorized into two major types.
The data contributions to models are organized in an \textit{objective-driven} manner and arranged according to their \textit{sequence in technical stages} of \MLLM development. 
First, to elicit foundational abilities from \MLLMs, it is crucial to provide more and higher-quality data for \MLLM training, aiming at the scaling of \MLLMs (\S\ref{sec-data-contrib-scaling}).
Some works focus on providing large-scale datasets to scale up \MLLMs (\S\ref{subsec-data-contrib-scaling-up}), while others enhance the effectiveness of scaling by improving data quality and organizing the data strategically (\S\ref{subsec-data-contrib-scaling-effectiveness}).
After obtaining an \MLLM with foundational capabilities, a series of approaches could be performed around data to enhance its usability from various aspects, including the instruction responsiveness (\S\ref{subsec-data-contrib-usability-following}), reasoning ability (\S\ref{subsec-data-contrib-usability-reasoning}) and ethics (\S\ref{subsec-data-contrib-usability-ethic}), followed by comprehensive evaluations (\S\ref{subsec-data-contrib-usability-benchmark}).

As datasets grow in size, their curation gradually relies on well-trained models (\MLLMs or their components such as \LLMs and foundation models). 
We review a series of data-centric approaches co-piloted with models and group them into two main categories, organized in a \textit{role-driven} manner and arranged according to their \textit{sequence in a data pipeline}, i.e., 
1) for the synthesis of data, where models directly participate in data curation to alleviate repetitive tasks for humans by acting as a data creator (\S\ref{subsec-model-contrib-synthesis-creator}), mapper (\S\ref{subsec-model-contrib-synthesis-mapper}), filter (\S\ref{subsec-model-contrib-synthesis-filter}), and evaluator (\S\ref{subsec-model-contrib-synthesis-evaluator}); 
and 
2) for insights into data, where models perform as data scientists to provide insights on \mmodal data by acting as a data navigator (\S\ref{subsec-model-contrib-application-navigator}), analyzer (\S\ref{subsec-model-contrib-application-analyzer}), extractor (\S\ref{subsec-model-contrib-application-extractor}), and visualizer (\S\ref{subsec-model-contrib-application-visualizer}). 

Based on these investigations, we summarize the public datasets for \MLLMs and clarify the models' participation during data curation (\S\ref{sec-resources}).
Finally, we identify some progressive future directions for \MLLMs, forming a roadmap for the future development of \MLLMs (\S\ref{sec-future}).

\subsection{Differences from Related Surveys}
\label{subsec-preliminary-related}
The popularity of \MLLMs has led researchers to catalog existing works.
Current surveys on \MLLMs mainly focus on model-centric perspectives with the taxonomy based on: 
1) training techniques that highlight the training stages and algorithms \cite{
mmsurvey-carolan2024review,
mmsurvey-caffagni2024revolution,
mmsurvey-xu2024survey,
mmsurvey-li2023multimodal,
mmsurvey-zhang2024mmllms,
mmsurvey-10445007,
liu2024multimodal,
tang2024video,
survey-jin2024efficient
};
2) \MLLM architectures \cite{
mmsurvey-huang2023visual,
mmsurvey-yin2024survey,
mmsurvey-wu2023multimodal,
mmsurvey-caffagni2024revolution,
mmsurvey-xu2024survey,
mmsurvey-zhang2024mmllms,
survey-10123038,
mmsurvey-10445007,
zhao2024deep,
survey-jin2024efficient
};
3) \MLLM capabilities for reasoning \cite{
mmsurvey-yin2024survey,
mmsurvey-li2023multimodal,
mmsurvey-wang2024exploring
} 
and 
applications 
\cite{
zhou2024survey,
mmsurvey-caffagni2024revolution
};
or 
4) \MLLM systems which focus on the key considerations and applications \cite{
mmsurvey-zhao2024survey,
mmsurvey-xu2024survey
}.
These model-centric surveys facilitate the development of \MLLMs, yet have not given dedicated consideration to data. 

The importance of data for \LLMs has been emphasized by some \LLM surveys from data-centric perspectives, with the taxonomy based on: 
1) data approaches across stages in the data pipeline \cite{survey-zha2023datacentric,survey-wang2023data,long2024llms};
2) training stages \cite{survey-albalak2024survey-selection};
or 
3) the adopted techniques for specific data-centric approaches \cite{survey-wang2024survey-selection,survey-ding2024data-aug,Salehi2023Data-green}. 
These surveys foster the attention of \LLM communities toward a data-centric perspective.
One recent survey extends the data-centric perspective from \LLMs to \MLLMs \cite{bai2024survey}, which organizes existing works according to their sequence in the data pipeline. 

Given 1) the potential of \codev for \MLLMs, and 2) the fact that existing data-centric surveys fall short of establishing connections between data-centric approaches and specific \MLLM capabilities, 
there is currently a lack of and an urgent need for a survey to articulate the contributions made by \mmodal data and \MLLMs to each other.
Thus, we provide this up-to-date survey on \MLLMs from a \codev perspective, clarifying how data technologies facilitate the development of \MLLMs and then demonstrating how models can promote \mmodal data technologies.
The above comparisons between the related surveys and ours are summarized in \tabref{tab-related}.

\subsection{\MLLM Architecture}
We briefly introduce \MLLM architecture for reference.
An \MLLM typically contains: 
1) an \LLM such as \texttt{LLaMA} \cite{touvron2023llama}; 
2) one or more foundation models (encoders) to encode non-text data (e.g., \texttt{ViT} \cite{dosovitskiy2021an} and \texttt{CLIP} \cite{radford2021learning}); 
and 
3) one or more projectors to align the encoded features of non-text data with the feature space of \LLMs. 
This setup allows \MLLMs to generate textual responses to \mmodal inputs \cite{mmsurvey-yin2024survey}. 
Further, by incorporating 4) modality-specific generators such as \texttt{Stable Diffusion} \cite{rombach2022high}, \MLLMs can produce \mmodal contents \cite{wu2023nextgpt,gao2024lumina}.
These components may be trained in different stages \cite{liu2023visual} and with different types of datasets \cite{gao2023llama,chen2024allava}. 
We do not focus on the pretraining of \LLMs, for which we refer readers to the surveys in Table \ref{tab-related}.
\section{\MModal Data Contributions for \MLLMs: Scaling}
\label{sec-data-contrib-scaling}
\definecolor{color-layer3}{HTML}{03497E}
\definecolor{color-layer2}{HTML}{0596D5}
\definecolor{color-layer1}{HTML}{9DEBFC}
\definecolor{color-layer0}{HTML}{FEB228}

\tikzstyle{leaf-style}=[
    rectangle,
    draw=hidden-draw,
    rounded corners,
    text opacity=1,
    minimum height=1.5em,
    minimum width=5em,
    inner sep=2pt,
    align=left,
    fill opacity=.5,
    line width=0.9pt,
]
\tikzstyle{leaf}=[leaf-style, minimum height=1.5em,
    fill=color-layer3!30, text=black, align=left,font=\scriptsize,
    inner xsep=2pt,
    inner ysep=2pt,
    line width=0.8pt,
]

\begin{figure*}[t]
    \centering
    \begin{forest}
        forked edges,
        for tree={
            grow=east,
            reversed=true,
            anchor=base west,
            parent anchor=east,
            child anchor=west,
            base=center,
            font=\small,
            rectangle,
            draw=hidden-draw,
            rounded corners,
            align=center,
            text centered,
            minimum width=4em,
            edge+={darkgray, line width=1pt},
            s sep=2pt,
            inner xsep=2pt,
            inner ysep=2pt,
            line width=0.8pt,
            ver/.style={rotate=90, child anchor=north, parent anchor=south, anchor=center},
        },
        where level=1{text width=13em,font=\footnotesize,}{},
        where level=2{text width=13em,font=\scriptsize,}{},
        where level=3{text width=13.5em,font=\scriptsize,}{},
        [
            \textbf{\MModal Data Contributions}\\\textbf{for \MLLMs: Scaling}, ver, fill=color-layer0!30
            [
                \textbf{For Scaling Up of \MLLMs:}\\\textbf{Larger Datasets} (\S\ref{subsec-data-contrib-scaling-up}), fill=color-layer1!30
                [
                    \textbf{Data Acquisition} (\S\ref{subsubsec-data-contrib-scaling-up-acquisition}), fill=color-layer2!30
                    [
                        For Encoders and Decoders, leaf
                    ]
                    [
                        For Projectors, leaf
                    ]
                    [
                        For Fine-Tuning, leaf
                    ]
                ]
                [
                    \textbf{Data Augmentation} (\S\ref{subsubsec-data-contrib-scaling-up-augmentation}), fill=color-layer2!30
                    [
                        Traditional Random Augmentation, leaf
                    ]
                    [
                        Generative Augmentation, leaf
                    ]
                ]
                [
                    \textbf{Data Diversity} (\S\ref{subsubsec-data-contrib-scaling-up-diversity}), fill=color-layer2!30
                    [
                        For Single-Modality Perception Abilities, leaf
                    ]
                    [
                        For Cross-Modality Cognition Abilities, leaf
                    ]
                ]
            ]
            [
                \textbf{For Scaling Effectiveness of}\\\textbf{MLLMs: Better Subsets} (\S\ref{subsec-data-contrib-scaling-effectiveness}), fill=color-layer1!30
                [
                    \textbf{Data Condensation} (\S\ref{subsubsec-data-contrib-scaling-effectiveness-condensation}), fill=color-layer2!30
                    [
                        Data Deduplication, leaf
                    ]
                    [
                        Low-Quality Data Filtering, leaf
                    ]
                    [
                        Kernel Set Construction, leaf
                    ]
                ]
                [
                    \textbf{Data Mixture} (\S\ref{subsubsec-data-contrib-scaling-effectiveness-condensation}), fill=color-layer2!30
                    [
                        Mitigating Distribution Bias, leaf
                    ]
                    [
                        Exploiting Distribution Bias, leaf
                    ]
                ]
                [
                    \textbf{Data Packing} (\S\ref{subsubsec-data-contrib-scaling-effectiveness-packing}), fill=color-layer2!30
                    [
                        For Better Pretraining, leaf
                    ]
                    [
                        For Long-Context Support, leaf
                    ]
                ]
                [
                    \textbf{Cross-Modal Alignment} (\S\ref{subsubsec-data-contrib-scaling-effectiveness-crossmodal-alignment}), fill=color-layer2!30
                    [
                        Joint Embedding Space, leaf
                    ]
                    [
                        Text-Centric Anchoring, leaf
                    ]
                ]
            ]
        ]
        \end{forest}
    \caption{Organization of data approaches tailored for scaling of \MLLMs.}
    \label{pic-data4model-scaling}
\end{figure*}
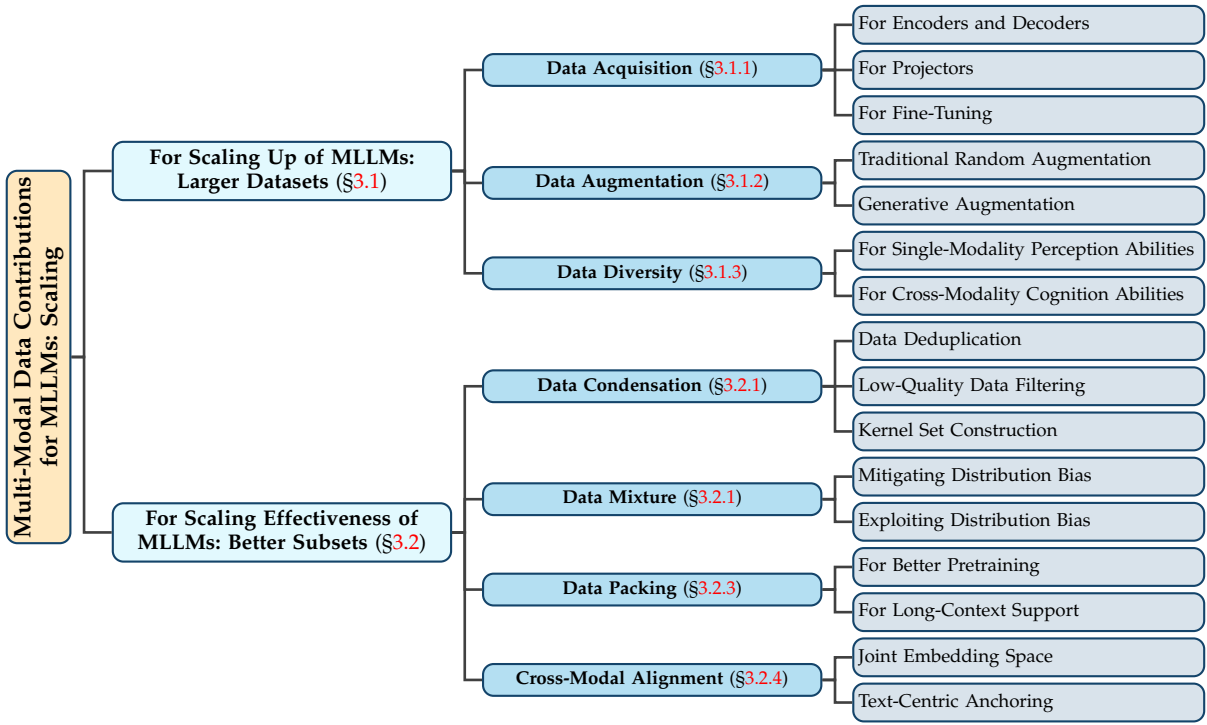
Building \MLLMs with satisfactory performance requires large-scale and high-quality \mmodal data.
This section summarizes existing works contributing to large-scale \MLLMs by providing datasets, organized by the \textit{logical sequence of top-level design and iterative optimization}, i.e., first to scale up the cardinality in \mmodal datasets (\S\ref{subsec-data-contrib-scaling-up}), followed by enhancing the scaling effectiveness of datasets (\S\ref{subsec-data-contrib-scaling-effectiveness}). 
The organization of this section is illustrated in Fig. \ref{pic-data4model-scaling}. 
After introducing existing works, a brief summarization and discussion are provided in Sec. \ref{subsec-data-contrib-scaling-discussion}.

\subsection{For Scaling Up of \MLLMs: Larger Datasets}
\label{subsec-data-contrib-scaling-up}
The excellent performance of \MLLMs benefits from a larger number of parameters, especially when the size of model parameters reaches a certain level, abilities that traditional \mmodal models lack begin to emerge, such as OCR-free math reasoning \cite{mmsurvey-yin2024survey}. 
Larger models require larger-scale data \cite{kaplan2020scaling,hoffmann2022training}, especially for \MLLMs which extend the feature space of inputs and/or outputs over \LLMs. 
This subsection summarizes works scaling up \MLLMs by providing large-scale data, focusing on data acquisition (\S\ref{subsubsec-data-contrib-scaling-up-acquisition}), data augmentation (\S\ref{subsubsec-data-contrib-scaling-up-augmentation}) and data diversity (\S\ref{subsubsec-data-contrib-scaling-up-diversity}).

\subsubsection{Data Acquisition}
\label{subsubsec-data-contrib-scaling-up-acquisition}
Data acquisition, a.k.a. data collection, acquires raw data to support large-scale datasets. 
Existing works adopt various data sources, including: 
crawling from the web \cite{udandarao2024zeroshot,radford2021learning,gadre2023datacomp,liu2024chartthinker,chen2023sharegpt4v,wu2024uiclip},
exploiting existing datasets \cite{jia2024gpt4mts,wu2023large,deshmukh2022audio,liu2024chartthinker,yan2024chartreformer,yan2024list,tang2024textsquare,zou2024implicitave,yu2024texthawk,zhao2023bubogpt,lu2022learn,liu2023visual,li2024hunyuan},
employing human efforts \cite{yan2024chartreformer,maaz2023video,xia2023structchart,li2024deep,wu2024uiclip,wu2023nextgpt},
synthetic with well-trained \MLLMs such as \texttt{GPT-4V} \cite{jia2024gpt4mts,wu2024multimodal,maaz2023video,liu2023mmc,li2024deep,yu2024texthawk,zhao2023bubogpt,chen2023sharegpt4v,mu2023embodiedgpt,chen2024allava,wu2023nextgpt}, 
or 
simulators \cite{sreeram2024probing}.
The components of \MLLMs may be trained in different stages.
Taking \texttt{LLaVA} \cite{liu2023visual} as an example \footnote{The terms of pretraining and fine-tuning originate from \cite{liu2023visual}.}, these stages include: 
1) pretraining of encoders and/or generators for basic understanding and/or generation, 
2) pretraining of projector(s) for feature space alignment,  
and
3) fine-tuning of the entire or parts of the \MLLM to promote task responsiveness or downstream-task performance.
Generally, these stages consume different types of datasets \cite{gao2023llama,chen2024allava}

\para{For Encoders and Decoders}
Pretraining of encoders typically relies on massive any-text pairs which are easy to obtain. 
For example, to train the visual encoder in an \MLLM, image-text pairs can be collected from the web \cite{radford2021learning}, merging existing datasets \cite{jia2021scaling} or even purchasing \cite{li2024hunyuan}.
Similarly, audio-text pairs can also be collected from websites \cite{wu2023large}.
The significance of scaling up datasets for vision-language pretraining has been experimentally demonstrated by \cite{jia2021scaling}.

\para{For Projectors}
Pretraining of the projectors for feature alignment usually consumes the instruction data converted from existing any-text pairs. 
Existing works obtain such data by converting any-text pairs into instruction styles such as conversations with native captions as responses \cite{liu2023visual}, providing fine-grained information such as background elements and notable features \cite{chen2024allava}, or enriching the captions with contextual details such as physical appearance \cite{maaz2023video}. 

\para{For Fine-Tuning}
Fine-tuning usually consumes carefully designed instructions to further enhance \MLLMs' general instruction-following capabilities or downstream-task performance. 
For instruction following, data are typically organized in the form of QA pairs by simulating conversations between humans and assistants, using \LLMs such as \texttt{GPT-4} \cite{liu2023visual} or \MLLMs such as \texttt{GPT-4V} \cite{chen2024allava,tang2024textsquare,yan2024list}. 
The instructions could be improved with more advanced tasks such as listing all the items marked in the images one by one \cite{yan2024list}.
Compared to instruction data for projectors, these data typically contain more elements of reasoning and contextual understanding.
\MLLMs' downstream-task performance can be improved by fine-tuning with instruction data targeted at specific downstream tasks such as audio retrieval \cite{deshmukh2022audio}, chart understanding \cite{xia2023structchart,yan2024chartreformer,liu2023mmc}, user interface (UI) assessment \cite{wu2024uiclip}, and 
high-resolution visual understanding \cite{yu2024texthawk}.

\subsubsection{Data Augmentation}
\label{subsubsec-data-contrib-scaling-up-augmentation}
Data augmentation expands and diversifies a dataset by transforming existing data or adding synthetic data, thereby enhancing model generalization.

\para{Traditional Random Augmentation}
Data augmentation is widely applied to improve the generalization of deep learning models, such as random cropping, flipping, scaling and color transformation for vision tasks, and random deletion and swapping for text tasks.
\MLLMs also benefit from these simple yet effective techniques, e.g., random cropping and flipping for vision encoders \cite{ye2024mplug} and projectors \cite{li2023blip,ye2023mplug}, and character, word and sentence-level text augmentation for visual instruction tuning \cite{chen2024visual}.
For data involving temporal information such as audio and video, time masking and random scaling could be adopted \cite{vallaeys2024improved}.

\para{Generative Augmentation}
Random augmentation can disrupt the semantics in \mmodal datasets. 
Well-trained \LLMs such as \texttt{LLaMA} \cite{touvron2023llama} can be employed to rewrite text descriptions \cite{fan2023improving}, enhancing the diversity of vocabularies and sentence structures while preserving the semantics.
Such a character has benefited text-based person retrieval \cite{li2024data}, human-action description \cite{chivereanu2024aligning} and image captions \cite{yu2024capsfusion}.  
In some datasets with two (implicit) classes, different types of data may be imbalanced, causing \MLLMs to overfit to one class of data.
By synthesizing scarce negative samples in a dataset that teaches \MLLM to use tools, \MLLMs can be taught to determine whether a tool is truly needed \cite{yang2023gpt4tools} instead of always outputting that tools are needed.
Similarly, issues with insufficient positive samples in image retrieval datasets can be addressed through caption generation and template-based text modification \cite{feng2024improving}.

\subsubsection{Data Diversity}
\label{subsubsec-data-contrib-scaling-up-diversity}
Diversified data help \MLLMs perform well across different situations while alleviating the bias in data distribution.

\para{For Single-Modality Perception Abilities}
Encoders for other modalities in \MLLMs benefit from diversified data sources, thereby improving the perception ability of \MLLMs for individual modalities.
It is found that \texttt{Flamingo} \cite{alayrac2022flamingo} trained with a mixture of complementary large-scale \mmodal datasets performs significantly better than the models trained on any single dataset.
Diverse datasets have a positive impact on the understanding of both audio \cite{deshmukh2022audio} and images \cite{gadre2023datacomp}.
The diversity of concepts in the dataset is also positively correlated with the accuracy of models \cite{udandarao2024zeroshot}.

\para{For Cross-Modality Cognition Abilities}
Higher task diversity \cite{gao2024sphinx} and data-pool diversity \cite{tong2024cambrian,wu2024multimodal} improve the cross-modality cognition of \MLLMs. 
Variations in data diversity affect \MLLMs' performance in processing information from different modalities \cite{driess2023palm}.
The diversity of demonstration examples can also improve the effectiveness of in-context learning (ICL) for \MLLM reasoning \cite{liu2023retrieval}.

\subsection{\makebox{For Scaling Effectiveness of \MLLMs: Better Subsets}}
\label{subsec-data-contrib-scaling-effectiveness}
Simply increasing the amount of data is not economical \cite{survey-jin2024efficient}. 
It is indicated that only exponential data growth can lead to linear performance improvements of models \cite{udandarao2024zeroshot}. 
In addition to the computational expense, low-quality data samples can jeopardize the performance of \MLLMs.
A well-designed data filtering/selection strategy can lead to better-performing models with less token count during training \cite{gadre2023datacomp,chen2024datajuicer,zhou2023lima}.
This section refers to the training cost and performance as the \textit{scale effectiveness} of \mmodal data. 
It aims to select and orchestrate the training data to achieve higher token efficiency or better results, including: 
data condensation (\S\ref{subsubsec-data-contrib-scaling-effectiveness-condensation}), 
data mixture (\S\ref{subsubsec-data-contrib-scaling-effectiveness-mixture}), 
data packing (\S\ref{subsubsec-data-contrib-scaling-effectiveness-packing})
and 
cross-modal alignment (\S\ref{subsubsec-data-contrib-scaling-effectiveness-crossmodal-alignment}).

\subsubsection{Data Condensation}
\label{subsubsec-data-contrib-scaling-effectiveness-condensation}
Training \MLLMs with a high-quality subset may lead to comparable or even superior performance while improving the token efficiency of training \cite{chen2024datajuicer}.
It is theoretically analyzed in active learning contexts that the training with a subset containing $n$ samples could outperform training on the original dataset with $N$ samples ($N > n$) \cite{kolossov2024towards}, since the information may be transmitted from the selection strategy to \MLLMs. 
Pruning redundant data can improve the scaling trend on vision datasets \cite{sorscher2022beyond}.
We classify the relevant methods into three categories: \emph{data deduplication}, \emph{low-quality data filtering} and \emph{kernel set construction}, where the first category mainly focuses on training efficiency and the latter two additionally emphasize \MLLM performance.

\para{Data Deduplication}
As the scale of datasets continues to grow, the likelihood of having duplicate samples within the dataset increases. 
When text-to-image \MLLMs are employed for dataset construction, training samples may be directly copied \cite{webster2023duplication}.
Duplicated data not only wastes valuable computational resources but also raises issues such as concerns on copyright infringement \cite{webster2023duplication}. 
Traditional deduplication based on hash value can precisely remove the redundant samples, but may suffer the scaling of datasets \cite{jafari2021survey}.
To perform deduplication with query efficiency, the criteria for determining duplicates can be appropriately relaxed \cite{beaumont-2022-clip-retrieval,theis2022lossy}.
Thus, embeddings by models can be used to identify and filter out data pairs that are semantically similar but not identical in data representation \cite{webster2023duplication,abbas2023semdedup}.

\para{Low-Quality Data Filtering}
Data filtering, a.k.a data pruning, eliminates data samples that do not meet certain criteria. 
It contributes comparable or even better performance than training on the complete data pool while lowering training costs. 
The assessment of data quality can be achieved with heuristic metrics or evaluation models \cite{huang2024multimodal}.

\emph{Heuristic-based Filtering} is easy to implement and often effective, such as discarding samples with over-length captions \cite{gadre2023datacomp} or low text complexity \cite{radenovic2023filtering}. 
Several works focus on designing data filters with fixed training code and computational budget \cite{gadre2023datacomp,mahmoud2023sieve,huang2024multimodal,maini2023t}.
To avoid bias by a single filter, an ensemble of filters could be employed \cite{huang2024multimodal}.
Despite being effective, heuristic-based filtering eliminates certain categories of data, lowering data diversity \cite{nguyen2023improving}.

\emph{Model-based Filtering} eliminates the reliance on human labor of heuristic-based filtering. 
The evaluation model can be a pretrained foundation model such as \texttt{CLIP} for image-text matching \cite{yu2023devil,gadre2023datacomp,mahmoud2023sieve}, or a finetuned \MLLM to score the quality of instructions \cite{wang2024finetuned}.
The neural network for data filtering has also been exploited \cite{fang2023data}.
Apart from discarding low-quality data samples, some works salvage the wrongly discarded data samples by rewriting the captions with well-trained models \cite{mahmoud2023sieve,nguyen2023improving}, since some of the data pairs may be still valuable with appropriate rewriting. 

\para{Kernel Set Construction}
Contrary to low-quality data filtering, kernel set construction starts from scratch and progressively adds data samples from a candidate pool to the selected training set. 
It is typically achieved by clustering the training data and selecting representative samples (e.g., those close to the centroid) from each cluster \cite{he2024efficient,wei2023instructiongpt}. 
Experiments indicate that fewer but higher-quality instruction data can enable \MLLMs to generate better outputs.

\subsubsection{Data Mixture}
\label{subsubsec-data-contrib-scaling-effectiveness-mixture}
A dataset can be a mixture of data from multiple domains and even various sources.
Different from the approaches focusing on data diversity (\S\ref{subsubsec-data-contrib-scaling-up-diversity}), those focusing on data mixture primarily enhance \MLLM performance by adjusting the proportion of different categories of data within the given training dataset \cite{yan2024list}, which is similar to adjusting weights of tasks in multi-task learning.

\para{Mitigating Distribution Bias}
Generally, a balanced mixture across different domains leads to better pretraining performance \cite{xu2024demystifying}. 
The ratios of synthetic and original data also have a certain impact \cite{tsai2024text}.
It is indicated that the mixed pretraining dataset should include more samples from higher quality sources, rather than randomly sampling from all candidate sources \cite{nguyen2022quality}.
Besides, the strategy of data mixing and the arrangements of model parameter updates also have a significant impact on pertaining \cite{alayrac2022flamingo}.

\para{Exploiting Distribution Bias}
Although eliminating bias between datasets can help improve model performance, in some cases, biased data distribution can be leveraged beneficially \cite{liu2024decade}.
During fine-tuning, adjusting the proportion of different categories of data in the dataset can meet varying requirements on image generation \cite{li2024hunyuan}.

\subsubsection{Data Packing}
\label{subsubsec-data-contrib-scaling-effectiveness-packing}
Data packing improves \MLLM performance through the optimal arrangement of data samples within each batch, without altering the dataset size. 
Compared to data mixture (\S\ref{subsubsec-data-contrib-scaling-effectiveness-mixture}) which combines different types of data at the dataset level, data packing combines them at the batch level. 

\para{For Better Pretraining}
Including hard negatives in each batch can improve the pretraining of \texttt{CLIP} \cite{ma2024mode}.
Packing visual patches from images of different resolutions into a single sequence allows for variable resolution while maintaining the aspect ratio \cite{dehghani2024patch}. 
It may be the solution of \texttt{Sora} to the variability in latent space dimensions \cite{liu2024sora}. 

\para{For Long-Context Support}
By strategically combining inputs of different lengths into a single batch, data packing enhances the long-context support of \MLLMs. 
Existing works exploit this by formulating an optimization problem to minimize the disruption of contextual information via arranging the chunked long documents and short documents \cite{ding2024fewer}, or by arranging mutually relevant documents in a single training sequence as much as possible, long-context utilization can be improved with only brief fine-tuning \cite{staniszewski2023structured}.
From our investigations, data packing is promising but not fully investigated in the \mmodal domain.

\subsubsection{Cross-Modal Alignment}
\label{subsubsec-data-contrib-scaling-effectiveness-crossmodal-alignment}
\MLLMs benefit from the alignment between different modalities of data. 
However, ensuring precise alignment is a labor-intensive task, potentially leading to mismatches \cite{han2023noisy}.
This section summarizes existing works to detect and improve the inner-sample cross-modal alignment.

\para{Joint Embedding Space}
By mapping different modality data into the same feature space, we can assess their matching with the similarity of the embedding vectors \cite{fuyu-8b}.
These methods are typically related to multi-task learning and contrastive learning, such as treating image-text pairs with \texttt{CLIP} scores low as cross-modal mismatches, \cite{gadre2023datacomp,huang2024multimodal,mahmoud2023sieve,maini2023t,nguyen2023improving,fang2023data} or frame-level mismatches between video and text \cite{blattmann2023stable}.
\texttt{CLIP} score helps to assign different levels of alignment capabilities to data samples \cite{zhao2024aligngpt}. 
Techniques of learning from noised data can also be used to remove data samples with potentially incorrect correspondences \cite{han2023noisy}.

\para{Text-Centric Anchoring}
The text can serve as an anchor to enable \MLLMs to boost cross-modal alignment \cite{tsai2024text} with their knowledge.
Some studies suggest that the mismatches between images and captions may not always stem from semantic dissimilarity, but rather from the low quality of the captions. 
Thus, some data samples initially deemed as mismatches can be reused by caption rewriting \cite{nguyen2023improving}. 
Similar approaches also facilitate the construction of \texttt{BLIP-2} \cite{li2023blip} and \texttt{ALLaVA} \cite{chen2024allava}.
Chain-of-thought (CoT) techniques can also improve the text-centric alignment of charts~\cite{liu2024chartthinker}.

\subsection{Brief Summarization and Discussion}
\label{subsec-data-contrib-scaling-discussion}
A high-performing large-scale \MLLM requires extensive datasets.
\MLLMs typically employ a modular training paradigm, where different components at different stages may consume different types of datasets (\S\ref{subsubsec-data-contrib-scaling-up-acquisition}). 
After datasets are constructed, data augmentation may be needed (\S\ref{subsubsec-data-contrib-scaling-up-augmentation}), which generally employs different strategies or different models for data in different modalities.
In both acquisition and augmentation, enhancing the diversity generally benefits \MLLMs (\S\ref{subsubsec-data-contrib-scaling-up-diversity}).
However, optimizing diversity may bias the dataset towards simpler samples, while difficulty-based selection can miss necessary simple data \cite{maharana2023d2}, highlighting the trade-offs between data diversity and sample importance.
Besides, predictive determining the mixture ratio of various data sources has been studied in \LLMs \cite{bimix}. 
However, the applicability and efficacy of such approaches for \MLLMs remain uncharted territory.

After scaling up datasets, we need to improve the scaling effectiveness.
To save computational resources and eliminate the low-quality data samples, we need to condense the dataset (\S\ref{subsubsec-data-contrib-scaling-effectiveness-condensation}) by deduplication, filtering, and kernel set construction, obtaining a true subset with high-density information.
Then, there are two methods of adjusting the proportions of different data for better \MLLMs:
data mixture (\S\ref{subsubsec-data-contrib-scaling-effectiveness-mixture}) which works at the dataset level for better task performance
and 
data packing (\S\ref{subsubsec-data-contrib-scaling-effectiveness-packing}) which works at the batch level to improve convergence and support long contexts.
While these methods should be orthogonal, their combined effects have not been extensively explored.
A unique challenge for \MLLMs is cross-modal alignment, which involves mapping data into a unified feature space or using text as an anchor. 
There is a lack of research on using other modalities as anchors, as discussed in Sec. \ref{subsec-future-automatic-modality-detection}.
\section{\MModal Data Contributions for \MLLMs: Usability}
\label{sec-data-contrib-usability}
\definecolor{color-layer0}{HTML}{00BFB0}
\definecolor{color-layer1}{HTML}{559D94}
\definecolor{color-layer2}{HTML}{E8FEFA}
\definecolor{color-layer3}{HTML}{F3EADA}

\tikzstyle{leaf-style}=[
    rectangle,
    draw=hidden-draw,
    rounded corners,
    text opacity=1,
    minimum height=1.5em,
    minimum width=5em,
    inner sep=2pt,
    align=center,
    fill opacity=.5,
    line width=0.9pt,
]
\tikzstyle{leaf}=[leaf-style, minimum height=1.5em,
    fill=hidden-pink!80, text=black, align=left,font=\scriptsize,
    inner xsep=2pt,
    inner ysep=2pt,
    line width=0.8pt,
]

\begin{figure*}[t]
    \centering
    \begin{forest}
        forked edges,
        for tree={
            grow=east,
            reversed=true,
            anchor=base west,
            parent anchor=east,
            child anchor=west,
            base=center,
            font=\small,
            rectangle,
            draw=hidden-draw,
            rounded corners,
            align=center,
            text centered,
            minimum width=4em,
            edge+={darkgray, line width=1pt},
            s sep=2pt,
            inner xsep=2pt,
            inner ysep=2pt,
            line width=0.8pt,
            ver/.style={rotate=90, child anchor=north, parent anchor=south, anchor=center},
        },
        where level=1{text width=15em,font=\footnotesize,}{},
        where level=2{text width=14.3em,font=\scriptsize,}{},
        where level=3{text width=14.6em,font=\scriptsize,}{},
        [
            \textbf{\MModal Data Contributions}\\\textbf{for \MLLMs: Usability}, ver, fill=color-layer0!30
            [
                \textbf{For Instruction Responsiveness}\\\textbf{of \MLLMs} (\S\ref{subsec-data-contrib-usability-following}), fill=color-layer1!30
                [
                    \textbf{Prompt Design} (\S\ref{subsubsec-data-contrib-usability-following-prompt}), fill=color-layer2!30
                    [
                        Prompts for Better Responsiveness, leaf
                    ]
                    [
                        Prompts for Dataset Curation, leaf
                    ]
                ]
                [
                    \textbf{ICL Data} (\S\ref{subsubsec-data-contrib-usability-following-icl}), fill=color-layer2!30
                    [
                        Demonstration Creation, leaf
                    ]
                    [
                        Demonstration Optimization, leaf
                    ]
                ]
                [
                    \textbf{Human-Behavior Alignment Data} (\S\ref{subsubsec-data-contrib-usability-following-human}), fill=color-layer2!30
                    [
                        Human Preference-Oriented Improvement, leaf
                    ]
                    [
                        Hallucination Reduction, leaf
                    ]
                ]
            ]
            [
                \textbf{For Reasoning Ability}\\\textbf{of MLLMs} (\S\ref{subsec-data-contrib-usability-reasoning}), fill=color-layer1!30
                [
                    \textbf{Data for Single-Hop Reasoning} (\S\ref{subsubsec-data-contrib-usability-reasoning-single-hop}), fill=color-layer2!30
                ]
                [
                    \textbf{Data for Multi-Hop Reasoning} (\S\ref{subsubsec-data-contrib-usability-reasoning-multi-hop}), fill=color-layer2!30
                ]
            ]
            [
                \textbf{For Ethics of MLLMs} (\S\ref{subsec-data-contrib-usability-ethic}), fill=color-layer1!30
                [
                    \textbf{Data Toxicity} (\S\ref{subsubsec-data-contrib-usability-ethic-toxicity}), fill=color-layer2!30
                    [
                        Data-Centric Attacks, leaf
                    ]
                    [
                        Data-Centric Defenses, leaf
                    ]
                ]
                [
                    \textbf{Data Privacy and Intellectual} \\ \textbf{Property} (\S\ref{subsubsec-data-contrib-usability-ethic-privacy}), fill=color-layer2!30
                    [
                        Privacy Attacks, leaf
                    ]
                    [
                        Privacy-Preserving Training, leaf
                    ]
                    [
                        License Analysis, leaf
                    ]
                    [
                        Model Watermarking, leaf
                    ]
                ]
            ]
            [
                \textbf{For Evaluation of \MLLMs} (\S\ref{subsec-data-contrib-usability-benchmark}), fill=color-layer1!30
                [
                    \textbf{Benchmarks for \MModal} \\ \textbf{Understanding} (\S\ref{subsubsec-data-contrib-usability-benchmark-understanding}), fill=color-layer2!30
                    [
                        Visual Understanding, leaf
                    ]
                    [
                        Spatial Understanding, leaf
                    ]
                    [
                        Temporal Understanding, leaf
                    ]
                ]
                [
                    \textbf{Benchmarks for \MModal} \\ \textbf{Generation} (\S\ref{subsubsec-data-contrib-usability-benchmark-generation}), fill=color-layer2!30
                    [
                        Quality of Generation, leaf
                    ]
                    [
                        Human-Preference Alignment of Generation, leaf
                    ]
                    [
                        Safety of Generation, leaf
                    ]
                ]
                [
                    \textbf{Benchmarks for \MModal} \\ \textbf{Retrieval} (\S\ref{subsubsec-data-contrib-usability-benchmark-retrieval}), fill=color-layer2!30
                ]
                [
                    \textbf{Benchmarks for \MModal} \\ \textbf{Reasoning} (\S\ref{subsubsec-data-contrib-usability-benchmark-reasoning}), fill=color-layer2!30
                ]
            ]
        ]
    \end{forest}
    \caption{Organization of data approaches tailored for the usability of \MLLMs.}
    \label{pic-data4model-usability}
\end{figure*}
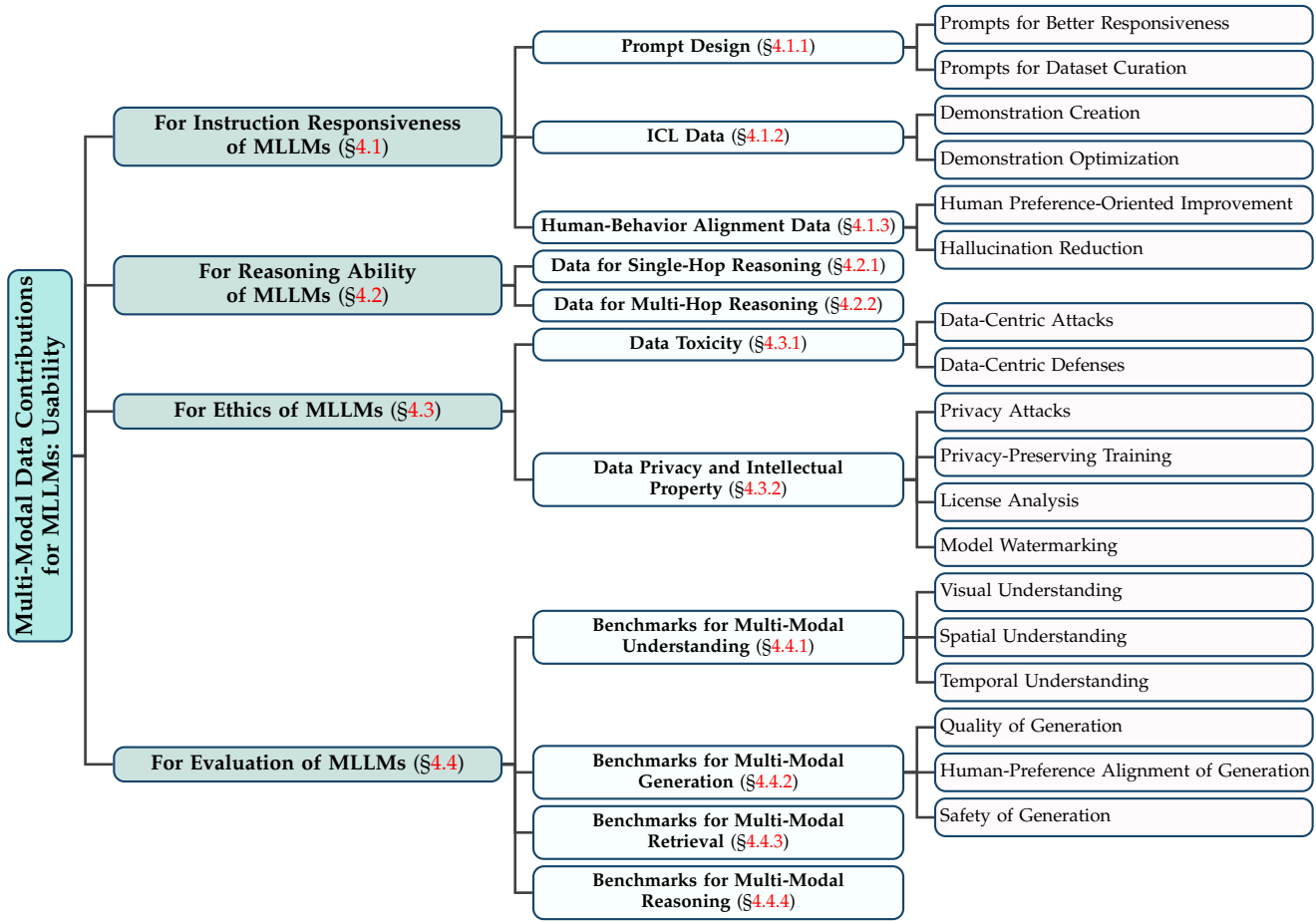
After obtaining an \MLLM with the works introduced in Sec. \ref{sec-data-contrib-scaling}, enhancement of its usability is needed.
We organize related works revolving around data into \textit{three progressive phases, followed by a retrospective}, as shown in Fig. \ref{pic-data4model-usability}. 
An \MLLM needs to go through the improvements on responsiveness to human instructions (\S\ref{subsec-data-contrib-usability-following}), reasoning capabilities for better intelligence (\S\ref{subsec-data-contrib-usability-reasoning}) and compliance with ethics (\S\ref{subsec-data-contrib-usability-ethic}). 
Finally, comprehensive evaluations are needed to review the effectiveness of previous methods and assist in model selection for application.
After reviewing existing works, we briefly summarize and discuss them in Sec. \ref{subsec-data-contrib-usability-discussion}.

\subsection{For Instruction Responsiveness of \MLLMs}
\label{subsec-data-contrib-usability-following}
The instruction responsiveness of \MLLMs can be improved through better prompt templates (\S\ref{subsubsec-data-contrib-usability-following-prompt}), ICL demonstrations (\S\ref{subsubsec-data-contrib-usability-following-icl}), or fine-tuning data aligned with human behaviors (\S\ref{subsubsec-data-contrib-usability-following-human}), where generally, the first two methods do not require changes to the model parameters. 

\subsubsection{Prompt Design}
\label{subsubsec-data-contrib-usability-following-prompt}
The prompt design optimizes the input context $\mathbf{c}$ of \MLLMs in Definition \ref{def-co-development}, helping \MLLMs to focus on specific tasks or objectives such as questions, instructions, or backgrounds.

\para{Prompts for Better Responsiveness}
With carefully designed templates, \MLLMs can be better guided to respond to various tasks.
The set-of-mark prompt indicates that simply overlaying identities on image regions can release the visual grounding capability \cite{yang2023set} and improve the responses of \texttt{GPT-4V} by enabling better identification of the objects referred to by visual prompts \cite{lin2024drawandunderstand}.
With special prompts, \MLLMs can detect whether their outputs contain harmful content \cite{gou2024eyes}.
Users can specify visual regions to \MLLMs with specially-designed templates \cite{chen2023shikra,peng2023kosmos}.
By packing the transformed time series patches with specific prompts such as statistics, time series could be better predicted \cite{jin2023time}. 

\para{Prompts for Dataset Curation}
Given specially designed prompts, \MLLMs such as \texttt{GPT-4V} can generate responses including detailed evidence and reasoning, enhancing the response quality in curated instruction datasets \cite{chen2024allava}. 
As indicated by \cite{fan2024scaling}, the diversity of synthetic data is positively correlated with the diversity of prompts. 
Thus, it is important to provide a diverse set of prompt templates.

\subsubsection{ICL Data}
\label{subsubsec-data-contrib-usability-following-icl}
\MLLMs can learn from ICL demonstrations to improve their outputs \cite{sun2024generative}.
Thus, the demonstrations play a crucial role in enhancing the responsiveness of \MLLMs.

\para{Demonstration Creation}
ICL demonstrations generally have less quantity compared to training data.
Existing works involve manually constructing high-quality ICL demonstrations to directly boost \MLLM performance \cite{li2024groundinggpt} and teach well-trained \MLLMs to generate datasets \cite{yan2024list}.
Some studies suggest that providing ICL demonstrations alone may not be sufficient for \MLLM capabilities, and fine-tuning with ICL datasets might be beneficial~\cite{zhao2024mmicl,doveh2024towards}.

\para{Demonstration Optimization}
One study finds that non-text information in ICL demonstrations (e.g., images) typically consumes a large number of tokens which leads to unsatisfactory ICL performance, thus, images in demonstrations are aggregated to the latent space of textual labels to shorten input length \cite{gao2024aim}. 
For a similar purpose, some works aggregate the demonstrations into images to avoid inaccurate text descriptions of complex images \cite{wang2024aggregatedimageinimagelearning}.
One study finds that demonstrations from diverse groups (higher diversity) may lead to better performance \cite{liu2023retrieval}.

\subsubsection{Human-Behavior Alignment Data}
\label{subsubsec-data-contrib-usability-following-human}
Human-behavior alignment for \MLLMs aims at making the behavior and decision-making of \MLLMs more consistent with human behaviors, aligning the \MLLM's output with human preferences and reducing hallucinations \cite{bai2024hallucination}.

\para{Human Preference-Oriented Improvement}
\MLLM outputs can better fit human preference with the help of curated human feedback datasets \cite{yin2023lamm}.
Due to the high cost of manually constructing such datasets, human efforts can often be replaced by well-trained \MLLMs such as \texttt{GPT-4V} since these well-trained \MLLMs are well-aligned with human preferences \cite{ge2024mllmbench,wu2024multimodal}. 
With the feedback, human preferences can be distilled to \MLLMs with reinforcement learning (RL) \cite{chen2024dress,wu2024multimodal} or direct preference optimization (DPO) \cite{Li2023SilkiePD,yu2024rlhf}.
Preferences can also be presented in a multi-level form, making \MLLMs better align with human preferences by learning from the differences between adjacent levels of preferences \cite{Zhang2024AutomatedMP}.
Note that merely pursuing high reward scores does not necessarily increase human preference for \MLLMs, a.k.a. reward hacking, which can be calibrated with additional information \cite{sun2023aligning}.

\para{Hallucination Reduction}
Humans answer questions based on facts and background knowledge rather than fabricating information, while \MLLMs may exhibit hallucinations \cite{bai2024hallucination,huang2024opera}.
Hallucinations can be mitigated through fine-tuning with instruction datasets \cite{liu2023mitigating,chen2024allava} and DPO with feedback datasets \cite{Zhang2024AutomatedMP} curated with \texttt{GPT-4V}. 
Some works promote the study of hallucinations from a benchmarking perspective, which we discuss in detail in Sec. \ref{subsec-data-contrib-usability-benchmark}.

\subsection{For Reasoning Ability of \MLLMs}
\label{subsec-data-contrib-usability-reasoning}
This section summarizes existing efforts on the reasoning abilities of \MLLMs with data as an intermediary, organized according to the number of thinking steps required.

\subsubsection{Data for Single-Hop Reasoning}
\label{subsubsec-data-contrib-usability-reasoning-single-hop}
Single-hop reasoning requires a direct relationship between the question and the answer. 
Some works enhance this capability of \MLLMs by curating corresponding datasets, emphasizing chart reasoning including mathematical reasoning and chart redrawing \cite{xia2023structchart}, and the understanding of humorous and creative content in videos \cite{xie2023funqa}.

\subsubsection{Data for Multi-Hop Reasoning}
\label{subsubsec-data-contrib-usability-reasoning-multi-hop}
\Mmodal multi-hop reasoning needs multiple reasoning steps to answer a question with information from multiple modalities.
Annotated reasoning steps in data help teach \MLLMs to better demonstrate the decision-making process  \cite{gai2024medthink}.
Providing relevant demonstrations can enhance the \mmodal CoT \cite{liu2023retrieval,zheng2023ddcot}, which benefits robotic reasoning \cite{mu2023embodiedgpt} and scientific question answering \cite{lu2022learn}.
Scaling up the corresponding data can significantly enhance the reasoning capabilities of \MLLMs in both 2D and 3D spaces \cite{cho2024language} as well as temporal localization in videos \cite{chen2023grounding}. 
Augmenting language models with vision experts and compositional reasoning modules leads to better reasoning performance on complex cross-modal tasks \cite{yang2023mm,lu2024chameleon}.
Additionally, neuro-symbolic approaches help in improving reasoning performance on complex visual tasks without training \cite{gupta2023visual}.

\subsection{For Ethics of \MLLMs}
\label{subsec-data-contrib-usability-ethic}
Ethical considerations for \MLLMs are crucial for their usability. 
It is essential to ensure the generated contents do not contain harmful content (\S\ref{subsubsec-data-contrib-usability-ethic-toxicity}), while guaranteeing that the training and inference do not violate privacy or infringe on intellectual property rights (\S\ref{subsubsec-data-contrib-usability-ethic-privacy}).
These issues can be addressed and resolved from data-centric aspects.

\subsubsection{Data Toxicity}
\label{subsubsec-data-contrib-usability-ethic-toxicity}
The integration of other modalities makes \MLLMs more vulnerable to attackers \cite{fan2024unbridled}.
When presented alongside input images with malicious content, \MLLMs are more prone to generating sensitive or harmful responses than facing purely text-based inputs \cite{pi2024mllmprotector,gou2024eyes}.
To protect \MLLMs from ethical violations and thereby improve their usability, it is necessary to research data toxicity faced by \MLLMs. 

\para{Data-Centric Attacks}
\MLLMs can be poisoned with training data rather than directly manipulating their parameters.
Some works construct adversarial examples to disrupt foundation models \cite{wang2023exploring,lu2023set,schlarmann2023adversarial,he2023sa}, but may not always be effective in the context of \MLLMs, since many \MLLMs are built with frozen encoders \cite{liu2023visual}.
Some works build malicious data to fine-tune \MLLMs to generate harmful responses, with images \cite{tao2024imgtrojan} or audio as carriers \cite{bagdasaryan2023ab}.
Such jailbreak attacks can be performed with prompt engineering \cite{wu2023jailbreaking}. 
Adversarial prompts can even be transmitted from malicious agents to benign ones to induce harmful outcomes in a multi-agent system \cite{tan2024wolf}.
Image triggers can be planted for backdoor attacks \cite{liang2024vl}.

\para{Data-Centric Defenses}
\MLLMs can counteract harmful queries by simply adding a simple safety prompt before the queries \cite{liu2024mmsafetybench}. 
Observing that \MLLMs can retain the safeguard when images are removed from the inputs, prompting \MLLMs to first detect and then remove or convert the malicious input images helps to defend against jailbreak attacks \cite{gou2024eyes}.  
Additionally, training to distinguish the authenticity with \mmodal information helps \MLLMs to avoid producing incorrect and confusing outputs \cite{sun2023med}.

\subsubsection{Data Privacy and Intellectual Property}
\label{subsubsec-data-contrib-usability-ethic-privacy}
Privacy and intellectual property are crucial for \MLLM usability. 
Dataset providers must address privacy issues during data collection, and \MLLM training must filter and remove private information. 
This ensures compliance with laws and encourages data contributions. 
For intellectual property, model and data releases typically include licenses that protect innovations and economic interests, providing legal safeguards and market competitiveness. 
Analyzing these licenses helps avoid intellectual property disputes.

\para{Privacy Attacks}
\MLLMs can memorize training data during training and may reveal portions of personal information under specific prompts similar to \LLMs \cite{lukas2023analyzing}.
Thus, training data must exclude personal information while retaining public information, which is evaluated by a benchmark as a red team \cite{li2024red}. 
Some works exploit this memory information to extract private data, such as using the alignment information between street images and resident details to identify residents in geospatial systems~\cite{rao2023building}.

\para{Privacy-Preserving Training}
\MLLMs can be fine-tuned to decline requests for private information \cite{achiam2023gpt}. 
Image synthesis can promote privacy protection while performing data augmentation \cite{joshi2024Human-Analysis}.
Differential privacy (DP) \cite{huang2023safeguarding} can be viewed as a formal transformation of data, i.e., optimizing the data to retain as much utility as possible while removing private information.
Federated learning (FL) \cite{mcmahan2017communication} can harness edge-side data in a privacy-preserving manner to expand the data scale while protecting privacy, which has demonstrated effectiveness in training \LLMs by combining zero-order optimization \cite{on-device-llm,zero-order-llm} and parameter-efficient fine-tuning (PEFT) \cite{bai2024federated,zero-order-llm} to reduce memory and communication overhead.
Although popular for \LLMs, it has not yet been comprehensively explored for \MLLMs.

\para{License Analysis}
Models and data often come with licenses restricting their use, which are a special type of data tied to the lifecycle of \MLLMs and require comprehensive analysis, as current remedies are post hoc such as dataset retractions and modifications \cite{andrews2023ethical}.
Traditional analysis for open-source software \cite{ombredanne2020free,german2010sentence} may be inadequate as \MLLM projects are composed of several components including software, data and models.
A tool named \texttt{ModelGo} is proposed to assess potential legal risks in machine learning projects \cite{duan2024modelgo}, providing several case studies covering five modalities. 
This topic is important but still in its early stages.

\para{Model Watermarking}
By embedding benign backdoors in the training data, i.e., watermarks, model-service providers can determine whether the outputs are generated with their model, thereby strengthening user compliance with the licenses published by the model-service providers, such as clarifying the adoption of specific embedding models \cite{tang2023watermarking}. 
As more and more \mmodal data are created with \MLLMs, the application of watermarking techniques for \MLLMs needs further exploration.
\subsection{For Evaluation of \MLLMs}
\label{subsec-data-contrib-usability-benchmark}
The evaluation of \MLLMs helps identify the improvements brought by specific data or approaches for further optimization, thereby facilitating the usability of \MLLMs.
This subsection summarizes existing works that curate datasets to benchmark \MLLMs' capabilities in \mmodal contexts, including:  
understanding (\S\ref{subsubsec-data-contrib-usability-benchmark-understanding}), 
generation (\S\ref{subsubsec-data-contrib-usability-benchmark-generation}), 
retrieval (\S\ref{subsubsec-data-contrib-usability-benchmark-retrieval})
and 
reasoning (\S\ref{subsubsec-data-contrib-usability-benchmark-reasoning}).

\subsubsection{Benchmarks for \MModal Understanding}
\label{subsubsec-data-contrib-usability-benchmark-understanding}
With more comprehensive and accurate \mmodal understanding capabilities, \MLLMs can better process and integrate information from different data modalities.

\para{Visual Understanding}
Visual understanding of \MLLMs can be evaluated with general visual tasks such as image understanding \cite{gadre2023datacomp} and visual perception \cite{fu2024blink}, as well as specific downstream scenarios such as chart understanding \cite{chen2024onechart}, chart reasoning \cite{liu2023mmc}, aesthetic assessment \cite{zhou2024uniaa}, visual tasks in intelligent transportation \cite{shi2023open} and identifying the attributes from the product descriptions \cite{zou2024implicitave}.

\para{Spatial Understanding}
3D understanding requires comprehension of spatial information, depth perception, and volume measurement over 2D understanding. 
Existing related evaluations cover point-cloud spatial understanding \cite{zhang20243dbench}, region-level and scene-level tasks \cite{li2023m3dbench}, as well as instruction-following abilities in 3D point clouds \cite{yin2023lamm}.

\para{Temporal Understanding}
Temporal understanding of \MLLMs often refers to the ability to handle tasks difficult to solve within a single frame \cite{li2023mvbench}. 
According to our survey, benchmarks for temporal understanding are usually accompanied by those for spatial understanding \cite{li2023seed,li2023mvbench}.

\subsubsection{Benchmarks for \MModal Generation}
\label{subsubsec-data-contrib-usability-benchmark-generation}
Generation tasks have an extraordinarily vast output space, making the evaluation quite challenging.
Existing works assess the generation of \MLLMs from various perspectives.

\para{Quality of Generation}
Some existing works on the evaluation of \MLLM generation cover interleaved image-text generation \cite{an2023openleaf} and video generation \cite{huang2023vbench,liu2023evalcrafter,ge2024worldgpt}. 
Current evaluations mainly emphasize subject and background consistency in temporal quality, as well as aesthetic quality and motion smoothness \cite{huang2023vbench}, motion quality \cite{liu2023evalcrafter}, and consistency with action sequence descriptions \cite{ge2024worldgpt}.

\para{Human-Preference Alignment of Generation}
Some works focus on evaluating the alignment of \MLLMs to human preference \cite{chen2024mllm,huang2023vbench,ge2024mllmbench}. 
Evaluations can be conducted in a unified standard with scoring and pair and batch ranking \cite{chen2024mllm}, or customized evaluation criteria can be applied to each evaluation sample \cite{ge2024mllmbench}.
Considering that \MLLMs often experience hallucinations, which result in outputs that are inconsistent with human expectations, some works tailor evaluation for hallucinations \cite{huang2024visual,chen2024unified}.

\para{Safety of Generation}
Although the \LLMs in \MLLMs are often equipped with abilities to avoid harmful outputs, \MLLMs can be affected by jailbreak attacks \cite{zhao2024survey}. 
Existing work constructs evaluation datasets encompassing different scenarios \cite{liu2024mmsafetybench} and harmful behaviors \cite{niu2024jailbreaking} to comprehensively assess the impact of jailbreak attacks on \MLLMs.

\subsubsection{Benchmarks for \MModal Retrieval}
\label{subsubsec-data-contrib-usability-benchmark-retrieval}
The retrieval capabilities of \MLLMs help break down barriers between different forms of information, facilitating better search and recommendation functions.
Existing benchmarks evaluate such capabilities for audio retrieval \cite{deshmukh2022audio} and in intelligent transportation scenarios which focus on cross-modality image retrieval \cite{shi2023open},  and vehicle retrieval, person re-identification and sketch-to-image \cite{ying2024mmt}.

\subsubsection{Benchmarks for \MModal Reasoning}
\label{subsubsec-data-contrib-usability-benchmark-reasoning}
Some benchmarks evaluate reasoning abilities of \MLLMs that do not require CoT, covering both images \cite{xia2024chartx,masry2022chartqa,zhang2024if} and videos \cite{xie2023funqa,sreeram2024probing} from a modality perspective, and include counter-intuitive and surprising information in videos \cite{xie2023funqa}, decision-making in autonomous driving \cite{sreeram2024probing}, and operations related to charts from an application scenario perspective \cite{xia2024chartx,masry2022chartqa}.
There are also benchmarks focusing on the \mmodal CoT tasks for \MLLMs \cite{chen2024m}.
\subsection{Brief Summarization and Discussion}
\label{subsec-data-contrib-usability-discussion}
Usually, we need to construct and optimize data to further improve the usability of \MLLMs trained with the works presented in Sec. \ref{sec-data-contrib-scaling} to enhance user satisfaction and protect the interests of involved parties, paralleled to evaluations to review the strengths and weaknesses of \MLLMs.

\MLLMs' instruction responsiveness (\S\ref{subsec-data-contrib-usability-following}) can be improved with better prompt templates (\S\ref{subsubsec-data-contrib-usability-following-prompt}), high-quality ICL demonstrations (\S\ref{subsubsec-data-contrib-usability-following-icl}), and fine-tuning aligned with human behavior (\S\ref{subsubsec-data-contrib-usability-following-human}). 
While the first two methods may not need to update model parameters, they could consume more tokens for inference.
A study on ICL shows that text is the main information source in \mmodal contexts, while images have little impact, indicating to increase in the emphasis on other modalities in ICL demonstrations \cite{baldassini2024makes}.

To improve the reasoning ability of \MLLMs (\S\ref{subsec-data-contrib-usability-reasoning}), corresponding datasets could be constructed for single-hop (\S\ref{subsubsec-data-contrib-usability-reasoning-single-hop}) and multi-hop reasoning (\S\ref{subsubsec-data-contrib-usability-reasoning-multi-hop}). 
However, most current reasoning tasks for \MLLMs focus on scenarios with intertwined visual and textual elements. 
There is a need to explore reasoning applications that deeply integrate information from other modalities to improve \MLLMs.

To address ethical concerns (\S\ref{subsec-data-contrib-usability-ethic}), it is important to prevent \MLLMs from toxic outputs (\S\ref{subsubsec-data-contrib-usability-ethic-toxicity}), privacy leakage and license disputes (\S\ref{subsubsec-data-contrib-usability-ethic-privacy}). 
Generally, non-text modalities suscept \MLLMs to jailbreak attacks.
However, research suggests visual cues can improve \MLLMs' ethical alignment \cite{tu2023sight}, warranting further related explorations.
While watermarking and privacy-preserving training are well-studied for \LLMs, they have been not fully investigated for \MLLMs.

To objectively compare \MLLMs and reveal their task-specific shortcomings, benchmarks are needed (\S\ref{subsec-data-contrib-usability-benchmark}) on understanding (\S\ref{subsubsec-data-contrib-usability-benchmark-understanding}), generation (\S\ref{subsubsec-data-contrib-usability-benchmark-generation}), retrieval (\S\ref{subsubsec-data-contrib-usability-benchmark-retrieval}) and reasoning (\S\ref{subsubsec-data-contrib-usability-benchmark-reasoning}). 
As the covered tasks continue to expand, data-efficient evaluations \cite{xu2024data} for \MLLMs may be promising due to the conservation of computing resources.

\section{Model Contributions for \MModal Data: Synthesis}
\label{sec-model-contrib-synthesis}
As the scale of datasets for \MLLMs continues to grow, manually constructing datasets becomes a labor-intensive task. 
As a key participant in the \codev, it is essential to identify the contributions made by models to \mmodal data.
This section reviews model efforts to synthesize data in existing works, categorized by \textit{the roles the models play}, and organized in the \textit{order of a common pipeline for building a dataset}, including: 
data \emph{creator} (\S\ref{subsec-model-contrib-synthesis-creator}), 
data \emph{mapper} (\S\ref{subsec-model-contrib-synthesis-mapper}), 
data \emph{filter} (\S\ref{subsec-model-contrib-synthesis-filter}),
and 
data \emph{evaluator} (\S\ref{subsec-model-contrib-synthesis-evaluator}), as illustrated in Fig. \ref{pic-model4data-synthesis}. 
At the end of this section, a brief summarization and discussion are provided (\S\ref{subsec-model-contrib-synthesis-discussion}).

\definecolor{color-layer0}{HTML}{3498DB}
\definecolor{color-layer1}{HTML}{79FAC5}
\definecolor{color-layer2}{HTML}{3AC08F}
\definecolor{color-layer3}{HTML}{008A5C}

\tikzstyle{leaf-style}=[
    rectangle,
    draw=hidden-draw,
    rounded corners,
    text opacity=1,
    minimum height=1.5em,
    minimum width=5em,
    inner sep=2pt,
    align=center,
    fill opacity=.5,
    line width=0.9pt,
]
\tikzstyle{leaf}=[leaf-style, minimum height=1.5em,
    fill=color-layer2!15, text=black, align=left,font=\scriptsize,
    inner xsep=2pt,
    inner ysep=2pt,
    line width=0.8pt,
]

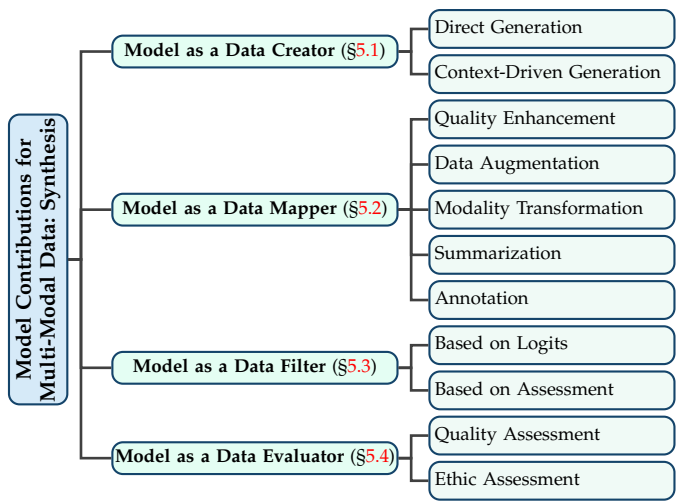
\begin{figure}[ht]
    \centering
    \begin{forest}
        forked edges,
        for tree={
            grow=east,
            reversed=true,
            anchor=base west,
            parent anchor=east,
            child anchor=west,
            base=center,
            font=\small,
            rectangle,
            draw=hidden-draw,
            rounded corners,
            align=center,
            text centered,
            minimum width=4em,
            edge+={darkgray, line width=1pt},
            s sep=2pt,
            inner xsep=1pt,
            inner ysep=2pt,
            line width=0.8pt,
            ver/.style={rotate=90, child anchor=north, parent anchor=south, anchor=center},
        },
        where level=0{font=\footnotesize,}{},
        where level=1{text width=11.1em,font=\scriptsize,}{},
        where level=2{text width=9.3em,font=\scriptsize,}{},
        [
            \textbf{Model Contributions for}\\\textbf{\MModal Data: Synthesis}, ver, fill=color-layer0!20
            [
                \textbf{Model as a Data Creator} (\S\ref{subsec-model-contrib-synthesis-creator}), fill=color-layer1!20
                [
                    Direct Generation, leaf
                ]
                [
                    Context-Driven Generation, leaf
                ]
            ]
            [
                \textbf{Model as a Data Mapper} (\S\ref{subsec-model-contrib-synthesis-mapper}), fill=color-layer1!20
                [
                    Quality Enhancement, leaf
                ]
                [
                    Data Augmentation, leaf
                ]
                [
                    Modality Transformation, leaf
                ]
                [
                    Summarization, leaf
                ]
                [
                    Annotation, leaf
                ]
            ]
            [
                \textbf{Model as a Data Filter} (\S\ref{subsec-model-contrib-synthesis-filter}), fill=color-layer1!20
                [
                    Based on Logits, leaf
                ]
                [
                    Based on Assessment, leaf
                ]
            ]
            [
                \textbf{Model as a Data Evaluator} (\S\ref{subsec-model-contrib-synthesis-evaluator}), fill=color-layer1!20
                [
                    Quality Assessment, leaf
                ]
                [
                    Ethic Assessment, leaf
                ]
            ]
        ]
    \end{forest}
    \caption{Overview of the model contributions to \mmodal data in terms of data synthesis, categorized by the roles of models.}
    \label{pic-model4data-synthesis}
\end{figure}
\subsection{Model as a Data Creator}
\label{subsec-model-contrib-synthesis-creator}
Models can act as a data creator to extend the cardinality of dataset $\mathcal{D}$, as
$\mathcal{D} = \mathcal{D} \cup \left\{\mathbf{w}(\mathbf{c}) \mid \mathcal{D}, \mathbf{c} \in \mathcal{C} \right\}^m$, 
where $m$ data samples are generated based on candidate contexts in $\mathcal{C}$.

\para{Direct Generation}
Models can directly generate the data ultimately used for training.
For example, \MLLMs can generate text responses to the provided instructions based on the given images \cite{du2023makes}, charts \cite{xia2024chartx}, and 3D inputs \cite{li20243dmit}.
Given existing videos, \MLLMs can generate multi-turn conversations \cite{li2023videochat}. 
Given requirements on topic and row count, charts can be automatically generated \cite{han2023chartllama,chen2024onechart} with rendering libraries.
This method can even be used to generate data containing \mmodal responses \cite{zhan2024anygpt}.
Based on existing images and human-crafted instructions, answers \cite{zhou2024uniaa}, reasoning contexts in steps \cite{tang2024textsquare,mu2023embodiedgpt} and results of counterfactual reasoning \cite{zhang2024if} can also be generated. 
\MLLMs can be employed to provide assessment on data quality \cite{wang2024finetuned} and aesthetics \cite{wu2024multimodal}, thereby building corresponding instruction datasets.

\para{Context-Driven Generation}
\MLLMs with limited capabilities may output contexts as medians to prompt other models for final generation.
\MLLMs without vision generation capabilities have been utilized to generate captions \cite{feng2024improving,gu2023compodiff,ventura2024covr} that then prompt \texttt{Stable Diffusion} for vision output to create fine-tuning data for compositional retrieval. 
A similar approach is applied for data that teach \MLLMs to edit images following human instructions \cite{brooks2023instructpix2pix}.

\subsection{Model as a Data Mapper}
\label{subsec-model-contrib-synthesis-mapper}
Models can enhance data quality by transforming data representations, formalized as:
Given a set of mapping functions $\mathcal{T} = \left\{T_1(\mathbf{u}, \mathbf{w}), T_2(\mathbf{u}, \mathbf{w}), \ldots\right\}$ based on model $\mathbf{w}$ such as rewriting and annotating, $\forall T \in \mathcal{T}$ is applied to each of the samples in $\mathcal{D}$, i.e., $\mathcal{D}' = \left\{ T(\mathbf{u}_i, \mathbf{w}) \mid \mathbf{u}_i \in \mathcal{D}, \forall T \in \mathcal{T} \right\}$.
The transformed data samples may also be merged into $\mathcal{D}$ instead of replacement as $\mathcal{D} = \mathcal{D} \cup \mathcal{D}'$ for a larger quantity.

\para{Quality Enhancement}
\MLLMs can rewrite data samples for higher quality or better application, such as rephrasing the labels in VQA datasets to suit compositional image retrieval \cite{levy2024data} and assembling short titles into long descriptions \cite{zhao2023bubogpt}.
\texttt{COCO} dataset \cite{liu2023visual} can be improved with more detailed captions for data samples generated by \texttt{GPT-4}.

\para{Data Augmentation}
\MLLMs and \LLMs can be used to rewrite data for augmentation and provide different data representations while preserving the original semantics. 
To increase the diversity of vocabulary and sentence structures \cite{li2024data,chivereanu2024aligning}, captions in vision-language datasets can be rewritten to generate data with different sentence structures \cite{fan2023improving}.

\para{Modality Transformation}
\MLLMs can convert data from other modalities into texts, making them understandable for \LLMs \cite{li2023videochat,ma2024aligned} or enable a text-centric alignment between modalities \cite{tsai2024text}. 
These approaches may be promising for semantic communication, where the costs associated with data storage and transmission are greatly alleviated.

\para{Summarization}
\MLLMs can perform summarization on existing data to facilitate data curation.
Time-series data collected from the web can be automatically summarized \cite{jia2024gpt4mts}, and the retrieved information in different modalities can be extracted and fused for the final text answers \cite{zhang2023moqagpt}.

\para{Annotation}
Opposite to summarization, annotation fills in more details between the question and the answer with \MLLMs' world knowledge.
For example, \MLLMs can complete the reasoning processes between the questions and answers \cite{gai2024medthink}, supplement captions with more information such as factual details \cite{chen2024unified}, and refine human annotations on aesthetics \cite{huang2024aesexpert}.
Data from different modalities in documents can be mapped to enable knowledge linkage \cite{tang2024pdfchatannotator}.

\subsection{Model as a Data Filter}
\label{subsec-model-contrib-synthesis-filter}
Models can be used to filter out data samples in $\mathcal{D}$ according to certain criteria as
$\mathcal{D}_{\text{filtered}} = \left\{\mathbf{u}_i \mid \mathbf{u}_i \in \mathcal{D}, F(\mathbf{u}_i, \mathbf{w}) = 1\right\}$, 
where $F$ is a filter function returning 0 or 1, and $\mathcal{D}_{\text{filtered}} \!\subseteq \mathcal{D}$.

\para{Based on Logits}
Some indicators calculated with the model-generated logits can potentially reflect the quality. 
For example, \texttt{CLIP} score can reflect the image-text semantic relevance \cite{gadre2023datacomp,yu2023devil,huang2024visual}, where data samples with low scores can be removed. 
In sampling-based data selection methods, the logits of the surrogate model can determine the sampling probability of each data instance \cite{kolossov2024towards}.

\para{Based on Assessment}
Indicators calculated from logits require manual thresholds and may not align with human behavior.
The assessment provided by \MLLMs on data samples can serve as a basis for data filtering.
Well-trained \MLLMs can be employed to generate numerical scores for image-text datasets \cite{wang2024finetuned} to quantify the matching between images and texts, and answer questions in datasets while removing the and wrongly-answered data \cite{zhang2024if}. 
\MLLMs can also evaluate their self-generated dataset \cite{tang2024textsquare}.

\subsection{Model as a Data Evaluator}
\label{subsec-model-contrib-synthesis-evaluator}
Models as data evaluators facilitate data synthesis via feedback, not manipulation.
This section summarizes existing works that employ models for sample-level data evaluation. 

\para{Quality Assessment}
As \texttt{GPT-4V} aligns with humans in terms of evaluation criteria \cite{ge2024mllmbench}, it can provide feedback on data samples in 1) quantitative evaluation, 2) preference ranking, and 3) quality enhancement suggestions \cite{wu2024multimodal}. 
As an example of \codev, \MLLMs can provide self-evaluation on the generated contents \cite{tang2024textsquare}.
However, while \MLLMs align with human capabilities in pair-ranking, their quantitative scores need improvement \cite{chen2024mllm}.

\para{Ethic Assessment}
Models aligned with human values can uphold ethical standards, helping to check web-crawled and model-generated data for violations.
With carefully designed prompt templates, \MLLMs can self-check their outputs for harmful content \cite{gou2024eyes}.
The success of malicious queries can be assessed with the help of \texttt{ChatGPT} \cite{tao2024imgtrojan}. 
\subsection{Brief Summarization and Discussion}
\label{subsec-model-contrib-synthesis-discussion}
With advanced model capabilities and rising data needs for training \MLLMs, models increasingly handle data curation instead of humans.
A model can serve as 
a data creator (\S\ref{subsec-model-contrib-synthesis-creator}), 
a mapper to transform existing data samples (\S\ref{subsec-model-contrib-synthesis-mapper}), 
a filter to exclude low-quality samples (\S\ref{subsec-model-contrib-synthesis-filter}), 
and 
an evaluator to evaluate or rank data samples (\S\ref{subsec-model-contrib-synthesis-evaluator}). 
From our investigation, current works mainly rely on well-trained \MLLMs such as \texttt{GPT-4V}. 
This paradigm may be insufficient to construct \MLLMs with top-tier scale and performance. 
Thus, it is promising to leverage the increasing knowledge and capabilities of the target \MLLM while training it. 

\section{Model Contributions for \MModal Data: Insights}
\label{sec-model-contrib-application}
\definecolor{color-layer0}{HTML}{FA86B8}
\definecolor{color-layer1}{HTML}{9CE7FF}
\definecolor{color-layer2}{HTML}{5BB0FF}
\definecolor{color-layer3}{HTML}{007BCC}

\tikzstyle{leaf-style}=[
    rectangle,
    draw=hidden-draw,
    rounded corners,
    text opacity=1,
    minimum height=1.5em,
    minimum width=5em,
    inner sep=0.8pt,
    align=center,
    fill opacity=.5,
    line width=0.9pt,
]
\tikzstyle{leaf}=[leaf-style, minimum height=1.5em,
    fill=color-layer2!20, text=black, align=left, font=\scriptsize,
    inner xsep=1.2pt,
    inner ysep=2pt,
    line width=0.8pt,
]

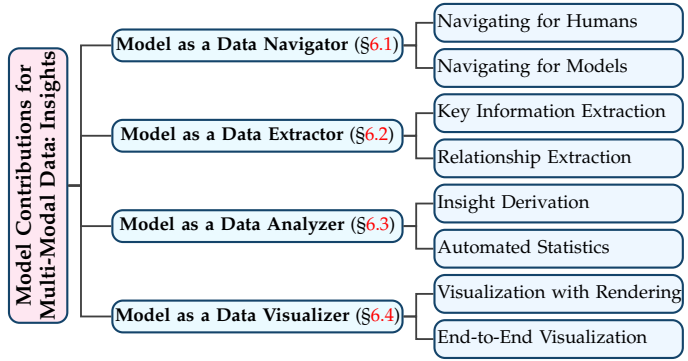
\begin{figure}[ht]
    \centering
    \begin{forest}
        forked edges,
        for tree={
            grow=east,
            reversed=true,
            anchor=base west,
            parent anchor=east,
            child anchor=west,
            base=center,
            font=\small,
            rectangle,
            draw=hidden-draw,
            rounded corners,
            align=center,
            text centered,
            minimum width=4em,
            edge+={darkgray, line width=0.7pt},
            s sep=2pt,
            inner xsep=1pt,
            inner ysep=2pt,
            line width=0.8pt,
            ver/.style={rotate=90, child anchor=north, parent anchor=south, anchor=center},
        },
        where level=0{font=\footnotesize,}{},
        where level=1{text width=11.3em,font=\scriptsize,}{},
        where level=2{text width=9.6em,font=\scriptsize,}{},
        [
            \textbf{Model Contributions for}\\\textbf{\MModal Data: Insights}, ver, fill=color-layer0!20
            [
                \textbf{Model as a Data Navigator} (\S\ref{subsec-model-contrib-application-navigator}), fill=color-layer1!20
                [
                    Navigating for Humans, leaf
                ]
                [
                    Navigating for Models, leaf
                ]
            ]
            [
                \textbf{Model as a Data Extractor} (\S\ref{subsec-model-contrib-application-extractor}), fill=color-layer1!20
                [
                    Key Information Extraction, leaf
                ]
                [
                    Relationship Extraction, leaf
                ]
            ]
            [
                \textbf{Model as a Data Analyzer} (\S\ref{subsec-model-contrib-application-analyzer}), fill=color-layer1!20
                [
                    Insight Derivation, leaf
                ]
                [
                    Automated Statistics, leaf
                ]
            ]
            [
                \textbf{Model as a Data Visualizer} (\S\ref{subsec-model-contrib-application-visualizer}), fill=color-layer1!20
                [
                    Visualization with Rendering, leaf
                ]
                [
                    End-to-End Visualization, leaf
                ]
            ]
        ]
    \end{forest}
    \caption{Overview of the model contributions to \mmodal data in terms of data insights, categorized by the roles of models.}
    \label{pic-model4data-application}
\end{figure}
In addition to data synthesis, models can provide insights to assist in dataset curation and the development of data pipelines, serving as data scientists. 
This section summarizes existing applications of models for data science, grouped by model roles and ordered by the \textit{sequence of tasks performed by a data scientist for dataset curation}, including
\emph{navigator} to assist users in locating the required information from a vast search space (\S\ref{subsec-model-contrib-application-navigator}),
\emph{extractor} to extract key content from vast and disorganized information (\S\ref{subsec-model-contrib-application-extractor}), 
\emph{analyzer} to provide data analysis and statistics (\S\ref{subsec-model-contrib-application-analyzer}),
and
\emph{visualizer} to facilitate data visualization (\S\ref{subsec-model-contrib-application-visualizer}), as illustrated in Fig. \ref{pic-model4data-application}. 
After the introduction of existing works, a brief summarization and discussion are provided in Sec. \ref{subsec-model-contrib-application-discussion}.
\subsection{Model as a Data Navigator}
\label{subsec-model-contrib-application-navigator}
The capability of \LLMs to retrieve external knowledge has been well-recognized \cite{zhu2024large,jing2024large,fernandez2023large}.
\MLLMs can help data consumers quickly locate the needed data or information, with the \emph{understanding} and \emph{retrieval} capabilities.

\para{Navigating for Humans}
\MLLMs can eliminate the need to separate search for data in different modalities for knowledge retrieval based on \mmodal queries, thereby allowing better utilization of the intertwined contextual information across modalities \cite{long2024generative}. 
\MLLMs also facilitate knowledge discovery \cite{miret2024llms}, which guide humans to knowledge yet undiscovered by humanity.

\para{Navigating for Models}
By extending data consumers from humans to models, external knowledge can be navigated for more precise responses \cite{long2024generative}, such as ICL demonstrations \cite{luo2024does}, external knowledge bases \cite{caffagni2024wiki}, and web video descriptions \cite{xu2024retrieval}.
\MLLM outputs also benefit from the entity relationship in \mmodal knowledge graph \cite{chen2024knowledge,lee2024multimodal}.
As Definition \ref{def-co-development}, these methods adopt \MLLMs to provide contexts with external information and knowledge.

\subsection{Model as a Data Extractor}
\label{subsec-model-contrib-application-extractor}
\MLLMs can extract information from unstructured raw data to facilitate further analysis, which is especially important for extensive data.
It can be regarded as a more fine-grained analysis, i.e., the data-sample level, compared to those using \MLLMs for analysis at the dataset level (\S\ref{subsec-model-contrib-application-analyzer}).

\para{Key Information Extraction}
Existing works extract key information from documents \cite{perot2023lmdx,biswas2024robustness}, and reorganize the text outputs generated by \MLLMs into segments and related claims \cite{chen2024unified}.
Named entity recognition for knowledge graph construction can be regarded as a special type of work for key information extraction.
With \MLLMs, entities can be automatically extracted from texts \cite{wu2024structured}, images \cite{udandarao2024zeroshot}, or even grounded \mmodal named entity recognition that further bounds the groundings of entities in images \cite{li2024llms}.

\para{Relationship Extraction}
Knowledge graph construction requires understanding syntax and semantics and, thus, typically involves labor-intensive efforts \cite{zhang2024extract}. 
\MLLMs can help the automatic construction of knowledge graphs by extracting the relationship between \mmodal entities \cite{he2023using}.
This capability can also help to bind different modalities of data within documents, enabling knowledge association~\cite{tang2024pdfchatannotator}.

\subsection{Model as a Data Analyzer}
\label{subsec-model-contrib-application-analyzer}
Data analysis is one of the responsibilities of data scientists and typically includes identifying data quality issues, understanding data distribution, etc.
By jointly leveraging the ability of \MLLMs and \LLMs to \emph{understand} human requirements and its capability to \emph{summarize} data, the workload of data scientists can be significantly reduced \cite{hassan2023chatgpt}.

\para{Insight Derivation}
\MLLMs and \LLMs have been validated to be capable of data analysis and insights \cite{cheng2023gpt,ye2023mplug,hu2024mplug,luo2024llm}. 
\MLLMs with visual understanding of mathematical formulas and tables can help capture critically important information in scientific diagrams \cite{hu2023mplug} or scientific literature \cite{jiang2024bridging,cai2024sciassess}, which can even benefit the preliminary analysis of academic paper quality.

\para{Automated Statistics}
With \emph{understanding} and \emph{reasoning} abilities, \MLLMs can perform automated statistics for charts \cite{xia2024chartx,han2023chartllama,chen2024onechart,masry2022chartqa,xia2023structchart}, where the charts for datasets can be generated by third-party software.
\MLLMs also help derive statistical conclusions from tables or databases \cite{wu2024daco}. 
\subsection{Model as a Data Visualizer}
\label{subsec-model-contrib-application-visualizer}
Data visualization is crucial in data science workflows for data comprehension and insights.
\MLLMs can provide personalized visualizations and handle unstructured data with understanding, reasoning, and generation capabilities.

\para{Visualization with Rendering}
Given visual and text constraints \cite{yang2024posterllava}, \MLLMs can generate layouts to alleviate the manual efforts data scientists expend on visualization styles.
Visualization of raw data and analysis results can be achieved by invoking APIs with \LLMs serving as a code copilot \cite{zhang2023data,dibia2023lida,vazquez2024llms}. 
This kind of method has been applied in some database management works \cite{hassan2023chatgpt,wu2024automated}.

\para{End-to-End Visualization}
Leveraging the image generation and editing abilities of \MLLMs \cite{he2024llms}, visualizations can be performed in an end-to-end manner according to user requirements, and even be adjusted interactively. 
This helps visualization be better aligned with human needs and reduces human effort.
Given an existing chart as a template, \MLLMs can be employed to generate new charts following human instructions \cite{han2023chartllama,yan2024chartreformer}. 
Although promising, this approach may require more exploration to ensure that it can produce results that precisely meet user requirements.

\subsection{Brief Summarization and Discussion}
\label{subsec-model-contrib-application-discussion}
With the emerging capabilities of \MLLMs, they can help data scientists to navigate data consumers to desired data and knowledge (\S\ref{subsec-model-contrib-application-navigator}), extract information and relationships from data (\S\ref{subsec-model-contrib-application-extractor}), provide data insights and statistics (\S\ref{subsec-model-contrib-application-analyzer}), and reduce the manual involvements in data visualization (\S\ref{subsec-model-contrib-application-visualizer}). 
These topics are beginning to take shape, but it will be a while before we can fully rely on \MLLMs as data scientists.

\newcommand{\Eval}{Eval\xspace}
\newcommand{\Pretraining}{Pretrain\xspace}
\newcommand{\PretrainingProj}{Pretrain(P)\xspace}
\newcommand{\Finetuning}{Finetune\xspace}

\newcommand{\Web}{Web\xspace}
\newcommand{\Merge}{Merge\xspace}
\newcommand{\Simulation}{Simulation\xspace}
\newcommand{\Human}{Human\xspace}
\newcommand{\Log}{Log\xspace}

\newcommand{\Image}{Image}
\newcommand{\Text}{Text}
\newcommand{\Audio}{Audio}
\newcommand{\Video}{Video}

\section{Public Datasets for \MLLMs}
\label{sec-resources}
Table \ref{tab-resources-datasets-wi-models} lists public datasets for \MLLMs based on this survey, which is continuously maintained to stay up-to-date.
The operator ``-'' denotes an order-sensitive mapping relationship. 
``Pretrain'' and ``Pretrain(P)'' denote the pretraining of encoders/decoders and projectors, respectively.  

\newcolumntype{C}[1]{>{\centering\arraybackslash}p{#1}}
\begin{table*}[!ht]
\centering
\renewcommand\arraystretch{1.2}
\caption{
Public datasets for \MLLMs based on this survey. 
The \textbf{top} and \underline{bottom} sections list datasets \textbf{with} and \underline{without} models as one of the data sources, respectively.
``Quantity'' indicates the count of data samples/instances.
``Merge'' denotes adopting/merging some existing datasets.
}
\label{tab-resources-datasets-wi-models}
\setlength\tabcolsep{4.75pt}
\scriptsize
\begin{tabularx}{\linewidth}{ccC{2.3cm}cXl}
\toprule[0.8pt]
\multicolumn{1}{c}{Modality} &
\multicolumn{1}{c}{Data Sources}  &
\multicolumn{1}{c}{\MLLM Stage}  &
\multicolumn{1}{c}{Quantity}  &
\multicolumn{1}{c}{Objective}  &
\multicolumn{1}{c}{Reference}
\\
\midrule[0.8pt]


\Video-\Text &
\MLLMs, \Merge &  
\Pretraining &
70M&
Video-language pretraining&
Panda-70M \cite{chen2024panda} \\
\cline{1-6}

\Video-\Text&
BLIP2, \Web &
\Pretraining&
234M&  
Video-related understanding and generation &
InternVid \cite{wang2023internvid} \\
\cline{1-6}

\Image-\Text&
GPT-3.5&
\Pretraining&
9K &
Chart understand \& reasoning&
SimChart9K \cite{xia2023structchart}\\
\cline{1-6}

\Image-\Text&
GPT-4V, \Web, \Merge& 
\Pretraining, \PretrainingProj, \Finetuning&
1.2M&
Pretraining encoder and projector finetuning \MLLM with high-quality image-text data&
ShareGPT4V \cite{chen2023sharegpt4v} \\
\cline{1-6}

\Image-\Text &
GPT-4, \Merge &  
\Pretraining, \Finetuning &
8M&
Enhancing coverage of chart themes &
Chart-Sum-QA \cite{liu2024chartthinker} \\
\cline{1-6}

\Image-\Text&
GPT4-V, \Merge, \Web&
\PretrainingProj, \Finetuning&
1.2M&
Image understanding&
ShareGPT4V \cite{chen2023sharegpt4v} \\
\cline{1-6}

\Image-\Text&
GPT4-V, \Merge& 
\PretrainingProj, \Finetuning &
664K&
Modality alignment \& instruction following&
ALLaVA (Data) \cite{chen2024allava} \\
\cline{1-6}

\Image-\Text&
GPT-4, \Merge&
\Finetuning&
400K&
Mitigating hallucinations&
LRV-Instruction \cite{liu2023mitigating} \\
\cline{1-6}

\Video-\Text&
GPT-4V, \Merge&
\Finetuning&
40K&
Text-video generation &
ShareGPT4Video \cite{chen2024sharegpt4video} \\
\cline{1-6}

\Image-\Text&
Genimi-Pro, \Merge&
\Finetuning&
195K &
Fine-grained image perception&
DocGemini \cite{yu2024texthawk} \\
\cline{1-6}

\Image-\Text&
GPT-4V, \Merge&
\Finetuning&
1.6M&
visual comprehension with textual interaction&
MDVP-Data \cite{lin2024drawandunderstand} \\
\cline{1-6}

\Image-\Text&
Gemini-Pro, \Web&
\Finetuning&
9.1M&
Text-centric VQA&
Square-10M \cite{tang2024textsquare}\\
\cline{1-6}

\Image-\Text&
GPT-4&
\Finetuning&
75K&
3D scene understanding &
3DMIT \cite{li20243dmit} \\
\cline{1-6}

\Video-\Text&
GPT-4, \Merge&  
\Finetuning&
7K&
Video understanding&
VideoChat \cite{li2023videochat}\\
\cline{1-6}

\Image-\Text&
GPT-4&
\Finetuning&
160K&
Chart understanding&
ChartLlama \cite{han2023chartllama} \\
\cline{1-6}

\Image-\Text&
GPT-4, \Merge&
\Finetuning&
32K&
Complex visual reasoning&
ComVint \cite{du2023makes} \\
\cline{1-6}

\Image-\Text&
ChatGPT, \Human, \Merge &
\Finetuning&
2.4M&
Human instruction alignment &
M$^3$IT \cite{li2023m} \\
\cline{1-6}

\Image-\Text &
GPT-4, \Merge, Human &
\Finetuning &
600K&
Chart understanding &
MMC-Instruction \cite{liu2023mmc} \\
\cline{1-6}

\Video-\Text&
BLIP-2, GRiT, \Merge&
\Finetuning&
100K&
Video understanding and conversation&
VideoInstruct-100K \cite{maaz2023video}\\
\cline{1-6}

\Image-\Text&
GPT4-V, \Merge&  
\Finetuning &
80K&
Set-of-Mark understanding &
SoM-LLaVA \cite{yan2024list} \\
\cline{1-6}

\Text-\Image &
GPT-4V, \Merge& 
\Finetuning&
120M&
Generation quality on human preference&
VisionPrefer \cite{wu2024multimodal} \\
\cline{1-6}

\Video-\Text &
ChatGPT, \Merge &
\Finetuning &
2M&
Spatial \& temporal understanding &
VideoChat2-IT \cite{li2023mvbench} \\
\cline{1-6}

\Image-\Text &
GPT-API, \Merge &
\Finetuning &
196K&
3D instruction following &
LAMM \cite{yin2023lamm} \\
\cline{1-6}

\Image-\Text&
ChatGPT, \Merge, \Web&
\Finetuning&
39M &
Chart understanding \& downstream tasks &
ChartSFT \cite{meng2024chartassisstant} \\
\cline{1-6}

\Image-\Text &
GPT-3.5, \Web &
\Finetuning, \Eval &
107K&
Misinformation detection in medical fields &
Med-MMHL \cite{sun2023med} \\
\cline{1-6}

\Image-\Text &
GPT, \Simulation &   
\Finetuning, \Eval &
231K&
Spatial understanding on point clouds &
3DBench \cite{zhang20243dbench} \\
\cline{1-6}

\Image-\Text &
GPT, \Merge, Human &
\Finetuning, \Eval &
11.3K&  
\Mmodal CoT &
M$^3$CoT \cite{chen2024m} \\
\cline{1-6}

\Video/\Image-\Text &
GPT-4, \Merge, \Human &  
\Eval & 
19K&
Spatial \& temporal understanding &
SEED-Bench \cite{li2023seed} \\
\cline{1-6}

\Text-\Video &
GPT-4, \Merge, Human &  
\Eval &
200K& 
Generation quality on human preference &
EvalCrafter \cite{liu2023evalcrafter} \\
\cline{1-6}

\Image-\Text &
GPT-4, Human &
\Eval &
6.9K &  
Image aesthetic assessment &
ImplicitAVE \cite{zou2024implicitave} \\
\cline{1-6}

\Image-\Text &
\MLLMs, \Merge, Human &  
\Eval &
4.4K&  
Evaluations on human-preference alignment &
MLLM-as-a-Judge \cite{chen2024mllm} \\
\cline{1-6}

\Image-\Text/\Image &
ChatGPT, \Merge &
\Eval &
31K &
Comprehensive evaluation of \MLLMs &
MMT-Bench \cite{ying2024mmt} \\
\cline{1-6}

\Image-\Text &
Stable Diffusion, GPT-4 &
\Eval &
5K&  
Generation safety &
MM-SafetyBench \cite{liu2024mmsafetybench} \\
\cline{1-6}

\Image-\Text &
\MLLMs, \Merge, Human  & 
\Eval &
1.2K &  
Visual hallucinations &
VHTest \cite{huang2024visual} \\
\cline{1-6}

\Image-\Text &
ChatGPT, \Merge, Human  &
\Eval &
1.9K&  
Visual hallucinations &
MHaluBench \cite{chen2024unified} \\
\cline{1-6}

\Image-\Text &
GPT-4, Human &
\Eval &
48K&
Reasoning for charts &
ChartX \cite{xia2024chartx} \\
\cline{1-6}

\Image-\Text &
GPT-3.5, Human &
\Eval &
6K&
Chart understanding &
ChartY \cite{chen2024onechart} \\

\midrule[0.8pt]


\Image-\Text&
\Web, \Human&
\Pretraining&
10M& 
Chinese video-language pretraining&
Youku-mPLUG \cite{xu2023youku} \\
\cline{1-6}

\Audio-\Text &
\Web&
\Pretraining &
4.5K &   
Text-audio retrieval &
WavText5K \cite{deshmukh2022audio} \\
\cline{1-6}

\Video-\Text&
\Web, \Human&
\Pretraining&
13K&  
Video understanding&
BLiSS \cite{he2023align} \\
\cline{1-6}

\Video-\Text &
\Merge, \Web&
\Pretraining &
101.2M& 
Image-Text interleaved pretraining &
MMC4 \cite{zhu2023multimodal} \\
\cline{1-6}

\Image-\Text &
\Merge&
\Pretraining &
12.8M-12.8B&
Image-language pretraining&
DataComp \cite{gadre2023datacomp} \\
\cline{1-6}

\Audio-\Text &
\Web &
\Pretraining &
630K &
Pretraining of audio \& text encoders &
LAION-Audio-630K \cite{wu2023large} \\
\cline{1-6}

\Image-\Text&
\Web&
\Finetuning&
21K & 
CoT generation&
ScienceQA \cite{lu2022learn} \\
\cline{1-6}

\Audio/\Image-\Text&
\Merge&
\Finetuning&
5K&
Triple-modality instruction following&
BuboGPT \cite{zhao2023bubogpt} \\
\cline{1-6}

\Image-\Text &
\Web, \Human&
\Eval &
130K &
Performance with long-tailed concepts &
Let it Wag \cite{udandarao2024zeroshot} \\
\cline{1-6}

\Text-\Video &
\Merge, Human &
\Eval &
800&
Generation quality on human preference &
VBench \cite{huang2023vbench} \\
\cline{1-6}

\Video-\Text &
Human &
\Eval &
11.6k& 
Temporal understanding, perception &
Perception Test \cite{patraucean2023perception} \\
\cline{1-6}

\Video-\Text &
\Web, Human &
\Eval &
312K & 
Reasoning, surprising video understanding &
FunQA \cite{xie2023funqa} \\
\cline{1-6}

\Text-\Video &
Human &
\Eval &
11M&
Generation quality, state transfer&
WorldNet-Wild \cite{ge2024worldgpt} \\
\cline{1-6}

\Image-\Text &
\Merge, Human &
\Eval &
32.7K&
Evaluation for logical operations with charts &
ChartQA \cite{masry2022chartqa} \\
\cline{1-6}

\Image-\Text &
\Merge, Human &
\Eval &
3.8K&  
Evaluation for visual understanding &
BLINK \cite{fu2024blink} \\
\cline{1-6}

\Image-\Text&
\Merge, \Human&
\Eval&
15K&
Evaluation for mathematical reasoning&
MathVerse \cite{zhang2024mathverse} \\
\cline{1-6}

\Image-\Text &
Human &
\Eval &
420&
Evaluation for open-ended QA &
MLLM-Bench \cite{ge2024mllmbench} \\

\bottomrule[0.8pt]
\end{tabularx}
\end{table*}

\section{Roadmap for Future \MLLMs}
\label{sec-future}
From this survey, existing \MLLM works of \codev are mainly performed in the externally-boosted paradigm as Definition \ref{def-co-development}. 
We provide some promising future directions progressively organized as follows: 
1) infrastructure development (\S\ref{subsec-future-infrastructure}), 
2) \MLLM research with \codev in an externally-boosted manner, where we try to point out the next phase of this widely applied paradigm (\S\ref{subsec-future-externally-boosted}), 
and 
3) \MLLM research with \codev in a self-boosted manner, a less explored but promising paradigm in the foreseeable future (\S\ref{subsec-future-self-boosted}). 
These progressive directions form a roadmap for \MLLM research in the coming few years.

\subsection{\CODEV Infrastructures}
\label{subsec-future-infrastructure}
To advance \codev for \MLLMs, more advancements are required in the infrastructure for a more conducive environment. 
Firstly, scalable data management suites for \MLLMs should be implemented to support efficient curation of large-scale \mmodal datasets.
Then, efficient hardware and algorithms, e.g., quantization and mixed precision algorithms, can enhance the iteration between \MLLM training and synthesis of \mmodal datasets.
Finally, there is a pressing need for more seamless integration between existing data-centric and model-centric infrastructures. 
These tools allow developers to effortlessly design and test prototypes for diverse datasets, models and application scenarios. 
An initial attempt \cite{dj-sandbox} in this direction builds a sandbox with several built-in experimental scaffolds for \codev of \MLLMs.

\subsection{Externally-Boosted \MLLM Development}
\label{subsec-future-externally-boosted}
Based on convenient infrastructures, we can explore some advanced research topics under the externally-boosted paradigm of \codev by leveraging well-trained \MLLMs, including data discovery, modality-compatibility detection and automatic knowledge transfer, organized in order of first scaling up the data and then improving \MLLMs' usability as that in Sec. \ref{sec-data-contrib-scaling} and \ref{sec-data-contrib-usability}.

\subsubsection{\MLLM-Based Data Discovery}  
\label{subsubsec-future-automatic-data-discovery}
Data discovery fundamentally supports scaling and usability of \MLLMs as in Sec. \ref{subsec-data-contrib-scaling-up} and \ref{subsec-data-contrib-scaling-effectiveness}. 
It needs understanding the requirements, retrieving relevant data, and statistics to gain an initial outline of data characteristics, often requiring heavy human efforts. 
As in Sec. \ref{sec-model-contrib-application}, existing works reveal the capabilities of \MLLMs in terms of knowledge discovery \cite{miret2024llms} and dataset analysis \cite{wu2024daco}. 
With these capabilities, it is promising to enable automatic data discovery to alleviate human labor and keep the knowledge contained in \MLLMs up to date by continually acquiring valuable \mmodal data from the internet, such as images and videos from news.
This topic leads to some minor research issues.

\para{Automatic Long-tail Data Discovery}
\MLLMs may perform sub-optimal with inputs belonging to long-tail domains \cite{udandarao2024zeroshot}, potentially misclassifying some long-tail as worthless, wasting data that could be even more meaningful.
Thus, it is needed to especially enhance \MLLMs in discerning whether data belongs to long-tail domains.

\para{Automatic Compliance with Privacy and Licensing Requirements}
Data discovery must comply with privacy and usage regulations \cite{duan2024modelgo}
Enabling \MLLMs to understand licenses and avoid privacy violations in an automatic data discovery pipeline, is a worthwhile research direction.

\subsubsection{Modality-Compatibility Detection with \MLLMs}  
\label{subsec-future-automatic-modality-detection}
From Sec. \ref{subsubsec-data-contrib-scaling-effectiveness-crossmodal-alignment}, prevalent model-driven techniques for assessing the compatibility across modalities mainly leverage pre-trained foundation models, which occasionally yield inaccurate assessments, as a low compatibility score could stem not from modal mismatches but rather from suboptimal data quality \cite{mahmoud2023sieve}.
Text-centric anchoring helps with this by refining captions \cite{nguyen2023improving}. 
Our study reveals that aligning across modalities by \MLLMs with non-text anchors remains an underexplored frontier.
Expanding anchoring to encompass additional modalities may harness a richer array of signals and broaden the representational landscape, potentially elevating the efficacy of alignment endeavors.

\subsubsection{Automatic Knowledge Transfer for \MLLMs}  
\label{subsec-future-automatic-knowledge-transfer}
Training \MLLMs on datasets generated with top-tier models essentially distills knowledge from them.
It still requires human involvement, such as specifying a transfer set.
As well-trained \MLLMs can provide suggestions to quality enhancement \cite{wu2024multimodal}, it is promising to explore automatic knowledge transfer, especially transferring between \MLLMs that excel in different modalities.
Specifically, a teacher \MLLM can question a student \MLLM to assess its shortcomings, and generate tailored data to bridge these gaps.
The key lies in creating an automated pipeline to use well-trained \MLLMs for evaluating target models.

\subsection{Self-Boosted \MLLM Development}
\label{subsec-future-self-boosted}
The self-boosted paradigm of \codev (\S\ref{sec-preliminary}) does not rely on the availability of well-trained \MLLMs, making it more versatile and capable of supporting the development of \MLLMs with top-tier scale and performance.
Potential points for consideration cover the scaling and usability of \MLLMs as discussed in Sec. \ref{subsec-data-contrib-scaling-up}, \ref{subsec-data-contrib-scaling-effectiveness}, \ref{sec-data-contrib-usability}. 

\subsubsection{Self Data Scaling with \MLLMs}
\label{subsec-future-self-data-boosting}
\para{Data Scaling with Single \MLLM}
Existing works show that alternatively using models to assist in data curation and improving models with the newly collected data can achieve \codev \cite{kirillov2023segment}. 
For any-to-any \MLLMs, it is promising to alternatively optimize data in different modalities, such as image recaptioning \cite{li2023blip,chen2024allava,mahmoud2023sieve}, while optimizing them with the continually optimized/growing data, which alleviates copyright concerns.

\para{Data Scaling with Cooperative \MLLMs}
Inspired by deep mutual learning \cite{zhang2018deep} which trains multiple models that learn from each other and results in a better performance than training a single model, data scaling may be enabled with multiple cooperative \MLLMs, i.e., training multiple \MLLMs simultaneously and having them jointly improve the datasets with their different knowledge. 
A recent study \cite{dj-sandbox} finds that alternative recaptioning with an image-text model and image regeneration with a text-image model can continuously boost the data quality and quantity \cite{dj-sandbox}, which provides a prototype for the exploration cooperative \MLLMs that excel in different modalities.

\subsubsection{Self Data Condensation with \MLLMs}
From Sec. \ref{subsubsec-data-contrib-scaling-effectiveness-condensation}, there are some model-based metrics and assessments \cite{gadre2023datacomp,wang2024finetuned,yu2023devil,mahmoud2023sieve,fang2023data,nguyen2023improving} that help to condensate \mmodal data without reliance of human heuristics.
They essentially use the criteria of one model to assess the value of the data for another model, failing to be tailored for the specific needs of target models.

If we consider \MLLMs as intelligent entities, they should best understand what type of data they need.
Thus, allowing the \MLLM to select the necessary training data for itself is promising, where the evaluation can be performed based on quality assessment \cite{wu2024multimodal,tang2024textsquare}, quantitative score \cite{wang2024finetuned}, or ranking among data samples \cite{chen2024mllm}. 
With this, the condensation can be dynamically adjusted based on the \MLLM's current state during training, as some data, mistakenly perceived as low quality or modality mismatch due to difficulties in model comprehension, can be reused for training after the capabilities of the \MLLM are improved.

\subsubsection{RL from Self Feedback of \MLLMs}
Reinforcement learning (RL) from human feedback (RLHF) aligns \LLMs and \MLLMs with human preferences based on human feedback \cite{bai2022constitutional}. 
However, reliance on human feedback limits scalability, prompting the rise of RL with artificial intelligence feedback (RLAIF) \cite{leerlaif}.
Curating human feedback for \MLLMs may be more labor-intensive than \LLMs due to multiple modalities. 
Well-trained \MLLMs can provide ranking assessment aligned with human feedback \cite{chen2024mllm}, enabling AI feedback for \mmodal contents \cite{yu2024rlaif}. 
Just like humans can evaluate the quality of their own statements, allowing an \MLLM under training to provide feedback on its own output may be a promising direction. 
In the early training stage of an \MLLM, some human involvement may still be needed.
Besides, fine-grained feedback rather than scores could be better for \MLLMs \cite{liang2024rich,wu2024multimodal}.

\section{Conclusions}
\label{sec-conclusion}
This paper highlights the dual effect and great potential of simultaneously improving both data and models for \MLLM development, sketching a new \codev paradigm through a comprehensive and systematic review of existing works for \MLLMs.
Specifically, we first examine the contributions that data can make to \MLLMs by first scaling up \MLLMs and then improving the usability of \MLLMs. 
Next, we discuss how models can facilitate data curation by serving for data synthesis as data developers and providing data insights as data scientists.
We identify that the current \codev paradigm in \MLLMs often still requires well-trained models or human assistance.
In light of this, we summarize numerous potential future directions ranging from infrastructure development to self-boosted \codev for \MLLMs, furnishing a roadmap for future \MLLMs from the \codev perspective. 
Through these efforts, we hope to offer timely guidance and inspire more innovation in both research and applications of \MLLMs, fostering a deeper integration of both data and model advancements.

\balance
\bibliographystyle{IEEEtran}
\bibliography{references}

\begin{thebibliography}{100}
\providecommand{\url}[1]{#1}
\csname url@samestyle\endcsname
\providecommand{\newblock}{\relax}
\providecommand{\bibinfo}[2]{#2}
\providecommand{\BIBentrySTDinterwordspacing}{\spaceskip=0pt\relax}
\providecommand{\BIBentryALTinterwordstretchfactor}{4}
\providecommand{\BIBentryALTinterwordspacing}{\spaceskip=\fontdimen2\font plus
\BIBentryALTinterwordstretchfactor\fontdimen3\font minus \fontdimen4\font\relax}
\providecommand{\BIBforeignlanguage}[2]{{%
\expandafter\ifx\csname l@#1\endcsname\relax
\typeout{** WARNING: IEEEtran.bst: No hyphenation pattern has been}%
\typeout{** loaded for the language `#1'. Using the pattern for}%
\typeout{** the default language instead.}%
\else
\language=\csname l@#1\endcsname
\fi
#2}}
\providecommand{\BIBdecl}{\relax}
\BIBdecl

\bibitem{reid2024gemini}
M.~Reid, N.~Savinov, D.~Teplyashin, D.~Lepikhin, T.~Lillicrap, J.-b. Alayrac, R.~Soricut, A.~Lazaridou, O.~Firat, J.~Schrittwieser \emph{et~al.}, ``Gemini 1.5: Unlocking multimodal understanding across millions of tokens of context,'' \emph{arXiv:2403.05530}, 2024.

\bibitem{openai2024sora}
OpenAI, ``Sora: Creating video from text,'' \url{https://openai.com/sora}, 2024.

\bibitem{openai2024gpt4o}
{OpenAI}, ``Hello gpt-4o,'' \url{https://openai.com/index/hello-gpt-4o/}, 2024.

\bibitem{wu2023nextgpt}
S.~Wu, H.~Fei, L.~Qu, W.~Ji, and T.-S. Chua, ``Next-gpt: Any-to-any multimodal llm,'' \emph{arXiv:2309.05519}, 2023.

\bibitem{survey-zhao2023survey}
W.~X. Zhao, K.~Zhou, J.~Li, T.~Tang, X.~Wang, Y.~Hou, Y.~Min, B.~Zhang, J.~Zhang, Z.~Dong \emph{et~al.}, ``A survey of large language models,'' \emph{arXiv:2303.18223}, 2023.

\bibitem{villalobos2022will}
P.~Villalobos, J.~Sevilla, L.~Heim, T.~Besiroglu, M.~Hobbhahn, and A.~Ho, ``Will we run out of data? an analysis of the limits of scaling datasets in machine learning,'' \emph{arXiv:2211.04325}, 2022.

\bibitem{goyal2024scaling}
S.~Goyal, P.~Maini, Z.~C. Lipton, A.~Raghunathan, and J.~Z. Kolter, ``Scaling laws for data filtering--data curation cannot be compute agnostic,'' in \emph{CVPR}, 2024, pp. 22\,702--22\,711.

\bibitem{gao2024lumina}
P.~Gao, L.~Zhuo, Z.~Lin, C.~Liu, J.~Chen, R.~Du, E.~Xie, X.~Luo, L.~Qiu, Y.~Zhang \emph{et~al.}, ``Lumina-t2x: Transforming text into any modality, resolution, and duration via flow-based large diffusion transformers,'' \emph{arXiv:2405.05945}, 2024.

\bibitem{kaplan2020scaling}
J.~Kaplan, S.~McCandlish, T.~Henighan, T.~B. Brown, B.~Chess, R.~Child, S.~Gray, A.~Radford, J.~Wu, and D.~Amodei, ``Scaling laws for neural language models,'' \emph{arXiv:2001.08361}, 2020.

\bibitem{aghajanyan2023scaling}
A.~Aghajanyan, L.~Yu, A.~Conneau, W.-N. Hsu, K.~Hambardzumyan, S.~Zhang, S.~Roller, N.~Goyal, O.~Levy, and L.~Zettlemoyer, ``Scaling laws for generative mixed-modal language models,'' in \emph{ICML}, 2023, pp. 265--279.

\bibitem{survey-zha2023datacentric}
D.~Zha, Z.~P. Bhat, K.-H. Lai, F.~Yang, Z.~Jiang, S.~Zhong, and X.~Hu, ``Data-centric artificial intelligence: A survey,'' \emph{arXiv:2303.10158}, 2023.

\bibitem{Salehi2023Data-green}
S.~Salehi and A.~Schmeink, ``Data-centric green artificial intelligence: A survey,'' \emph{IEEE Transactions on Artificial Intelligence}, pp. 1--18, 2023.

\bibitem{survey-jakubik2024datacentric}
J.~Jakubik, M.~V{\"o}ssing, N.~K{\"u}hl, J.~Walk, and G.~Satzger, ``Data-centric artificial intelligence,'' \emph{Business \& Information Systems Engineering}, pp. 1--9, 2024.

\bibitem{gadre2023datacomp}
S.~Y. Gadre, G.~Ilharco, A.~Fang, J.~Hayase, G.~Smyrnis, T.~Nguyen, R.~Marten, M.~Wortsman, D.~Ghosh, J.~Zhang \emph{et~al.}, ``Datacomp: In search of the next generation of multimodal datasets,'' \emph{NeurIPS}, vol.~36, 2023.

\bibitem{fan2023improving}
L.~Fan, D.~Krishnan, P.~Isola, D.~Katabi, and Y.~Tian, ``Improving clip training with language rewrites,'' \emph{NeurIPS}, vol.~36, 2023.

\bibitem{he2024efficient}
M.~He, Y.~Liu, B.~Wu, J.~Yuan, Y.~Wang, T.~Huang, and B.~Zhao, ``Efficient multimodal learning from data-centric perspective,'' \emph{arXiv:2402.11530}, 2024.

\bibitem{mmsurvey-carolan2024review}
K.~Carolan, L.~Fennelly, and A.~F. Smeaton, ``A review of multi-modal large language and vision models,'' \emph{arXiv:2404.01322}, 2024.

\bibitem{mmsurvey-caffagni2024revolution}
D.~Caffagni, F.~Cocchi, L.~Barsellotti, N.~Moratelli, S.~Sarto, L.~Baraldi, M.~Cornia, and R.~Cucchiara, ``The (r)evolution of multimodal large language models: A survey,'' \emph{arXiv:2402.12451}, 2024.

\bibitem{mmsurvey-xu2024survey}
M.~Xu, W.~Yin, D.~Cai, R.~Yi, D.~Xu, Q.~Wang, B.~Wu, Y.~Zhao, C.~Yang, S.~Wang \emph{et~al.}, ``A survey of resource-efficient llm and multimodal foundation models,'' \emph{arXiv:2401.08092}, 2024.

\bibitem{mmsurvey-li2023multimodal}
C.~Li, Z.~Gan, Z.~Yang, J.~Yang, L.~Li, L.~Wang, J.~Gao \emph{et~al.}, ``Multimodal foundation models: From specialists to general-purpose assistants,'' \emph{Foundations and Trends{\textregistered} in Computer Graphics and Vision}, vol.~16, no. 1-2, pp. 1--214, 2024.

\bibitem{mmsurvey-zhang2024mmllms}
D.~Zhang, Y.~Yu, C.~Li, J.~Dong, D.~Su, C.~Chu, and D.~Yu, ``{MM-LLMs}: Recent advances in multimodal large language models,'' \emph{arXiv:2401.13601}, 2024.

\bibitem{mmsurvey-10445007}
J.~Zhang, J.~Huang, S.~Jin, and S.~Lu, ``Vision-language models for vision tasks: A survey,'' \emph{IEEE Transactions on Pattern Analysis and Machine Intelligence}, pp. 1--20, 2024.

\bibitem{liu2024multimodal}
Q.~Liu, J.~Zhu, Y.~Yang, Q.~Dai, Z.~Du, X.-M. Wu, Z.~Zhao, R.~Zhang, and Z.~Dong, ``Multimodal pretraining, adaptation, and generation for recommendation: A survey,'' \emph{arXiv:2404.00621}, 2024.

\bibitem{tang2024video}
Y.~Tang, J.~Bi, S.~Xu, L.~Song, S.~Liang, T.~Wang, D.~Zhang, J.~An, J.~Lin, R.~Zhu \emph{et~al.}, ``Video understanding with large language models: A survey,'' \emph{arXiv:2312.17432}, 2023.

\bibitem{survey-jin2024efficient}
Y.~Jin, J.~Li, Y.~Liu, T.~Gu, K.~Wu, Z.~Jiang, M.~He, B.~Zhao, X.~Tan, Z.~Gan \emph{et~al.}, ``Efficient multimodal large language models: A survey,'' \emph{arXiv:2405.10739}, 2024.

\bibitem{mmsurvey-huang2023visual}
J.~Huang, J.~Zhang, K.~Jiang, H.~Qiu, and S.~Lu, ``Visual instruction tuning towards general-purpose multimodal model: A survey,'' \emph{arXiv:2312.16602}, 2023.

\bibitem{mmsurvey-yin2024survey}
S.~Yin, C.~Fu, S.~Zhao, K.~Li, X.~Sun, T.~Xu, and E.~Chen, ``A survey on multimodal large language models,'' \emph{arXiv:2306.13549}, 2023.

\bibitem{mmsurvey-wu2023multimodal}
J.~Wu, W.~Gan, Z.~Chen, S.~Wan, and S.~Y. Philip, ``Multimodal large language models: A survey,'' in \emph{IEEE BigData}, 2023, pp. 2247--2256.

\bibitem{survey-10123038}
P.~Xu, X.~Zhu, and D.~A. Clifton, ``Multimodal learning with transformers: A survey,'' \emph{IEEE Transactions on Pattern Analysis and Machine Intelligence}, vol.~45, no.~10, pp. 12\,113--12\,132, 2023.

\bibitem{zhao2024deep}
F.~Zhao, C.~Zhang, and B.~Geng, ``Deep multimodal data fusion,'' \emph{ACM Computing Surveys}, vol.~56, no.~9, pp. 1--36, 2024.

\bibitem{mmsurvey-wang2024exploring}
Y.~Wang, W.~Chen, X.~Han, X.~Lin, H.~Zhao, Y.~Liu, B.~Zhai, J.~Yuan, Q.~You, and H.~Yang, ``Exploring the reasoning abilities of multimodal large language models (mllms): A comprehensive survey on emerging trends in multimodal reasoning,'' \emph{arXiv:2401.06805}, 2024.

\bibitem{zhou2024survey}
P.~Zhou, L.~Wang, Z.~Liu, Y.~Hao, P.~Hui, S.~Tarkoma, and J.~Kangasharju, ``A survey on generative ai and llm for video generation, understanding, and streaming,'' \emph{arXiv:2404.16038}, 2024.

\bibitem{mmsurvey-zhao2024survey}
T.~Zhao, L.~Zhang, Y.~Ma, and L.~Cheng, ``A survey on safe multi-modal learning system,'' \emph{arXiv:2402.05355}, 2024.

\bibitem{bai2024survey}
T.~Bai, H.~Liang, B.~Wan, L.~Yang, B.~Li, Y.~Wang, B.~Cui, C.~He, B.~Yuan, and W.~Zhang, ``A survey of multimodal large language model from a data-centric perspective,'' \emph{arXiv:2405.16640}, 2024.

\bibitem{kirillov2023segment}
A.~Kirillov, E.~Mintun, N.~Ravi, H.~Mao, C.~Rolland, L.~Gustafson, T.~Xiao, S.~Whitehead, A.~C. Berg, W.-Y. Lo \emph{et~al.}, ``Segment anything,'' in \emph{ICCV}, 2023, pp. 4015--4026.

\bibitem{xu2023cit}
H.~Xu, S.~Xie, P.-Y. Huang, L.~Yu, R.~Howes, G.~Ghosh, L.~Zettlemoyer, and C.~Feichtenhofer, ``Cit: Curation in training for effective vision-language data,'' in \emph{ICCV}, 2023, pp. 15\,180--15\,189.

\bibitem{chen2024allava}
G.~H. Chen, S.~Chen, R.~Zhang, J.~Chen, X.~Wu, Z.~Zhang, Z.~Chen, J.~Li, X.~Wan, and B.~Wang, ``Allava: Harnessing gpt4v-synthesized data for a lite vision-language model,'' \emph{arXiv:2402.11684}, 2024.

\bibitem{rombach2022high}
R.~Rombach, A.~Blattmann, D.~Lorenz, P.~Esser, and B.~Ommer, ``High-resolution image synthesis with latent diffusion models,'' in \emph{CVPR}, 2022, pp. 10\,684--10\,695.

\bibitem{li2023self}
X.~Li, P.~Yu, C.~Zhou, T.~Schick, L.~Zettlemoyer, O.~Levy, J.~Weston, and M.~Lewis, ``Self-alignment with instruction backtranslation,'' \emph{arXiv:2308.06259}, 2023.

\bibitem{survey-zha2023data}
D.~Zha, K.-H. Lai, F.~Yang, N.~Zou, H.~Gao, and X.~Hu, ``Data-centric ai: Techniques and future perspectives,'' in \emph{KDD}, 2023, pp. 5839--5840.

\bibitem{survey-wang2023data}
Z.~Wang, W.~Zhong, Y.~Wang, Q.~Zhu, F.~Mi, B.~Wang, L.~Shang, X.~Jiang, and Q.~Liu, ``Data management for large language models: A survey,'' \emph{arXiv:2312.01700}, 2023.

\bibitem{long2024llms}
L.~Long, R.~Wang, R.~Xiao, J.~Zhao, X.~Ding, G.~Chen, and H.~Wang, ``On {LLMs}-driven synthetic data generation, curation, and evaluation: A survey,'' \emph{arXiv:2406.15126}, 2024.

\bibitem{survey-albalak2024survey-selection}
A.~Albalak, Y.~Elazar, S.~M. Xie, S.~Longpre, N.~Lambert, X.~Wang, N.~Muennighoff, B.~Hou, L.~Pan, H.~Jeong \emph{et~al.}, ``A survey on data selection for language models,'' \emph{arXiv:2402.16827}, 2024.

\bibitem{survey-wang2024survey-selection}
J.~Wang, B.~Zhang, Q.~Du, J.~Zhang, and D.~Chu, ``A survey on data selection for llm instruction tuning,'' \emph{arXiv:2402.05123}, 2024.

\bibitem{survey-ding2024data-aug}
B.~Ding, C.~Qin, R.~Zhao, T.~Luo, X.~Li, G.~Chen, W.~Xia, J.~Hu, A.~T. Luu, and S.~Joty, ``Data augmentation using llms: Data perspectives, learning paradigms and challenges,'' \emph{arXiv:2403.02990}, 2024.

\bibitem{touvron2023llama}
H.~Touvron, T.~Lavril, G.~Izacard, X.~Martinet, M.~Lachaux, T.~Lacroix, B.~Rozi{\`{e}}re, N.~Goyal, E.~Hambro, F.~Azhar, A.~Rodriguez, A.~Joulin, E.~Grave, and G.~Lample, ``{LLaMA}: Open and efficient foundation language models,'' \emph{arXiv:2302.13971}, 2023.

\bibitem{dosovitskiy2021an}
A.~Dosovitskiy, L.~Beyer, A.~Kolesnikov, D.~Weissenborn, X.~Zhai, T.~Unterthiner, M.~Dehghani, M.~Minderer, G.~Heigold, S.~Gelly, J.~Uszkoreit, and N.~Houlsby, ``An image is worth 16x16 words: Transformers for image recognition at scale,'' in \emph{ICLR}, 2021.

\bibitem{radford2021learning}
A.~Radford, J.~W. Kim, C.~Hallacy, A.~Ramesh, G.~Goh, S.~Agarwal, G.~Sastry, A.~Askell, P.~Mishkin, J.~Clark \emph{et~al.}, ``Learning transferable visual models from natural language supervision,'' in \emph{ICML}, 2021, pp. 8748--8763.

\bibitem{liu2023visual}
H.~Liu, C.~Li, Q.~Wu, and Y.~J. Lee, ``Visual instruction tuning,'' \emph{NeurIPS}, vol.~36, 2023.

\bibitem{gao2023llama}
P.~Gao, J.~Han, R.~Zhang, Z.~Lin, S.~Geng, A.~Zhou, W.~Zhang, P.~Lu, C.~He, X.~Yue \emph{et~al.}, ``Llama-adapter v2: Parameter-efficient visual instruction model,'' \emph{arXiv:2304.15010}, 2023.

\bibitem{hoffmann2022training}
J.~Hoffmann, S.~Borgeaud, A.~Mensch, E.~Buchatskaya, T.~Cai, E.~Rutherford, D.~d.~L. Casas, L.~A. Hendricks, J.~Welbl, A.~Clark \emph{et~al.}, ``Training compute-optimal large language models,'' \emph{arXiv:2203.15556}, 2022.

\bibitem{udandarao2024zeroshot}
V.~Udandarao, A.~Prabhu, A.~Ghosh, Y.~Sharma, P.~H. Torr, A.~Bibi, S.~Albanie, and M.~Bethge, ``No" zero-shot" without exponential data: Pretraining concept frequency determines multimodal model performance,'' \emph{arXiv:2404.04125}, 2024.

\bibitem{liu2024chartthinker}
M.~Liu, D.~Chen, Y.~Li, G.~Fang, and Y.~Shen, ``Chartthinker: A contextual chain-of-thought approach to optimized chart summarization,'' \emph{arXiv:2403.11236}, 2024.

\bibitem{chen2023sharegpt4v}
L.~Chen, J.~Li, X.~Dong, P.~Zhang, C.~He, J.~Wang, F.~Zhao, and D.~Lin, ``Sharegpt4v: Improving large multi-modal models with better captions,'' \emph{arXiv:2311.12793}, 2023.

\bibitem{wu2024uiclip}
J.~Wu, Y.-H. Peng, A.~Li, A.~Swearngin, J.~P. Bigham, and J.~Nichols, ``Uiclip: A data-driven model for assessing user interface design,'' \emph{arXiv:2404.12500}, 2024.

\bibitem{jia2024gpt4mts}
F.~Jia, K.~Wang, Y.~Zheng, D.~Cao, and Y.~Liu, ``{GPT4MTS}: Prompt-based large language model for multimodal time-series forecasting,'' in \emph{AAAI}, vol.~38, no.~21, 2024, pp. 23\,343--23\,351.

\bibitem{wu2023large}
Y.~Wu, K.~Chen, T.~Zhang, Y.~Hui, T.~Berg-Kirkpatrick, and S.~Dubnov, ``Large-scale contrastive language-audio pretraining with feature fusion and keyword-to-caption augmentation,'' in \emph{ICASSP}, 2023, pp. 1--5.

\bibitem{deshmukh2022audio}
S.~Deshmukh, B.~Elizalde, and H.~Wang, ``Audio retrieval with {WavText5k} and clap training,'' \emph{arXiv:2209.14275}, 2022.

\bibitem{yan2024chartreformer}
P.~Yan, M.~Bhosale, J.~Lal, B.~Adhikari, and D.~Doermann, ``Chartreformer: Natural language-driven chart image editing,'' \emph{arXiv:2403.00209}, 2024.

\bibitem{yan2024list}
A.~Yan, Z.~Yang, J.~Wu, W.~Zhu, J.~Yang, L.~Li, K.~Lin, J.~Wang, J.~McAuley, J.~Gao \emph{et~al.}, ``List items one by one: A new data source and learning paradigm for multimodal {LLMs},'' \emph{arXiv:2404.16375}, 2024.

\bibitem{tang2024textsquare}
J.~Tang, C.~Lin, Z.~Zhao, S.~Wei, B.~Wu, Q.~Liu, H.~Feng, Y.~Li, S.~Wang, L.~Liao \emph{et~al.}, ``{TextSquare}: Scaling up text-centric visual instruction tuning,'' \emph{arXiv:2404.12803}, 2024.

\bibitem{zou2024implicitave}
H.~P. Zou, V.~Samuel, Y.~Zhou, W.~Zhang, L.~Fang, Z.~Song, P.~S. Yu, and C.~Caragea, ``{ImplicitAVE}: An open-source dataset and multimodal llms benchmark for implicit attribute value extraction,'' \emph{arXiv:2404.15592}, 2024.

\bibitem{yu2024texthawk}
Y.-Q. Yu, M.~Liao, J.~Wu, Y.~Liao, X.~Zheng, and W.~Zeng, ``Texthawk: Exploring efficient fine-grained perception of multimodal large language models,'' \emph{arXiv:2404.09204}, 2024.

\bibitem{zhao2023bubogpt}
Y.~Zhao, Z.~Lin, D.~Zhou, Z.~Huang, J.~Feng, and B.~Kang, ``{BuboGPT}: Enabling visual grounding in multi-modal {LLMs},'' \emph{arXiv:2307.08581}, 2023.

\bibitem{lu2022learn}
P.~Lu, S.~Mishra, T.~Xia, L.~Qiu, K.-W. Chang, S.-C. Zhu, O.~Tafjord, P.~Clark, and A.~Kalyan, ``Learn to explain: Multimodal reasoning via thought chains for science question answering,'' \emph{NeurIPS}, vol.~35, pp. 2507--2521, 2022.

\bibitem{li2024hunyuan}
Z.~Li, J.~Zhang, Q.~Lin, J.~Xiong, Y.~Long, X.~Deng, Y.~Zhang, X.~Liu, M.~Huang, Z.~Xiao \emph{et~al.}, ``Hunyuan-dit: A powerful multi-resolution diffusion transformer with fine-grained chinese understanding,'' \emph{arXiv:2405.08748}, 2024.

\bibitem{maaz2023video}
M.~Maaz, H.~Rasheed, S.~Khan, and F.~S. Khan, ``Video-chatgpt: Towards detailed video understanding via large vision and language models,'' \emph{arXiv:2306.05424}, 2023.

\bibitem{xia2023structchart}
R.~Xia, B.~Zhang, H.~Peng, N.~Liao, P.~Ye, B.~Shi, J.~Yan, and Y.~Qiao, ``Structchart: Perception, structuring, reasoning for visual chart understanding,'' \emph{arXiv:2309.11268}, 2023.

\bibitem{li2024deep}
Y.-Y. Li, Y.~Bai, C.~Wang, M.~Qu, Z.~Lu, R.~Soria, and J.~Liu, ``Deep learning and llm-based methods applied to stellar lightcurve classification,'' \emph{arXiv:2404.10757}, 2024.

\bibitem{wu2024multimodal}
X.~Wu, S.~Huang, and F.~Wei, ``Multimodal large language model is a human-aligned annotator for text-to-image generation,'' \emph{arXiv:2404.15100}, 2024.

\bibitem{liu2023mmc}
F.~Liu, X.~Wang, W.~Yao, J.~Chen, K.~Song, S.~Cho, Y.~Yacoob, and D.~Yu, ``{MMC}: Advancing multimodal chart understanding with large-scale instruction tuning,'' \emph{arXiv:2311.10774}, 2023.

\bibitem{mu2023embodiedgpt}
Y.~Mu, Q.~Zhang, M.~Hu, W.~Wang, M.~Ding, J.~Jin, B.~Wang, J.~Dai, Y.~Qiao, and P.~Luo, ``{EmbodiedGPT}: Vision-language pre-training via embodied chain of thought,'' \emph{NeurIPS}, vol.~36, 2023.

\bibitem{sreeram2024probing}
S.~Sreeram, T.-H. Wang, A.~Maalouf, G.~Rosman, S.~Karaman, and D.~Rus, ``Probing multimodal llms as world models for driving,'' \emph{arXiv:2405.05956}, 2024.

\bibitem{jia2021scaling}
C.~Jia, Y.~Yang, Y.~Xia, Y.-T. Chen, Z.~Parekh, H.~Pham, Q.~Le, Y.-H. Sung, Z.~Li, and T.~Duerig, ``Scaling up visual and vision-language representation learning with noisy text supervision,'' in \emph{ICML}, 2021, pp. 4904--4916.

\bibitem{ye2024mplug}
Q.~Ye, H.~Xu, J.~Ye, M.~Yan, A.~Hu, H.~Liu, Q.~Qian, J.~Zhang, and F.~Huang, ``mplug-owl2: Revolutionizing multi-modal large language model with modality collaboration,'' in \emph{CVPR}, 2024, pp. 13\,040--13\,051.

\bibitem{li2023blip}
J.~Li, D.~Li, S.~Savarese, and S.~Hoi, ``Blip-2: Bootstrapping language-image pre-training with frozen image encoders and large language models,'' in \emph{ICML}, 2023, pp. 19\,730--19\,742.

\bibitem{ye2023mplug}
J.~Ye, A.~Hu, H.~Xu, Q.~Ye, M.~Yan, Y.~Dan, C.~Zhao, G.~Xu, C.~Li, J.~Tian \emph{et~al.}, ``mplug-docowl: Modularized multimodal large language model for document understanding,'' \emph{arXiv:2307.02499}, 2023.

\bibitem{chen2024visual}
D.~Chen, J.~Liu, W.~Dai, and B.~Wang, ``Visual instruction tuning with polite flamingo,'' in \emph{AAAI}, vol.~38, no.~16, 2024, pp. 17\,745--17\,753.

\bibitem{vallaeys2024improved}
T.~Vallaeys, M.~Shukor, M.~Cord, and J.~Verbeek, ``Improved baselines for data-efficient perceptual augmentation of llms,'' \emph{arXiv:2403.13499}, 2024.

\bibitem{li2024data}
Z.~Li, L.~Si, C.~Guo, Y.~Yang, and Q.~Cao, ``Data augmentation for text-based person retrieval using large language models,'' \emph{arXiv:2405.11971}, 2024.

\bibitem{chivereanu2024aligning}
R.~Chivereanu, A.~Cosma, A.~Catruna, R.~Rughinis, and E.~Radoi, ``Aligning actions and walking to llm-generated textual descriptions,'' \emph{arXiv:2404.12192}, 2024.

\bibitem{yu2024capsfusion}
Q.~Yu, Q.~Sun, X.~Zhang, Y.~Cui, F.~Zhang, Y.~Cao, X.~Wang, and J.~Liu, ``Capsfusion: Rethinking image-text data at scale,'' in \emph{CVPR}, 2024, pp. 14\,022--14\,032.

\bibitem{yang2023gpt4tools}
R.~Yang, L.~Song, Y.~Li, S.~Zhao, Y.~Ge, X.~Li, and Y.~Shan, ``{GPT4Tools}: Teaching large language model to use tools via self-instruction,'' \emph{NeurIPS}, vol.~36, 2023.

\bibitem{feng2024improving}
Z.~Feng, R.~Zhang, and Z.~Nie, ``Improving composed image retrieval via contrastive learning with scaling positives and negatives,'' \emph{arXiv:2404.11317}, 2024.

\bibitem{alayrac2022flamingo}
J.-B. Alayrac, J.~Donahue, P.~Luc, A.~Miech, I.~Barr, Y.~Hasson, K.~Lenc, A.~Mensch, K.~Millican, M.~Reynolds \emph{et~al.}, ``Flamingo: a visual language model for few-shot learning,'' \emph{NeurIPS}, vol.~35, pp. 23\,716--23\,736, 2022.

\bibitem{gao2024sphinx}
P.~Gao, R.~Zhang, C.~Liu, L.~Qiu, S.~Huang, W.~Lin, S.~Zhao, S.~Geng, Z.~Lin, P.~Jin \emph{et~al.}, ``{SPHINX-X}: Scaling data and parameters for a family of multi-modal large language models,'' \emph{arXiv:2402.05935}, 2024.

\bibitem{tong2024cambrian}
S.~Tong, E.~Brown, P.~Wu, S.~Woo, M.~Middepogu, S.~C. Akula, J.~Yang, S.~Yang, A.~Iyer, X.~Pan \emph{et~al.}, ``Cambrian-1: A fully open, vision-centric exploration of multimodal {LLMs},'' \emph{arXiv:2406.16860}, 2024.

\bibitem{driess2023palm}
D.~Driess, F.~Xia, M.~S. Sajjadi, C.~Lynch, A.~Chowdhery, B.~Ichter, A.~Wahid, J.~Tompson, Q.~Vuong, T.~Yu \emph{et~al.}, ``{PaLM-E}: An embodied multimodal language model,'' in \emph{ICML}, 2023, pp. 8469--8488.

\bibitem{liu2023retrieval}
B.~Liu, C.~Lyu, Z.~Min, Z.~Wang, J.~Su, and L.~Wang, ``Retrieval-augmented multi-modal chain-of-thoughts reasoning for large language models,'' \emph{arXiv:2312.01714}, 2023.

\bibitem{chen2024datajuicer}
D.~Chen, Y.~Huang, Z.~Ma, H.~Chen, X.~Pan, C.~Ge, D.~Gao, Y.~Xie, Z.~Liu, J.~Gao, Y.~Li, B.~Ding, and J.~Zhou, ``Data-juicer: A one-stop data processing system for large language models,'' in \emph{SIGMOD}, 2024.

\bibitem{zhou2023lima}
C.~Zhou, P.~Liu, P.~Xu, S.~Iyer, J.~Sun, Y.~Mao, X.~Ma, A.~Efrat, P.~Yu, L.~Yu \emph{et~al.}, ``Lima: Less is more for alignment,'' \emph{NeurIPS}, vol.~36, 2023.

\bibitem{kolossov2024towards}
G.~Kolossov, A.~Montanari, and P.~Tandon, ``Towards a statistical theory of data selection under weak supervision,'' in \emph{ICLR}, 2024.

\bibitem{sorscher2022beyond}
B.~Sorscher, R.~Geirhos, S.~Shekhar, S.~Ganguli, and A.~Morcos, ``Beyond neural scaling laws: beating power law scaling via data pruning,'' \emph{NeurIPS}, vol.~35, pp. 19\,523--19\,536, 2022.

\bibitem{webster2023duplication}
R.~Webster, J.~Rabin, L.~Simon, and F.~Jurie, ``On the de-duplication of laion-2b,'' \emph{arXiv:2303.12733}, 2023.

\bibitem{jafari2021survey}
O.~Jafari, P.~Maurya, P.~Nagarkar, K.~M. Islam, and C.~Crushev, ``A survey on locality sensitive hashing algorithms and their applications,'' \emph{arXiv:2102.08942}, 2021.

\bibitem{beaumont-2022-clip-retrieval}
R.~Beaumont, ``Clip retrieval: Easily compute clip embeddings and build a clip retrieval system with them,'' \url{https://github.com/rom1504/clip-retrieval}, 2022.

\bibitem{theis2022lossy}
L.~Theis, W.~Shi, A.~Cunningham, and F.~Husz{\'a}r, ``Lossy image compression with compressive autoencoders,'' in \emph{ICLR}, 2022.

\bibitem{abbas2023semdedup}
A.~Abbas, K.~Tirumala, D.~Simig, S.~Ganguli, and A.~S. Morcos, ``Semdedup: Data-efficient learning at web-scale through semantic deduplication,'' \emph{arXiv:2303.09540}, 2023.

\bibitem{huang2024multimodal}
T.-H. Huang, C.~Shin, S.~J. Tay, D.~Adila, and F.~Sala, ``Multimodal data curation via object detection and filter ensembles,'' \emph{arXiv:2401.12225}, 2024.

\bibitem{radenovic2023filtering}
F.~Radenovic, A.~Dubey, A.~Kadian, T.~Mihaylov, S.~Vandenhende, Y.~Patel, Y.~Wen, V.~Ramanathan, and D.~Mahajan, ``Filtering, distillation, and hard negatives for vision-language pre-training,'' in \emph{CVPR}, 2023, pp. 6967--6977.

\bibitem{mahmoud2023sieve}
A.~Mahmoud, M.~Elhoushi, A.~Abbas, Y.~Yang, N.~Ardalani, H.~Leather, and A.~Morcos, ``Sieve: Multimodal dataset pruning using image captioning models,'' \emph{arXiv:2310.02110}, 2023.

\bibitem{maini2023t}
P.~Maini, S.~Goyal, Z.~C. Lipton, J.~Z. Kolter, and A.~Raghunathan, ``T-mars: Improving visual representations by circumventing text feature learning,'' \emph{arXiv:2307.03132}, 2023.

\bibitem{nguyen2023improving}
T.~Nguyen, S.~Y. Gadre, G.~Ilharco, S.~Oh, and L.~Schmidt, ``Improving multimodal datasets with image captioning,'' \emph{NeurIPS}, vol.~36, 2023.

\bibitem{yu2023devil}
H.~Yu, Y.~Tian, S.~Kumar, L.~Yang, and H.~Wang, ``The devil is in the details: A deep dive into the rabbit hole of data filtering,'' \emph{arXiv:2309.15954}, 2023.

\bibitem{wang2024finetuned}
W.~Wang, K.~Mrini, L.~Yang, S.~Kumar, Y.~Tian, X.~Yan, and H.~Wang, ``Finetuned multimodal language models are high-quality image-text data filters,'' \emph{arXiv:2403.02677}, 2024.

\bibitem{fang2023data}
A.~Fang, A.~M. Jose, A.~Jain, L.~Schmidt, A.~Toshev, and V.~Shankar, ``Data filtering networks,'' \emph{arXiv:2309.17425}, 2023.

\bibitem{wei2023instructiongpt}
L.~Wei, Z.~Jiang, W.~Huang, and L.~Sun, ``Instructiongpt-4: A 200-instruction paradigm for fine-tuning minigpt-4,'' \emph{arXiv:2308.12067}, 2023.

\bibitem{xu2024demystifying}
H.~Xu, S.~Xie, X.~Tan, P.-Y. Huang, R.~Howes, V.~Sharma, S.-W. Li, G.~Ghosh, L.~Zettlemoyer, and C.~Feichtenhofer, ``Demystifying {CLIP} data,'' in \emph{ICLR}, 2024.

\bibitem{tsai2024text}
Y.-D. Tsai, T.-Y. Yen, P.-F. Guo, Z.-Y. Li, and S.-D. Lin, ``Text-centric alignment for multi-modality learning,'' \emph{arXiv:2402.08086}, 2024.

\bibitem{nguyen2022quality}
T.~Nguyen, G.~Ilharco, M.~Wortsman, S.~Oh, and L.~Schmidt, ``Quality not quantity: On the interaction between dataset design and robustness of clip,'' \emph{NeurIPS}, vol.~35, pp. 21\,455--21\,469, 2022.

\bibitem{liu2024decade}
Z.~Liu and K.~He, ``A decade's battle on dataset bias: Are we there yet?'' \emph{arXiv:2403.08632}, 2024.

\bibitem{ma2024mode}
J.~Ma, P.-Y. Huang, S.~Xie, S.-W. Li, L.~Zettlemoyer, S.-F. Chang, W.-T. Yih, and H.~Xu, ``Mode: Clip data experts via clustering,'' \emph{arXiv:2404.16030}, 2024.

\bibitem{dehghani2024patch}
M.~Dehghani, B.~Mustafa, J.~Djolonga, J.~Heek, M.~Minderer, M.~Caron, A.~Steiner, J.~Puigcerver, R.~Geirhos, I.~M. Alabdulmohsin \emph{et~al.}, ``Patch n’pack: Navit, a vision transformer for any aspect ratio and resolution,'' \emph{NeurIPS}, vol.~36, 2024.

\bibitem{liu2024sora}
Y.~Liu, K.~Zhang, Y.~Li, Z.~Yan, C.~Gao, R.~Chen, Z.~Yuan, Y.~Huang, H.~Sun, J.~Gao \emph{et~al.}, ``Sora: A review on background, technology, limitations, and opportunities of large vision models,'' \emph{arXiv:2402.17177}, 2024.

\bibitem{ding2024fewer}
H.~Ding, Z.~Wang, G.~Paolini, V.~Kumar, A.~Deoras, D.~Roth, and S.~Soatto, ``Fewer truncations improve language modeling,'' \emph{arXiv:2404.10830}, 2024.

\bibitem{staniszewski2023structured}
K.~Staniszewski, S.~Tworkowski, S.~Jaszczur, H.~Michalewski, {\L}.~Kuci{\'n}ski, and P.~Mi{\l}o{\'s}, ``Structured packing in llm training improves long context utilization,'' \emph{arXiv:2312.17296}, 2023.

\bibitem{han2023noisy}
H.~Han, K.~Miao, Q.~Zheng, and M.~Luo, ``Noisy correspondence learning with meta similarity correction,'' in \emph{CVPR}, 2023, pp. 7517--7526.

\bibitem{fuyu-8b}
R.~Bavishi, E.~Elsen, C.~Hawthorne, M.~Nye, A.~Odena, A.~Somani, and S.~Ta\c{s}\i{}rlar, ``Introducing our multimodal models,'' 2023.

\bibitem{blattmann2023stable}
A.~Blattmann, T.~Dockhorn, S.~Kulal, D.~Mendelevitch, M.~Kilian, D.~Lorenz, Y.~Levi, Z.~English, V.~Voleti, A.~Letts \emph{et~al.}, ``Stable video diffusion: Scaling latent video diffusion models to large datasets,'' \emph{arXiv:2311.15127}, 2023.

\bibitem{zhao2024aligngpt}
F.~Zhao, T.~Pang, C.~Li, Z.~Wu, J.~Guo, S.~Xing, and X.~Dai, ``Aligngpt: Multi-modal large language models with adaptive alignment capability,'' \emph{arXiv:2405.14129}, 2024.

\bibitem{maharana2023d2}
A.~Maharana, P.~Yadav, and M.~Bansal, ``$\mathbb{D}^2$ pruning: Message passing for balancing diversity \& difficulty in data pruning,'' in \emph{ICLR}, 2024.

\bibitem{bimix}
C.~Ge, Z.~Ma, D.~Chen, Y.~Li, and B.~Ding, ``Data mixing made efficient: A bivariate scaling law for language model pretraining,'' \emph{arXiv:2405.14908}, 2024.

\bibitem{yang2023set}
J.~Yang, H.~Zhang, F.~Li, X.~Zou, C.~Li, and J.~Gao, ``Set-of-mark prompting unleashes extraordinary visual grounding in gpt-4v,'' \emph{arXiv:2310.11441}, 2023.

\bibitem{lin2024drawandunderstand}
W.~Lin, X.~Wei, R.~An, P.~Gao, B.~Zou, Y.~Luo, S.~Huang, S.~Zhang, and H.~Li, ``Draw-and-understand: Leveraging visual prompts to enable {MLLMs} to comprehend what you want,'' \emph{arXiv:2403.20271}, 2024.

\bibitem{gou2024eyes}
Y.~Gou, K.~Chen, Z.~Liu, L.~Hong, H.~Xu, Z.~Li, D.-Y. Yeung, J.~T. Kwok, and Y.~Zhang, ``Eyes closed, safety on: Protecting multimodal llms via image-to-text transformation,'' \emph{arXiv:2403.09572}, 2024.

\bibitem{chen2023shikra}
K.~Chen, Z.~Zhang, W.~Zeng, R.~Zhang, F.~Zhu, and R.~Zhao, ``Shikra: Unleashing multimodal llm's referential dialogue magic,'' \emph{arXiv:2306.15195}, 2023.

\bibitem{peng2023kosmos}
Z.~Peng, W.~Wang, L.~Dong, Y.~Hao, S.~Huang, S.~Ma, and F.~Wei, ``Kosmos-2: Grounding multimodal large language models to the world,'' \emph{arXiv:2306.14824}, 2023.

\bibitem{jin2023time}
M.~Jin, S.~Wang, L.~Ma, Z.~Chu, J.~Y. Zhang, X.~Shi, P.-Y. Chen, Y.~Liang, Y.-F. Li, S.~Pan \emph{et~al.}, ``{Time-LLM}: Time series forecasting by reprogramming large language models,'' \emph{arXiv:2310.01728}, 2023.

\bibitem{fan2024scaling}
L.~Fan, K.~Chen, D.~Krishnan, D.~Katabi, P.~Isola, and Y.~Tian, ``Scaling laws of synthetic images for model training... for now,'' in \emph{CVPR}, 2024, pp. 7382--7392.

\bibitem{sun2024generative}
Q.~Sun, Y.~Cui, X.~Zhang, F.~Zhang, Q.~Yu, Y.~Wang, Y.~Rao, J.~Liu, T.~Huang, and X.~Wang, ``Generative multimodal models are in-context learners,'' in \emph{CVPR}, 2024, pp. 14\,398--14\,409.

\bibitem{li2024groundinggpt}
Z.~Li, Q.~Xu, D.~Zhang, H.~Song, Y.~Cai, Q.~Qi, R.~Zhou, J.~Pan, Z.~Li, V.~T. Vu \emph{et~al.}, ``Groundinggpt: Language enhanced multi-modal grounding model,'' \emph{arXiv:2401.06071}, 2024.

\bibitem{zhao2024mmicl}
H.~Zhao, Z.~Cai, S.~Si, X.~Ma, K.~An, L.~Chen, Z.~Liu, S.~Wang, W.~Han, and B.~Chang, ``{MMICL}: Empowering vision-language model with multi-modal in-context learning,'' in \emph{ICLR}, 2024.

\bibitem{doveh2024towards}
S.~Doveh, S.~Perek, M.~J. Mirza, A.~Alfassy, A.~Arbelle, S.~Ullman, and L.~Karlinsky, ``Towards multimodal in-context learning for vision \& language models,'' \emph{arXiv:2403.12736}, 2024.

\bibitem{gao2024aim}
J.~Gao, Q.~Qiao, Z.~Cao, Z.~Wang, and W.~Li, ``{AIM}: Let any multi-modal large language models embrace efficient in-context learning,'' \emph{arXiv:2406.07588}, 2024.

\bibitem{wang2024aggregatedimageinimagelearning}
L.~Wang, W.~Xu, Z.~Hu, Y.~Lan, S.~Dong, H.~Wang, R.~K.-W. Lee, and E.-P. Lim, ``All in an aggregated image for in-image learning,'' \emph{arXiv:2402.17971}, 2024.

\bibitem{bai2024hallucination}
Z.~Bai, P.~Wang, T.~Xiao, T.~He, Z.~Han, Z.~Zhang, and M.~Z. Shou, ``Hallucination of multimodal large language models: A survey,'' \emph{arXiv:2404.18930}, 2024.

\bibitem{yin2023lamm}
Z.~Yin, J.~Wang, J.~Cao, Z.~Shi, D.~Liu, M.~Li, X.~Huang, Z.~Wang, L.~Sheng, L.~Bai \emph{et~al.}, ``Lamm: Language-assisted multi-modal instruction-tuning dataset, framework, and benchmark,'' \emph{NeurIPS}, vol.~36, 2023.

\bibitem{ge2024mllmbench}
W.~Ge, S.~Chen, G.~Chen, J.~Chen, Z.~Chen, S.~Yan, C.~Zhu, Z.~Lin, W.~Xie, X.~Wang \emph{et~al.}, ``{MLLM}-bench: Evaluating multimodal llms with per-sample criteria,'' \emph{arXiv:2311.13951}, 2023.

\bibitem{chen2024dress}
Y.~Chen, K.~Sikka, M.~Cogswell, H.~Ji, and A.~Divakaran, ``Dress: Instructing large vision-language models to align and interact with humans via natural language feedback,'' in \emph{CVPR}, 2024, pp. 14\,239--14\,250.

\bibitem{Li2023SilkiePD}
L.~Li, Z.~Xie, M.~Li, S.~Chen, P.~Wang, L.~Chen, Y.~Yang, B.~Wang, and L.~Kong, ``Silkie: Preference distillation for large visual language models,'' \emph{arXiv:2405.2312.10665}, 2023.

\bibitem{yu2024rlhf}
T.~Yu, Y.~Yao, H.~Zhang, T.~He, Y.~Han, G.~Cui, J.~Hu, Z.~Liu, H.-T. Zheng, M.~Sun \emph{et~al.}, ``{RLHF-V}: Towards trustworthy mllms via behavior alignment from fine-grained correctional human feedback,'' in \emph{CVPR}, 2024, pp. 13\,807--13\,816.

\bibitem{Zhang2024AutomatedMP}
M.~Zhang and K.~Rong, ``Automated multi-level preference for mllms,'' \emph{arXiv:2405.11165}, 2024.

\bibitem{sun2023aligning}
Z.~Sun, S.~Shen, S.~Cao, H.~Liu, C.~Li, Y.~Shen, C.~Gan, L.-Y. Gui, Y.-X. Wang, Y.~Yang \emph{et~al.}, ``Aligning large multimodal models with factually augmented {RLHF},'' \emph{arXiv:2309.14525}, 2023.

\bibitem{huang2024opera}
Q.~Huang, X.~Dong, P.~Zhang, B.~Wang, C.~He, J.~Wang, D.~Lin, W.~Zhang, and N.~Yu, ``Opera: Alleviating hallucination in multi-modal large language models via over-trust penalty and retrospection-allocation,'' in \emph{CVPR}, 2024, pp. 13\,418--13\,427.

\bibitem{liu2023mitigating}
F.~Liu, K.~Lin, L.~Li, J.~Wang, Y.~Yacoob, and L.~Wang, ``Mitigating hallucination in large multi-modal models via robust instruction tuning,'' in \emph{ICLR}, 2023.

\bibitem{xie2023funqa}
B.~Xie, S.~Zhang, Z.~Zhou, B.~Li, Y.~Zhang, J.~Hessel, J.~Yang, and Z.~Liu, ``{FunQA}: Towards surprising video comprehension,'' \emph{arXiv:2306.14899}, 2023.

\bibitem{gai2024medthink}
X.~Gai, C.~Zhou, J.~Liu, Y.~Feng, J.~Wu, and Z.~Liu, ``{MedThink}: Explaining medical visual question answering via multimodal decision-making rationale,'' \emph{arXiv:2404.12372}, 2024.

\bibitem{zheng2023ddcot}
G.~Zheng, B.~Yang, J.~Tang, H.-Y. Zhou, and S.~Yang, ``{DDCoT}: Duty-distinct chain-of-thought prompting for multimodal reasoning in language models,'' \emph{NeurIPS}, vol.~36, pp. 5168--5191, 2023.

\bibitem{cho2024language}
J.~H. Cho, B.~Ivanovic, Y.~Cao, E.~Schmerling, Y.~Wang, X.~Weng, B.~Li, Y.~You, P.~Kr{\"a}henb{\"u}hl, Y.~Wang \emph{et~al.}, ``Language-image models with 3d understanding,'' \emph{arXiv:2405.03685}, 2024.

\bibitem{chen2023grounding}
H.~Chen, X.~Wang, H.~Chen, Z.~Song, J.~Jia, and W.~Zhu, ``{Grounding-Prompter}: Prompting {LLM} with multimodal information for temporal sentence grounding in long videos,'' \emph{arXiv:2312.17117}, 2023.

\bibitem{yang2023mm}
Z.~Yang, L.~Li, J.~Wang, K.~Lin, E.~Azarnasab, F.~Ahmed, Z.~Liu, C.~Liu, M.~Zeng, and L.~Wang, ``Mm-react: Prompting chatgpt for multimodal reasoning and action,'' \emph{arXiv:2303.11381}, 2023.

\bibitem{lu2024chameleon}
P.~Lu, B.~Peng, H.~Cheng, M.~Galley, K.-W. Chang, Y.~N. Wu, S.-C. Zhu, and J.~Gao, ``Chameleon: Plug-and-play compositional reasoning with large language models,'' \emph{NeurIPS}, vol.~36, 2023.

\bibitem{gupta2023visual}
T.~Gupta and A.~Kembhavi, ``Visual programming: Compositional visual reasoning without training,'' in \emph{CVPR}, 2023, pp. 14\,953--14\,962.

\bibitem{fan2024unbridled}
Y.~Fan, Y.~Cao, Z.~Zhao, Z.~Liu, and S.~Li, ``Unbridled icarus: A survey of the potential perils of image inputs in multimodal large language model security,'' \emph{arXiv:2404.05264}, 2024.

\bibitem{pi2024mllmprotector}
R.~Pi, T.~Han, Y.~Xie, R.~Pan, Q.~Lian, H.~Dong, J.~Zhang, and T.~Zhang, ``Mllm-protector: Ensuring mllm's safety without hurting performance,'' \emph{arXiv:2401.02906}, 2024.

\bibitem{wang2023exploring}
Y.~Wang, W.~Hu, Y.~Dong, and R.~Hong, ``Exploring transferability of multimodal adversarial samples for vision-language pre-training models with contrastive learning,'' \emph{arXiv:2308.12636}, 2023.

\bibitem{lu2023set}
D.~Lu, Z.~Wang, T.~Wang, W.~Guan, H.~Gao, and F.~Zheng, ``Set-level guidance attack: Boosting adversarial transferability of vision-language pre-training models,'' in \emph{ICCV}, 2023, pp. 102--111.

\bibitem{schlarmann2023adversarial}
C.~Schlarmann and M.~Hein, ``On the adversarial robustness of multi-modal foundation models,'' in \emph{CVPR}, 2023, pp. 3677--3685.

\bibitem{he2023sa}
B.~He, X.~Jia, S.~Liang, T.~Lou, Y.~Liu, and X.~Cao, ``Sa-attack: Improving adversarial transferability of vision-language pre-training models via self-augmentation,'' \emph{arXiv:2312.04913}, 2023.

\bibitem{tao2024imgtrojan}
X.~Tao, S.~Zhong, L.~Li, Q.~Liu, and L.~Kong, ``Imgtrojan: Jailbreaking vision-language models with one image,'' \emph{arXiv:2403.02910}, 2024.

\bibitem{bagdasaryan2023ab}
E.~Bagdasaryan, T.-Y. Hsieh, B.~Nassi, and V.~Shmatikov, ``(ab) using images and sounds for indirect instruction injection in multi-modal llms,'' \emph{arXiv:2307.10490}, 2023.

\bibitem{wu2023jailbreaking}
Y.~Wu, X.~Li, Y.~Liu, P.~Zhou, and L.~Sun, ``Jailbreaking gpt-4v via self-adversarial attacks with system prompts,'' \emph{arXiv:2311.09127}, 2023.

\bibitem{tan2024wolf}
Z.~Tan, C.~Zhao, R.~Moraffah, Y.~Li, Y.~Kong, T.~Chen, and H.~Liu, ``The wolf within: Covert injection of malice into mllm societies via an mllm operative,'' \emph{arXiv:2402.14859}, 2024.

\bibitem{liang2024vl}
J.~Liang, S.~Liang, M.~Luo, A.~Liu, D.~Han, E.-C. Chang, and X.~Cao, ``Vl-trojan: Multimodal instruction backdoor attacks against autoregressive visual language models,'' \emph{arXiv:2402.13851}, 2024.

\bibitem{liu2024mmsafetybench}
X.~Liu, Y.~Zhu, J.~Gu, Y.~Lan, C.~Yang, and Y.~Qiao, ``Mm-safetybench: A benchmark for safety evaluation of multimodal large language models,'' \emph{arXiv:2311.17600}, 2023.

\bibitem{sun2023med}
Y.~Sun, J.~He, S.~Lei, L.~Cui, and C.-T. Lu, ``Med-mmhl: A multi-modal dataset for detecting human-and llm-generated misinformation in the medical domain,'' \emph{arXiv:2306.08871}, 2023.

\bibitem{lukas2023analyzing}
N.~Lukas, A.~Salem, R.~Sim, S.~Tople, L.~Wutschitz, and S.~Zanella-B{\'e}guelin, ``Analyzing leakage of personally identifiable information in language models,'' in \emph{SP}.\hskip 1em plus 0.5em minus 0.4em\relax IEEE, 2023, pp. 346--363.

\bibitem{li2024red}
M.~Li, L.~Li, Y.~Yin, M.~Ahmed, Z.~Liu, and Q.~Liu, ``Red teaming visual language models,'' \emph{arXiv:2401.12915}, 2024.

\bibitem{rao2023building}
J.~Rao, S.~Gao, G.~Mai, and K.~Janowicz, ``Building privacy-preserving and secure geospatial artificial intelligence foundation models (vision paper),'' in \emph{SIGSPATIAL}, 2023, pp. 1--4.

\bibitem{achiam2023gpt}
J.~Achiam, S.~Adler, S.~Agarwal, L.~Ahmad, I.~Akkaya, F.~L. Aleman, D.~Almeida, J.~Altenschmidt, S.~Altman, S.~Anadkat \emph{et~al.}, ``{GPT-4} technical report,'' \emph{arXiv:2303.08774}, 2023.

\bibitem{joshi2024Human-Analysis}
I.~Joshi, M.~Grimmer, C.~Rathgeb, C.~Busch, F.~Bremond, and A.~Dantcheva, ``Synthetic data in human analysis: A survey,'' \emph{IEEE Transactions on Pattern Analysis and Machine Intelligence}, pp. 1--20, 2024.

\bibitem{huang2023safeguarding}
A.~Huang, P.~Liu, R.~Nakada, L.~Zhang, and W.~Zhang, ``Safeguarding data in multimodal {AI}: A differentially private approach to clip training,'' \emph{arXiv:2306.08173}, 2023.

\bibitem{mcmahan2017communication}
B.~McMahan, E.~Moore, D.~Ramage, S.~Hampson, and B.~A. y~Arcas, ``Communication-efficient learning of deep networks from decentralized data,'' in \emph{AISTATS}, 2017, pp. 1273--1282.

\bibitem{on-device-llm}
Z.~Qin, D.~Chen, B.~Qian, B.~Ding, Y.~Li, and S.~Deng, ``Federated full-parameter tuning of billion-sized language models with communication cost under 18 kilobytes,'' in \emph{ICML}, 2024.

\bibitem{zero-order-llm}
Z.~Ling, D.~Chen, L.~Yao, Y.~Li, and Y.~Shen, ``On the convergence of zeroth-order federated tuning for large language models,'' in \emph{KDD}, 2024.

\bibitem{bai2024federated}
J.~Bai, D.~Chen, B.~Qian, L.~Yao, and Y.~Li, ``Federated fine-tuning of large language models under heterogeneous language tasks and client resources,'' \emph{arXiv:2402.11505}, 2024.

\bibitem{andrews2023ethical}
J.~Andrews, D.~Zhao, W.~Thong, A.~Modas, O.~Papakyriakopoulos, and A.~Xiang, ``Ethical considerations for responsible data curation,'' \emph{NeurIPS}, vol.~36, 2023.

\bibitem{ombredanne2020free}
P.~Ombredanne, ``Free and open source software license compliance: Tools for software composition analysis,'' \emph{Computer}, vol.~53, no.~10, pp. 105--109, 2020.

\bibitem{german2010sentence}
D.~M. German, Y.~Manabe, and K.~Inoue, ``A sentence-matching method for automatic license identification of source code files,'' in \emph{ASE}, 2010, pp. 437--446.

\bibitem{duan2024modelgo}
M.~Duan, Q.~Li, and B.~He, ``Modelgo: A practical tool for machine learning license analysis,'' in \emph{WWW}, 2024, pp. 1158--1169.

\bibitem{tang2023watermarking}
Y.~Tang, J.~Yu, K.~Gai, X.~Qu, Y.~Hu, G.~Xiong, and Q.~Wu, ``Watermarking vision-language pre-trained models for multi-modal embedding as a service,'' \emph{arXiv:2311.05863}, 2023.

\bibitem{fu2024blink}
X.~Fu, Y.~Hu, B.~Li, Y.~Feng, H.~Wang, X.~Lin, D.~Roth, N.~A. Smith, W.-C. Ma, and R.~Krishna, ``Blink: Multimodal large language models can see but not perceive,'' \emph{arXiv:2404.12390}, 2024.

\bibitem{chen2024onechart}
J.~Chen, L.~Kong, H.~Wei, C.~Liu, Z.~Ge, L.~Zhao, J.~Sun, C.~Han, and X.~Zhang, ``Onechart: Purify the chart structural extraction via one auxiliary token,'' \emph{arXiv:2404.09987}, 2024.

\bibitem{zhou2024uniaa}
Z.~Zhou, Q.~Wang, B.~Lin, Y.~Su, R.~Chen, X.~Tao, A.~Zheng, L.~Yuan, P.~Wan, and D.~Zhang, ``Uniaa: A unified multi-modal image aesthetic assessment baseline and benchmark,'' \emph{arXiv:2404.09619}, 2024.

\bibitem{shi2023open}
Y.~Shi, F.~Lv, X.~Wang, C.~Xia, S.~Li, S.~Yang, T.~Xi, and G.~Zhang, ``Open-transmind: A new baseline and benchmark for 1st foundation model challenge of intelligent transportation,'' in \emph{CVPR}, 2023, pp. 6327--6334.

\bibitem{zhang20243dbench}
J.~Zhang, T.~Hu, X.~Huang, Y.~Gong, and D.~Zeng, ``3dbench: A scalable 3d benchmark and instruction-tuning dataset,'' \emph{arXiv:2404.14678}, 2024.

\bibitem{li2023m3dbench}
M.~Li, X.~Chen, C.~Zhang, S.~Chen, H.~Zhu, F.~Yin, G.~Yu, and T.~Chen, ``M3dbench: Let's instruct large models with multi-modal 3d prompts,'' \emph{arXiv:2312.10763}, 2023.

\bibitem{li2023mvbench}
K.~Li, Y.~Wang, Y.~He, Y.~Li, Y.~Wang, Y.~Liu, Z.~Wang, J.~Xu, G.~Chen, P.~Luo \emph{et~al.}, ``{MVBench}: A comprehensive multi-modal video understanding benchmark,'' \emph{arXiv:2311.17005}, 2023.

\bibitem{li2023seed}
B.~Li, R.~Wang, G.~Wang, Y.~Ge, Y.~Ge, and Y.~Shan, ``{Seed-Bench}: Benchmarking multimodal {LLMs} with generative comprehension,'' \emph{arXiv:2307.16125}, 2023.

\bibitem{an2023openleaf}
J.~An, Z.~Yang, L.~Li, J.~Wang, K.~Lin, Z.~Liu, L.~Wang, and J.~Luo, ``Openleaf: Open-domain interleaved image-text generation and evaluation,'' \emph{arXiv:2310.07749}, 2023.

\bibitem{huang2023vbench}
Z.~Huang, Y.~He, J.~Yu, F.~Zhang, C.~Si, Y.~Jiang, Y.~Zhang, T.~Wu, Q.~Jin, N.~Chanpaisit \emph{et~al.}, ``Vbench: Comprehensive benchmark suite for video generative models,'' \emph{arXiv:2311.17982}, 2023.

\bibitem{liu2023evalcrafter}
Y.~Liu, X.~Cun, X.~Liu, X.~Wang, Y.~Zhang, H.~Chen, Y.~Liu, T.~Zeng, R.~Chan, and Y.~Shan, ``Evalcrafter: Benchmarking and evaluating large video generation models,'' \emph{arXiv:2310.11440}, 2023.

\bibitem{ge2024worldgpt}
Z.~Ge, H.~Huang, M.~Zhou, J.~Li, G.~Wang, S.~Tang, and Y.~Zhuang, ``Worldgpt: Empowering llm as multimodal world model,'' \emph{arXiv:2404.18202}, 2024.

\bibitem{chen2024mllm}
D.~Chen, R.~Chen, S.~Zhang, Y.~Liu, Y.~Wang, H.~Zhou, Q.~Zhang, P.~Zhou, Y.~Wan, and L.~Sun, ``{MLLM}-as-a-judge: Assessing multimodal {LLM}-as-a-judge with vision-language benchmark,'' \emph{arXiv:2402.04788}, 2024.

\bibitem{huang2024visual}
W.~Huang, H.~Liu, M.~Guo, and N.~Z. Gong, ``Visual hallucinations of multi-modal large language models,'' \emph{arXiv:2402.14683}, 2024.

\bibitem{chen2024unified}
X.~Chen, C.~Wang, Y.~Xue, N.~Zhang, X.~Yang, Q.~Li, Y.~Shen, J.~Gu, and H.~Chen, ``Unified hallucination detection for multimodal large language models,'' \emph{arXiv:2402.03190}, 2024.

\bibitem{zhao2024survey}
T.~Zhao, L.~Zhang, Y.~Ma, and L.~Cheng, ``A survey on safe multi-modal learning system,'' \emph{arXiv:2402.05355}, 2024.

\bibitem{niu2024jailbreaking}
Z.~Niu, H.~Ren, X.~Gao, G.~Hua, and R.~Jin, ``Jailbreaking attack against multimodal large language model,'' \emph{arXiv:2402.02309}, 2024.

\bibitem{ying2024mmt}
K.~Ying, F.~Meng, J.~Wang, Z.~Li, H.~Lin, Y.~Yang, H.~Zhang, W.~Zhang, Y.~Lin, S.~Liu \emph{et~al.}, ``{MMT}-bench: A comprehensive multimodal benchmark for evaluating large vision-language models towards multitask {AGI},'' \emph{arXiv:2404.16006}, 2024.

\bibitem{xia2024chartx}
R.~Xia, B.~Zhang, H.~Ye, X.~Yan, Q.~Liu, H.~Zhou, Z.~Chen, M.~Dou, B.~Shi, J.~Yan \emph{et~al.}, ``Chartx \& chartvlm: A versatile benchmark and foundation model for complicated chart reasoning,'' \emph{arXiv:2402.12185}, 2024.

\bibitem{masry2022chartqa}
A.~Masry, D.~X. Long, J.~Q. Tan, S.~Joty, and E.~Hoque, ``Chartqa: A benchmark for question answering about charts with visual and logical reasoning,'' \emph{arXiv:2203.10244}, 2022.

\bibitem{zhang2024if}
L.~Zhang, X.~Zhai, Z.~Zhao, Y.~Zong, X.~Wen, and B.~Zhao, ``What if the tv was off? examining counterfactual reasoning abilities of multi-modal language models,'' in \emph{CVPR}, 2024, pp. 21\,853--21\,862.

\bibitem{chen2024m}
Q.~Chen, L.~Qin, J.~Zhang, Z.~Chen, X.~Xu, and W.~Che, ``M$^3$cot: A novel benchmark for multi-domain multi-step multi-modal chain-of-thought,'' \emph{arXiv:2405.16473}, 2024.

\bibitem{baldassini2024makes}
F.~B. Baldassini, M.~Shukor, M.~Cord, L.~Soulier, and B.~Piwowarski, ``What makes multimodal in-context learning work?'' in \emph{CVPR}, 2024, pp. 1539--1550.

\bibitem{tu2023sight}
H.~Tu, B.~Zhao, C.~Wei, and C.~Xie, ``Sight beyond text: Multi-modal training enhances llms in truthfulness and ethics,'' \emph{arXiv:2309.07120}, 2023.

\bibitem{xu2024data}
C.~Xu, G.~Saranathan, M.~P. Alam, A.~Shah, J.~Lim, S.~Y. Wong, F.~Martin, and S.~Bhattacharya, ``Data efficient evaluation of large language models and text-to-image models via adaptive sampling,'' \emph{arXiv:2406.15527}, 2024.

\bibitem{du2023makes}
Y.~Du, H.~Guo, K.~Zhou, W.~X. Zhao, J.~Wang, C.~Wang, M.~Cai, R.~Song, and J.-R. Wen, ``What makes for good visual instructions? synthesizing complex visual reasoning instructions for visual instruction tuning,'' \emph{arXiv:2311.01487}, 2023.

\bibitem{li20243dmit}
Z.~Li, C.~Zhang, X.~Wang, R.~Ren, Y.~Xu, R.~Ma, and X.~Liu, ``3dmit: 3d multi-modal instruction tuning for scene understanding,'' \emph{arXiv:2401.03201}, 2024.

\bibitem{li2023videochat}
K.~Li, Y.~He, Y.~Wang, Y.~Li, W.~Wang, P.~Luo, Y.~Wang, L.~Wang, and Y.~Qiao, ``Videochat: Chat-centric video understanding,'' \emph{arXiv:2305.06355}, 2023.

\bibitem{han2023chartllama}
Y.~Han, C.~Zhang, X.~Chen, X.~Yang, Z.~Wang, G.~Yu, B.~Fu, and H.~Zhang, ``{ChartLlama}: A multimodal llm for chart understanding and generation,'' \emph{arXiv:2311.16483}, 2023.

\bibitem{zhan2024anygpt}
J.~Zhan, J.~Dai, J.~Ye, Y.~Zhou, D.~Zhang, Z.~Liu, X.~Zhang, R.~Yuan, G.~Zhang, L.~Li \emph{et~al.}, ``{AnyGPT}: Unified multimodal {LLM} with discrete sequence modeling,'' \emph{arXiv:2402.12226}, 2024.

\bibitem{gu2023compodiff}
G.~Gu, S.~Chun, W.~Kim, H.~Jun, Y.~Kang, and S.~Yun, ``Compodiff: Versatile composed image retrieval with latent diffusion,'' \emph{arXiv:2303.11916}, 2023.

\bibitem{ventura2024covr}
L.~Ventura, A.~Yang, C.~Schmid, and G.~Varol, ``Covr: Learning composed video retrieval from web video captions,'' in \emph{AAAI}, vol.~38, no.~6, 2024, pp. 5270--5279.

\bibitem{brooks2023instructpix2pix}
T.~Brooks, A.~Holynski, and A.~A. Efros, ``Instructpix2pix: Learning to follow image editing instructions,'' in \emph{CVPR}, 2023, pp. 18\,392--18\,402.

\bibitem{levy2024data}
M.~Levy, R.~Ben-Ari, N.~Darshan, and D.~Lischinski, ``Data roaming and quality assessment for composed image retrieval,'' in \emph{AAAI}, vol.~38, no.~4, 2024, pp. 2991--2999.

\bibitem{ma2024aligned}
S.~Ma, L.~Wang, S.~Hou, and B.~Yan, ``Aligned with llm: a new multi-modal training paradigm for encoding fmri activity in visual cortex,'' \emph{arXiv:2401.03851}, 2024.

\bibitem{zhang2023moqagpt}
L.~Zhang, Y.~Wu, F.~Mo, J.-Y. Nie, and A.~Agrawal, ``Moqagpt: Zero-shot multi-modal open-domain question answering with large language model,'' \emph{arXiv:2310.13265}, 2023.

\bibitem{huang2024aesexpert}
Y.~Huang, X.~Sheng, Z.~Yang, Q.~Yuan, Z.~Duan, P.~Chen, L.~Li, W.~Lin, and G.~Shi, ``{AesExpert}: Towards multi-modality foundation model for image aesthetics perception,'' \emph{arXiv:2404.09624}, 2024.

\bibitem{tang2024pdfchatannotator}
Y.~Tang, C.-M. Chang, and X.~Yang, ``Pdfchatannotator: A human-llm collaborative multi-modal data annotation tool for pdf-format catalogs,'' in \emph{IUI}, 2024, pp. 419--430.

\bibitem{zhu2024large}
Y.~Zhu, H.~Yuan, S.~Wang, J.~Liu, W.~Liu, C.~Deng, Z.~Dou, and J.-R. Wen, ``Large language models for information retrieval: A survey,'' \emph{arXiv:2308.07107}, 2023.

\bibitem{jing2024large}
Z.~Jing, Y.~Su, Y.~Han, B.~Yuan, C.~Liu, H.~Xu, and K.~Chen, ``When large language models meet vector databases: A survey,'' \emph{arXiv:2402.01763}, 2024.

\bibitem{fernandez2023large}
R.~C. Fernandez, A.~J. Elmore, M.~J. Franklin, S.~Krishnan, and C.~Tan, ``How large language models will disrupt data management,'' \emph{Proceedings of the VLDB Endowment}, vol.~16, no.~11, pp. 3302--3309, 2023.

\bibitem{long2024generative}
X.~Long, J.~Zeng, F.~Meng, Z.~Ma, K.~Zhang, B.~Zhou, and J.~Zhou, ``Generative multi-modal knowledge retrieval with large language models,'' \emph{arXiv:2401.08206}, 2024.

\bibitem{miret2024llms}
S.~Miret and N.~Krishnan, ``Are llms ready for real-world materials discovery?'' \emph{arXiv:2402.05200}, 2024.

\bibitem{luo2024does}
Y.~Luo, Z.~Zheng, Z.~Zhu, and Y.~You, ``How does the textual information affect the retrieval of multimodal in-context learning?'' \emph{arXiv:2404.12866}, 2024.

\bibitem{caffagni2024wiki}
D.~Caffagni, F.~Cocchi, N.~Moratelli, S.~Sarto, M.~Cornia, L.~Baraldi, and R.~Cucchiara, ``Wiki-llava: Hierarchical retrieval-augmented generation for multimodal llms,'' \emph{arXiv:2404.15406}, 2024.

\bibitem{xu2024retrieval}
J.~Xu, Y.~Huang, J.~Hou, G.~Chen, Y.~Zhang, R.~Feng, and W.~Xie, ``Retrieval-augmented egocentric video captioning,'' \emph{arXiv:2401.00789}, 2024.

\bibitem{chen2024knowledge}
Z.~Chen, Y.~Zhang, Y.~Fang, Y.~Geng, L.~Guo, X.~Chen, Q.~Li, W.~Zhang, J.~Chen, Y.~Zhu \emph{et~al.}, ``Knowledge graphs meet multi-modal learning: A comprehensive survey,'' \emph{arXiv:2402.05391}, 2024.

\bibitem{lee2024multimodal}
J.~Lee, Y.~Wang, J.~Li, and M.~Zhang, ``Multimodal reasoning with multimodal knowledge graph,'' \emph{arXiv:2406.02030}, 2024.

\bibitem{perot2023lmdx}
V.~Perot, K.~Kang, F.~Luisier, G.~Su, X.~Sun, R.~S. Boppana, Z.~Wang, J.~Mu, H.~Zhang, and N.~Hua, ``{LMDX}: Language model-based document information extraction and localization,'' \emph{arXiv:2309.10952}, 2023.

\bibitem{biswas2024robustness}
A.~Biswas and W.~Talukdar, ``Robustness of structured data extraction from in-plane rotated documents using multi-modal large language models (llm),'' \emph{Journal of Artificial Intelligence Research}, vol.~4, no.~1, pp. 176--195, 2024.

\bibitem{wu2024structured}
H.~Wu, Y.~Yuan, L.~Mikaelyan, A.~Meulemans, X.~Liu, J.~Hensman, and B.~Mitra, ``Structured entity extraction using large language models,'' \emph{arXiv:2402.04437}, 2024.

\bibitem{li2024llms}
J.~Li, H.~Li, D.~Sun, J.~Wang, W.~Zhang, Z.~Wang, and G.~Pan, ``{LLMs} as bridges: Reformulating grounded multimodal named entity recognition,'' \emph{arXiv:2402.09989}, 2024.

\bibitem{zhang2024extract}
B.~Zhang and H.~Soh, ``Extract, define, canonicalize: An llm-based framework for knowledge graph construction,'' \emph{arXiv:2404.03868}, 2024.

\bibitem{he2023using}
W.~He, H.~Ma, S.~Li, H.~Dong, H.~Zhang, and J.~Feng, ``Using augmented small multimodal models to guide large language models for multimodal relation extraction,'' \emph{Applied Sciences}, vol.~13, no.~22, p. 12208, 2023.

\bibitem{hassan2023chatgpt}
M.~M. Hassan, A.~Knipper, and S.~K.~K. Santu, ``Chatgpt as your personal data scientist,'' \emph{arXiv:2305.13657}, 2023.

\bibitem{cheng2023gpt}
L.~Cheng, X.~Li, and L.~Bing, ``Is gpt-4 a good data analyst?'' \emph{arXiv:2305.15038}, 2023.

\bibitem{hu2024mplug}
A.~Hu, H.~Xu, J.~Ye, M.~Yan, L.~Zhang, B.~Zhang, C.~Li, J.~Zhang, Q.~Jin, F.~Huang \emph{et~al.}, ``mplug-docowl 1.5: Unified structure learning for ocr-free document understanding,'' \emph{arXiv:2403.12895}, 2024.

\bibitem{luo2024llm}
Y.~Luo, R.~An, B.~Zou, Y.~Tang, J.~Liu, and S.~Zhang, ``{LLM} as dataset analyst: Subpopulation structure discovery with large language model,'' \emph{arXiv:2405.02363}, 2024.

\bibitem{hu2023mplug}
A.~Hu, Y.~Shi, H.~Xu, J.~Ye, Q.~Ye, M.~Yan, C.~Li, Q.~Qian, J.~Zhang, and F.~Huang, ``mplug-paperowl: Scientific diagram analysis with the multimodal large language model,'' \emph{arXiv:2311.18248}, 2023.

\bibitem{jiang2024bridging}
F.~Jiang, K.~Wang, and H.~Li, ``Bridging research and readers: A multi-modal automated academic papers interpretation system,'' \emph{arXiv:2401.09150}, 2024.

\bibitem{cai2024sciassess}
H.~Cai, X.~Cai, J.~Chang, S.~Li, L.~Yao, C.~Wang, Z.~Gao, Y.~Li, M.~Lin, S.~Yang \emph{et~al.}, ``Sciassess: Benchmarking llm proficiency in scientific literature analysis,'' \emph{arXiv:2403.01976}, 2024.

\bibitem{wu2024daco}
X.~Wu, R.~Zheng, J.~Sha, T.-L. Wu, H.~Zhou, M.~Tang, K.-W. Chang, N.~Peng, and H.~Huang, ``Daco: Towards application-driven and comprehensive data analysis via code generation,'' \emph{arXiv:2403.02528}, 2024.

\bibitem{yang2024posterllava}
T.~Yang, Y.~Luo, Z.~Qi, Y.~Wu, Y.~Shan, and C.~W. Chen, ``Posterllava: Constructing a unified multi-modal layout generator with llm,'' \emph{arXiv:2406.02884}, 2024.

\bibitem{zhang2023data}
W.~Zhang, Y.~Shen, W.~Lu, and Y.~Zhuang, ``Data-copilot: Bridging billions of data and humans with autonomous workflow,'' \emph{arXiv:2306.07209}, 2023.

\bibitem{dibia2023lida}
V.~Dibia, ``{LIDA}: A tool for automatic generation of grammar-agnostic visualizations and infographics using large language models,'' in \emph{ACL}, 2023.

\bibitem{vazquez2024llms}
P.-P. V{\'a}zquez, ``Are llms ready for visualization?'' in \emph{PacificVis}.\hskip 1em plus 0.5em minus 0.4em\relax IEEE, 2024, pp. 343--352.

\bibitem{wu2024automated}
Y.~Wu, Y.~Wan, H.~Zhang, Y.~Sui, W.~Wei, W.~Zhao, G.~Xu, and H.~Jin, ``Automated data visualization from natural language via large language models: An exploratory study,'' \emph{PACMMOD}, vol.~2, no.~3, pp. 1--28, 2024.

\bibitem{he2024llms}
Y.~He, Z.~Liu, J.~Chen, Z.~Tian, H.~Liu, X.~Chi, R.~Liu, R.~Yuan, Y.~Xing, W.~Wang \emph{et~al.}, ``Llms meet multimodal generation and editing: A survey,'' \emph{arXiv:2405.19334}, 2024.

\bibitem{chen2024panda}
T.-S. Chen, A.~Siarohin, W.~Menapace, E.~Deyneka, H.-w. Chao, B.~E. Jeon, Y.~Fang, H.-Y. Lee, J.~Ren, M.-H. Yang \emph{et~al.}, ``Panda-70m: Captioning 70m videos with multiple cross-modality teachers,'' \emph{arXiv:2402.19479}, 2024.

\bibitem{wang2023internvid}
Y.~Wang, Y.~He, Y.~Li, K.~Li, J.~Yu, X.~Ma, X.~Li, G.~Chen, X.~Chen, Y.~Wang \emph{et~al.}, ``{InternVid}: A large-scale video-text dataset for multimodal understanding and generation,'' \emph{arXiv:2307.06942}, 2023.

\bibitem{chen2024sharegpt4video}
L.~Chen, X.~Wei, J.~Li, X.~Dong, P.~Zhang, Y.~Zang, Z.~Chen, H.~Duan, B.~Lin, Z.~Tang \emph{et~al.}, ``Sharegpt4video: Improving video understanding and generation with better captions,'' \emph{arXiv:2406.04325}, 2024.

\bibitem{li2023m}
L.~Li, Y.~Yin, S.~Li, L.~Chen, P.~Wang, S.~Ren, M.~Li, Y.~Yang, J.~Xu, X.~Sun \emph{et~al.}, ``M$^3$it: A large-scale dataset towards multi-modal multilingual instruction tuning,'' \emph{arXiv:2306.04387}, 2023.

\bibitem{meng2024chartassisstant}
F.~Meng, W.~Shao, Q.~Lu, P.~Gao, K.~Zhang, Y.~Qiao, and P.~Luo, ``Chartassisstant: A universal chart multimodal language model via chart-to-table pre-training and multitask instruction tuning,'' \emph{arXiv:2401.02384}, 2024.

\bibitem{xu2023youku}
H.~Xu, Q.~Ye, X.~Wu, M.~Yan, Y.~Miao, J.~Ye, G.~Xu, A.~Hu, Y.~Shi, G.~Xu \emph{et~al.}, ``Youku-mplug: A 10 million large-scale chinese video-language dataset for pre-training and benchmarks,'' \emph{arXiv:2306.04362}, 2023.

\bibitem{he2023align}
B.~He, J.~Wang, J.~Qiu, T.~Bui, A.~Shrivastava, and Z.~Wang, ``Align and attend: Multimodal summarization with dual contrastive losses,'' in \emph{CVPR}, 2023, pp. 14\,867--14\,878.

\bibitem{zhu2023multimodal}
W.~Zhu, J.~Hessel, A.~Awadalla, S.~Y. Gadre, J.~Dodge, A.~Fang, Y.~Yu, L.~Schmidt, W.~Y. Wang, and Y.~Choi, ``{Multimodal C4}: An open, billion-scale corpus of images interleaved with text,'' \emph{arXiv:2304.06939}, 2023.

\bibitem{patraucean2023perception}
V.~Patraucean, L.~Smaira, A.~Gupta, A.~Recasens, L.~Markeeva, D.~Banarse, S.~Koppula, M.~Malinowski, Y.~Yang, C.~Doersch \emph{et~al.}, ``Perception test: A diagnostic benchmark for multimodal video models,'' \emph{NeurIPS}, vol.~36, 2023.

\bibitem{zhang2024mathverse}
R.~Zhang, D.~Jiang, Y.~Zhang, H.~Lin, Z.~Guo, P.~Qiu, A.~Zhou, P.~Lu, K.-W. Chang, P.~Gao \emph{et~al.}, ``Mathverse: Does your multi-modal llm truly see the diagrams in visual math problems?'' \emph{arXiv:2403.14624}, 2024.

\bibitem{dj-sandbox}
D.~Chen, H.~Wang, Y.~Huang, C.~Ge, Y.~Li, B.~Ding, and J.~Zhou, ``Data-juicer sandbox: A comprehensive suite for multimodal data-model co-development,'' \emph{arXiv:2407.11784}, 2024.

\bibitem{zhang2018deep}
Y.~Zhang, T.~Xiang, T.~M. Hospedales, and H.~Lu, ``Deep mutual learning,'' in \emph{CVPR}, 2018, pp. 4320--4328.

\bibitem{bai2022constitutional}
Y.~Bai, S.~Kadavath, S.~Kundu, A.~Askell, J.~Kernion, A.~Jones, A.~Chen, A.~Goldie, A.~Mirhoseini, C.~McKinnon \emph{et~al.}, ``Constitutional ai: harmlessness from ai feedback. 2022,'' \emph{arXiv:2212.08073}, 2022.

\bibitem{leerlaif}
H.~Lee, S.~Phatale, H.~Mansoor, T.~Mesnard, J.~Ferret, K.~R. Lu, C.~Bishop, E.~Hall, V.~Carbune, A.~Rastogi \emph{et~al.}, ``Rlaif vs. rlhf: Scaling reinforcement learning from human feedback with ai feedback,'' in \emph{ICML}, 2024.

\bibitem{yu2024rlaif}
T.~Yu, H.~Zhang, Y.~Yao, Y.~Dang, D.~Chen, X.~Lu, G.~Cui, T.~He, Z.~Liu, T.-S. Chua \emph{et~al.}, ``Rlaif-v: Aligning mllms through open-source ai feedback for super gpt-4v trustworthiness,'' \emph{arXiv:2405.17220}, 2024.

\bibitem{liang2024rich}
Y.~Liang, J.~He, G.~Li, P.~Li, A.~Klimovskiy, N.~Carolan, J.~Sun, J.~Pont-Tuset, S.~Young, F.~Yang \emph{et~al.}, ``Rich human feedback for text-to-image generation,'' in \emph{CVPR}, 2024, pp. 19\,401--19\,411.

\end{thebibliography}

\end{document}